\documentclass[10pt]{article}
\usepackage{float}
\usepackage[utf8]{inputenc}
\usepackage[T1]{fontenc}
\usepackage{xhfill}
\usepackage{ae,aecompl}
\usepackage[none]{hyphenat}
\usepackage{setspace}
\usepackage{pgfplots}
\usepackage{comment}
\pgfplotsset{every tick label/.append style={font=\scriptsize}}
\pgfkeys{/pgf/number format/.cd,1000 sep={\,}}
\relax
\usepackage{amsmath}
\usepackage{mathptmx} 
\usepackage[hmargin=2cm,vmargin=3cm]{geometry}
\usepackage[tablesfirst,notablist,noheads]{}
\usepackage{hyperref}
\usepackage{xspace}
\usepackage[edges]{forest}
\usetikzlibrary{shadows.blur}
\usepackage{booktabs}
\usepackage{multirow}
\usepackage{authblk}
\usepackage{graphicx}
\usepackage{refcount}
\usepackage{enumitem}
\usepackage{xparse}
\usepackage{geometry}
\usepackage{xcolor}
\usepackage{fancyhdr}
\usepackage{titlesec}
\usepackage{indentfirst}
\usepackage{soul}

\hypersetup {
  colorlinks = true,
  urlcolor = {blue}
}
\graphicspath{{.}}
\DeclareGraphicsExtensions{.pdf, .png}

\definecolor{darkolivegreen}{rgb}{0.33, 0.42, 0.18}
\definecolor{darkpastelgreen}{rgb}{0.01, 0.75, 0.24}

\title{Comprehensive Taxonomies of Nature- and Bio-inspired Optimization: Inspiration \textit{versus} Algorithmic Behavior, Critical Analysis\\and Recommendations (from 2020 to 2024)}

\author[1]{\textbf{Daniel Molina}}
\affil[1]{
Department of Computer Science and Artificial Intelligence, 
DaSCI Andalusian Institute of Data Science and Computation Intelligence\\
University of Granada, Spain\\
\{dmolina, savalgl, herrera\}@decsai.ugr.es, jpoyatosamador@ugr.es
}

\author[1]{\textbf{Javier Poyatos}}

\author[2,3]{\textbf{Javier Del Ser}}
\affil[2]{
TECNALIA, Basque Research \& Technology Alliance (BRTA), Spain\\
javier.delser@tecnalia.com
}
\affil[3]{
Dept. of Communications Engineering\\
University of the Basque Country (UPV/EHU), Spain
}

\author[1]{\textbf{Salvador García}}
\author[4]{\\\textbf{Amir Hussain}}
\affil[4]{
  Edinburgh Napier University, United Kingdom\\
  a.hussain@napier.ac.uk
}
\author[1]{\textbf{Francisco Herrera}}

\date{}

\newcommand{\flying}{Flying\xspace}
\newcommand{\water}{Aquatic\xspace}
\newcommand{\lands}{Terrestrial\xspace}
\newcommand{\micro}{Micro\xspace}
\newcommand{\others}{Other\xspace}
\newcommand{\foraging}{Foraging\xspace}
\newcommand{\move}{Movement\xspace}

\begin{document}

\maketitle
\vspace{-1.3cm}

\begin{abstract}

In recent years, bio-inspired optimization has garnered significant attention in the literature. This algorithmic family mimics various biological processes observed in nature to effectively tackle complex optimization problems. The proliferation of nature- and bio-inspired algorithms, accompanied by a plethora of applications, tools, and guidelines, underscores the growing interest in this field. However, the exponential rise in the number of bio-inspired algorithms poses a challenge to the future trajectory of this research domain. Along the five versions of this document, the number of approaches grows incessantly, and where having a new biological description takes precedence over real problem-solving. This document, in its fifth revision since the original published version in \cite{Molina2020comprehensive}, presents two comprehensive taxonomies. One is based on principles of biological similarity, and the other one is based on operational aspects associated with the iteration of population models that initially have a biological inspiration. Therefore, these taxonomies enable researchers to categorize existing algorithmic developments into well-defined classes, considering two criteria: the source of inspiration and the behavior exhibited by each algorithm. Using these taxonomies, we classify 518 algorithms based on nature-inspired and bio-inspired principles. Each algorithm within these categories is thoroughly examined, allowing for a critical synthesis of design trends and similarities, and identifying the most analogous classical algorithm for each proposal. From our analysis, we conclude that a poor relationship is often found between the natural inspiration of an algorithm and its behavior. Furthermore, similarities in terms of behavior between different algorithms are greater than what is claimed in their public disclosure: specifically, we show that more than one-fourth of the reviewed bio-inspired solvers are versions of classical algorithms. The conclusions from the analysis of the algorithms lead to several learned lessons. 

Moreover, in this new update we have decided to take a brief tour of literature towards three broad directions, providing a more extensive approach to the original document:
\begin{itemize}[leftmargin=*]

\item First, we offer a critical perspective on the field following our insights in \cite{Molina2022nature}, highlighting the \emph{good} (a present and future plenty of exciting applications), the \emph{bad} (novel metaphors not leading to innovative solvers), and the \emph{ugly} (poor methodological practices) in metaheuristic optimization, with an expansion of these perspectives. 
\item Second, we revisit evolutionary and bio-inspired algorithms from a threefold approach: i) where we stand and what’s next in evolutionary algorithms and population-based nature and bio-inspired optimization, based on a structured proposal of challenges that were discussed in 2020, but still exist today \cite{DELSER2019220}; ii) a prescription of methodological guidelines for comparing bio-inspired optimization algorithms \cite{LATORRE2021}; and iii) a tutorial on the design, experimentation, and application of metaheuristic algorithms to real-world optimization problems \cite{OSABA2021}. 
\item Third, we perform a brief review of recent studies that propose good practices for designing metaheuristic algorithms, alongside a few highlighted taxonomies, overviews, and general approaches that, far from without attempting to be exhaustive with the literature, showcase the rich activity and attention received by this field in recent years.

\end{itemize}

This updated study ends with an analysis that exposes the double vision of the wide range of proposals that contributed to the field of metaheuristic optimization after five years of analysis: on one hand, we note a lack of analysis of the real optimization challenges and useful proposals instead of new metaheuristics only focused on a basic comparison with very classical problems versus algorithms. On the other hand, we offer a positive vision of the crucial role that population-based optimization models can take in the design of modern Artificial Intelligence algorithms.



This document is an update as of April 2024, and contains 518 algorithms as opposed to the originally published version which amounted to 323 revised metaheuristics. This arXiv document corresponds with an extension (already mentioned) of the 2 following papers, with references:
\begin{itemize}[leftmargin=*]
\item Molina, D., Poyatos, J., Del Ser, J., García, S., Hussain, A., \& Herrera, F. (2020). Comprehensive taxonomies of nature-and bio-inspired optimization: Inspiration versus algorithmic behavior, critical analysis recommendations. Cognitive Computation, 12, 897-939. DOI: \url{https://doi.org/10.1007/s12559-020-09730-8}

\item Molina-Cabrera, D., Poyatos, J., Osaba, E., Del Ser, J., Herrera, F. (2022). Nature- and Bio-inspired Optimization: the Good, the Bad, the Ugly and the Hopeful. DYNA, 97(2). 114-117. DOI: \url{https://doi.org/10.6036/10331}

\end{itemize}

\end{abstract}

\def\keyword{\vspace{.5em}
  {\textbf{\textit{Keywords --}}\,\relax%
  }}
\def\endkeyword{\par}

\section{Introduction}
\label{sec:intro}

Traditional optimization techniques are motivated by the complexity of the problem and the mathematical properties of its fitness function and constraints. However, in many real-world optimization problems, no exact solver can be applied to solve them at an affordable computational cost or within a reasonable time. Moreover, in some cases, there is no analytical form for the problem's objective and constraints. Under such circumstances, the use of traditional techniques has been widely proven to be unsuccessful, thereby calling for the consideration of alternative optimization approaches.

In this context, complexity is not unusual in Nature: a plethora of complex systems, processes and behaviors have evinced a surprising performance to efficiently address intricate optimization tasks. The most clear example can be found in the different animal species, which have developed over generations very specialized capabilities by evolutionary mechanisms. Indeed, evolution has allowed animals to adapt to harsh environments, foraging, very difficult tasks of orientation, and to resiliently withstand radical climatic changes, among other threats. Animals, when organized in independent systems, groups or swarms or colonies (systems quite complex on their own) have managed to colonize the Earth completely, and eventually achieve a global equilibrium that has permitted them to endure for thousands of years. This renowned success of biological organisms has inspired all kinds of solvers for optimization problems, which have been so far referred to as \emph{bio-inspired optimization algorithms}. This family of optimization methods simulates biological processes such as natural evolution, where solutions are represented by individuals that reproduce and mutate to generate new, potentially improved candidate solutions for the problem at hand. 

Disregarding their source of inspiration, there is clear evidence of the increasing popularity and notoriety gained by nature- and bio-inspired optimization algorithms in the last two decades. This momentum finds its reason in the capability of these algorithms to learn, adapt, and provide good solutions to complex problems that otherwise would have remained unsolved. Many overviews have capitalized on this spectrum of algorithms applied to a wide range of problem casuistry, from combinatorial problems \cite{Pintea2014} to large-scale optimization \cite{mahdavi2015metaheuristics}, evolutionary deep learning \cite{MARTINEZ2021161} and other alike. As a result, almost all business sectors have leveraged this success in recent times. 

From a design perspective, nature- and bio-inspired optimization algorithms are usually conceived after observing a natural process or the behavioral patterns of biological organisms, which are then converted into a computational optimization algorithm. New discoveries in Nature and the undoubted increase of worldwide investigation efforts have ignited the interest of the research community in biological processes and their extrapolation to computational problems. As a result, many new bio-inspired meta-heuristics have appeared in the literature, increasing the outbreak of proposals and applications every year. Nowadays, every natural process can be thought to be adaptable and emulated to produce a new meta-heuristic approach, yet with different capabilities of reaching global optimum solutions to optimization problems.
\begin{figure}[ht]
\centering
\includegraphics[width=\textwidth]{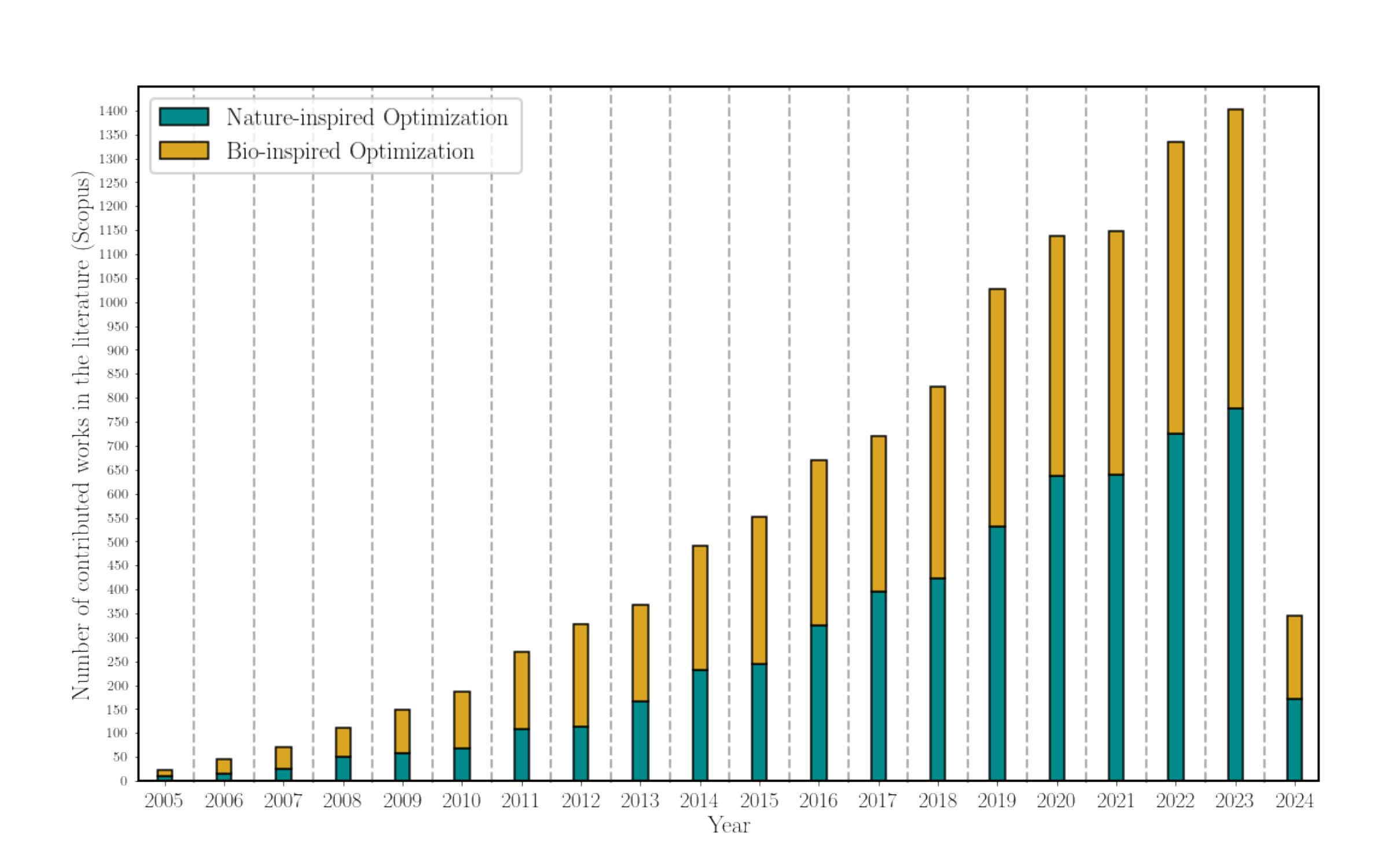}
\caption{\label{fig:papers} Number of papers with \textit{bio-inspired optimization} and \textit{nature-inspired optimization} in the title, abstract and/or keywords, over the period 2005–April 2024 (Scopus database).}
\end{figure}

The above statement is quantitatively supported by Figure \ref{fig:papers}, which depicts the increasing number of papers/book chapters published in the last years with \textit{bio-inspired optimization} and \textit{nature-inspired optimization} in their title, abstract and/or keywords. We have considered both \textit{bio-inspired} and \textit{nature-inspired optimization} because sometimes both terms are confused and indistinctly used, although nature-inspiration includes bio-inspired inspiration and complements it with other sources of inspirations (like physics-based optimization, chemistry-based optimization, ...). A major fraction of the publications comprising this plot proposed new bio-inspired algorithms at their time. From this rising number of nature and bio-inspired algorithms, one can easily conclude that it would be convenient to organize them into a taxonomy with well-defined criteria where to classify both the existing bio-inspired algorithms and new proposals to appear in the future. Unfortunately, the majority of such publications do not include any type of taxonomy, nor do they perform an exhaustive analysis of the similarity of their proposed algorithms concerning other counterparts. Instead, they only focus on the nature or biological metaphor motivating the design of their meta-heuristic. 

This metaphor-driven research trend has been already underscored in several contributions \cite{review8,fister2016new}, which have unleashed hot debates around specific meta-heuristic schemes that remain unresolved to date \cite{weyland2015critical,saka2016metaheuristics}. This problem gets exacerbated when important challenges are overseen and if more and more biological inspirations are used as the primary driver for research, as we can observe in 2024 with more than 500 proposals. It is our firm belief that this controversy could be lessened by a comprehensive taxonomy of nature and bio-inspired optimization algorithms that settled the criteria to justify the novelty and true contributions of current and future advances in the field.

In this fifth version of the original study published in \cite{Molina2020comprehensive}, we have classified 518 works proposing different types of meta-heuristic algorithms. Building upon this knowledge, we herein propose two different taxonomies for nature- and bio-inspired optimization algorithms:
\begin{itemize}[leftmargin=*]
\item The first taxonomy classifies algorithms based on their natural or biological inspiration so that given a specific algorithm, we can find its category quickly and without any ambiguity. The goal of this first taxonomy is to allow easy grouping the upsurge of solvers published in the literature.

\item The second taxonomy classifies the reviewed algorithms based exclusively on their behavior, i.e., how they generate new candidate solutions for the function to be optimized. Our aim is to group together algorithms with similar behavior, without considering its inspirational metaphor. 
\end{itemize}

We believe that this dual criterion can be very useful for researchers. The first one helps classify the different proposals by their origin of inspiration, whereas the second one provides valuable information about their algorithmic similarities and differences. This double classification allows researchers to identify each new proposal in an adequate context. To the best of our knowledge, there has been no previous attempt as ambitious as the one presented in this overview to organize the existing literature on nature- and bio-inspired optimization.

Considering the classifications obtained in our study, we have critically examined the reviewed literature classification in the different taxonomies proposed in this work. The goal is to analyze if there is a relationship between the algorithms classified in the same category in one taxonomy and their classification in the other taxonomy. Next, similarities detected among algorithms will allow discovering the most influential meta-heuristic algorithms, whose behavior has inspired many other nature- and bio-inspired proposals. 

These previous research tasks provide several insights to conduct a comprehensive two-fold analysis of the field:
\begin{itemize}[leftmargin=*]
    \item The first analysis focuses on taxonomies. Specifically, we provide several recommendations to improve research practices in this area. The growing number of nature-inspired proposals could be seen as a symptom of the active status of this field; however, its sharp evolution suggests that research efforts should be also invested towards new behavioral differences and verifiable performance evidence in practical problems.
    \item The second analysis delves into a critical perspective on bio-inspired optimization. It discusses the strengths, weaknesses, and challenges that have been identified in the field in recent years, while it also highlights the potential held for future developments in bio-inspired optimization.
\end{itemize}

Both taxonomies and the analysis provide a full overview of the situation of the bio-inspired optimization field. However, Figure \ref{fig:papers} reflects the interest of research in this field, as the number of papers is in continuous growth of interest. We believe that it is essential to highlight and reflect on what is expected from this field in the coming years, in terms of where it is currently being used and how researchers are proposing methodologies to properly design and apply bio-inspired algorithms not in real-world applications, but also in other emerging areas of Artificial Intelligence (AI). As a consequence, an analysis of the field in terms of \textit{Bio-inspired Optimization}, \textit{Evolutionary Computation}, \textit{Guidelines}, \textit{Comparison Methodology} and \textit{Benchmarking} are found in this report. 

As we have mentioned in the abstract, in this final version of the report we have decided to take a brief tour of literature from three broad perspectives with a more extensive approach to the document: 

In Section \ref{sec:criticalanalysis}, we pay attention from a triple critical position as it was pointed out in \cite{Molina2022nature}, highlighting the \textit{good} (a present and future plenty of exciting applications), the \textit{bad} (novel metaphors not leading to innovative solvers, going deeper into the group of works that criticize the lack of novelty of the new proposals \cite{PIOTROWSKI2014191,review8,weyland2015critical,fister2016new,saka2016metaheuristics,Camacho2020,Tzanetos2021,Aranha2022,Piotrowski2018,Camacho2018,Camacho2019intelligent,Camacho2022,Kudela2022,Campelo2023,Tzanetos2023,Camacho2023b,kudela2023evolutionary}), and the \textit{ugly} (poor methodological practices) as it was pointed out in \cite{Molina2022nature}, with an expansion of these analyses. As we have mentioned, we must emphasize that in these new algorithms, there exists a lack of justification together with the lack of comparison with the state of the art and of real interest in achieving reasonable levels of quality from the point of view of the optimization of well-known problems in recent competitions. Good methodological practices must be followed in forthcoming studies when designing, describing, and comparing new algorithms. 

The analysis of the issues undergone by the field enables us to provide potential solutions and an analysis toward best practices. Hence, in Section \ref{sec:directions}, we introduce three previous works, outlined as follows:
\begin{itemize}[leftmargin=*]
    \item Bio-inspired computation: Where we stand and what’s next \cite{DELSER2019220}.
    \item A prescription of methodological guidelines for comparing bio-inspired optimization algorithms \cite{LATORRE2021}.
    \item A tutorial on the design, experimentation, and application of metaheuristic algorithms to real-world optimization problems \cite{OSABA2021}.
\end{itemize}

Lastly, Section \ref{sec:othertax} presents an analysis of metaheuristics based on studies, guidelines, and other works of a more theoretical nature that help to solve the problems of the field.  We perform a brief review of recent studies that address good practices for designing metaheuristics and discussions from this perspective, and a short review of references -- without attempting to be exhaustive -- that address taxonomies, overviews, and general approaches in bio-inspired optimization. Therefore, this section considers such studies from a double vision: 
\begin{itemize}[leftmargin=*]
    \item Good practices for designing metaheuristics: It gathers several works that are guidelines for good practices related to research orientation to measure novelty \cite{Hu2024}, to measure similarity in metaheuristics \cite{DeArmas2022},  Metaheuristics ``In the Large” (to support the development, analysis, and comparison of new approaches) \cite{Swan2022}, to design manual or automatic new metaheuristics \cite{Camacho2023}, to guide the learning strategy in design and improvement of metaheuristics \cite{Jia2024}, to use statistical test in metaheuristics \cite{Walden2024}, and to detect the novelties in metaphor-based algorithms \cite{Fister2021}.
    \item Latest metaheuristics based studies which include, non-exhaustively, a dozen of recent studies about taxonomies \cite{review2,Rajwar2023,Sarhani2023}, overviews \cite{Tzanetos2021,Ferrer2023,Sharma2024,Velasco2024,Brahim2024,Marti2024} and general approaches \cite{Tang2024}. 
\end{itemize}

This work has been updated almost every year with several improvements, as shown in the trace of changes shown in Table \ref{fig:updates}. The latest version includes both novel bio-inspired proposals up to April 2024 and several analyses of the field, ranging from the situation of the field to the vision towards the future of this field.
\begin{table}[htp]
	\begin{center}
		\caption{Updates of the manuscript via arXiv.}
		\label{fig:updates}
		\begin{tabular}{lll}
			\toprule
			Update  & Date & Contribution \\
			\midrule
			Version \# 1  & Feb. 2020 & Initial version of the manuscript with 323 reviewed algorithms. \\ 
                Version \# 2  & Feb. 2020 & Changes in title and figures for better quality. \\ 
	        Version \# 3  & Apr. 2021 & Update with +31 new algorithms (up to 361), figures and tables changed. \\ 
                Version \# 4  & May 2022 & Update with +51 new algorithms (up to 412), figures and tables changed. \\ 
                Version \# 5  & Apr. 2024 & Update with +88 new algorithms (up to 518); figures and tables changed, and  \\   &  & three analyses included as Sections 7, 8 and 9.\\  
			\bottomrule
		\end{tabular}
	\end{center}
\end{table}

As we have mentioned in the abstract, this fifth and last version of this series of documents ends with an analysis that addresses the double vision of a wide range of proposals, which after five years of analysis must be indicated that they border on a lack of analysis of the real problems and useful proposals, and on the other hand, a positive vision of the role that population-based optimization models can contribute in the design of AI systems, in a new scenario of continuous emergence of AI.

The rest of this paper is organized as follows. In Section \ref{sec:previous}, we examine previous surveys, taxonomies, and reviews of nature- and bio-inspired algorithms reported so far in the literature. Section \ref{sec:taxonomy} delves into the taxonomy based on the inspiration of the algorithms. In Section \ref{sec:behavior}, we present and populate the taxonomy based on the behavior of the algorithm. In Section \ref{sec:analysis-survey}, we analyze similarities and differences found between both taxonomies, ultimately identifying the most influential algorithms in our reviewed papers. In Section \ref{sec:lessons}, we report several lessons learned and recommendations as the result of the previous analysis. In addition, as novel contributions of this version over its preceding ones, Section \ref{sec:criticalanalysis} provides an extended critical analysis of the state of the art in the field, highlighting the aforementioned \emph{good}, the \emph{bad}, and the \emph{ugly} in the metaheuristic landscape \cite{Molina2022nature}. In Section \ref{sec:directions}, we discuss future directions in bio-inspired optimization algorithms, and prescribe potential solutions and analysis toward ensuring good practices and correct experimental procedures with these algorithms. Section \ref{sec:othertax} shows studies and guidelines for good practices, together with recent studies including taxonomies, overviews, and general approaches related to metaheuristics. Finally, in Section \ref{sec:conclusions}, we summarize our current main conclusions and reflections on the field, with builds upon a five-year reflection and literature study.

\section{Related Literature Studies (before 2020 according to the first version of this report, Feb. 2020)} \label{sec:previous}

The diversity of bio-inspired algorithms and their flexibility to tackle optimization problems for many research fields have inspired several surveys and overviews to date. Most of them have focused on different types of problems \cite{KAR201620, walkMH}, including continuous \cite{Molina2018Comp}, combinatorial \cite{Pintea2014}, or multi-objective optimization problems \cite{Zavala2014}. For specific real-world problems, the prolific literature about nature- and bio-inspired algorithms has sparked specific state-of-the-art studies revolving around their application to different fields, such as Telecommunications \cite{BioTele}, Robotics \cite{SI}, Data Mining
\cite{FONG2013385}, Structural Engineering \cite{Zavala2014} or Transportation \cite{del2019bioinspired}. Even specific real-world problems have been dedicated exclusive overviews due to the vast research reported around the topic, like power systems
\cite{PSOPower}, the design of computer networks \cite{DRESSLER2010881}, automatic clustering \cite{JOSEGARCIA2016192}, face recognition \cite{BioFace}, molecular docking \cite{garcia2019bio}, or intrusion detection \cite{KOLIAS2011625}, to mention a few.

Seen from the algorithmic perspective, many particular bio-inspired solvers have been modified over the years to yield different versions analyzed in surveys devoted to that type of algorithms, from classical approaches such as PSO \cite{PSOReview} and DE \cite{Neri2010,DEDas1,DEDas2} to more modern ones, e.g., ABC \cite{Karaboga2014,bereview}, Bacterial Foraging Optimization Algorithm (BFOA, \cite{BFADas}) or the Bat Algorithm \cite{BatReview}. From a more general albeit still algorithmic point of view, \cite{review8} explains how the metaphor and the biological idea are used to create a bio-inspired meta-heuristic optimization algorithm. In this study, the reader is also provided with some examples and characteristics of this design process. Books like \cite{SIBook} or \cite{naturebook} show many nature-inspired algorithms. However, they are more focused on describing the different algorithms available in the literature than on classifying and analyzing them in depth.

Several comparison studies among bio-inspired algorithms with very different behaviors can be found in the current literature, which mostly aims at deciding which approach to use for solving a problem. In \cite{PSODEDas}, the popular PSO and DE versions are compared. This research line is followed by \cite{ELBELTAGI200543}, which compared the performance of different bio-inspired algorithms, again with prescribing which one to use as its primary goal. More recently, \cite{pazhaniraja2017} examined the features of several recent bio-inspired algorithms, suggesting, on a concluding note, to which type of problem each of the examined algorithms should be applied. More specific is the work in \cite{BioGPU}, which compares several different algorithms considering its parallel implementation on GPU devices. More recently, the focus has shifted towards comparing \emph{groups} of algorithms instead of making a comparison between individual solvers: this is the case of \cite{review10}, which compares Swarm Intelligence and Evolutionary Computation methods in order to assess their properties and behavior (e.g., their convergence speed). Once again, the main purpose of this recent literature strand is to compare bio-inspired algorithms, not to classify them nor to find similarities and design patterns among them. In \cite{ForagingvsOther}, foraging algorithms (such as the aforementioned BFOA) are compared against other evolutionary algorithms. In that paper, algorithms are classified into two large groups: algorithms \textit{with} and \textit{without} cooperation. In \cite{chouikhi2019bi,chouikhi2017pso}, PSO was proven to outperform other bio-inspired approaches (namely, DE, GA and ABC) when used for efficiently training and configuring Echo State Networks.

It has not been until relatively recent times that the community has embraced the need for arranging the myriad of existing bio-inspired algorithms and classifying them under principled, coherent criteria. In 2013, \cite{review_nature} presented a classification of meta-heuristic algorithms as per their biological inspiration that discerned categories with similar approaches in this regard: \textit{Swarm Intelligence}, \textit{Physics and Chemistry Based}, \textit{Bio-inspired algorithms (not SI-based)}, and an \textit{Other algorithms} category. However, the classification given in this paper is not actually hierarchical, so it can not be regarded as a true taxonomy. Another classification was proposed in \cite{Venga2015,Venga2016}, composed by \textit{Evolution Based Methods}, \textit{Physics Based Methods}, \textit{Swarm Based Methods}, and \textit{Human-Based Methods}. With respect to the preceding classification, a new \textit{Human-Based} category is proposed, which collectively refers to algorithms inspired by human behavior. The classification criteria underneath these categories are used to build up a catalog of more than 50 algorithmic proposals, obtaining similar groups in size. In this case, there is no \textit{miscellaneous} category as in the previous classification. Similarly to \cite{review_nature}, categories are disjoint groups without subcategories.
 
Recently, \cite{review9} offers a review of meta-heuristics from the 1970s until 2015, i.e., from the development of neural networks to novel algorithms like Cuckoo Search. Specifically, a broad view of new proposals is given, but without proposing any category. The most recent survey to date is that in \cite{Chu2018}, in which nature-inspired algorithms are classified not in terms of their source of inspiration, but rather by their behavior. Consequently, algorithms are classified as per three different principles. The first one is \textit{learning behavior}, namely, how solutions are learned from others preceding them. The learning behavior can be individual, local (i.e., only inside a neighborhood), global (between all individuals), and none (no learning). The second principle is \textit{interaction-collective behavior}, denoting whether individuals cooperate or compete between them. The third principle is referred to as \textit{diversification-population control}, by which algorithms are classified based on whether the population has a converging tendency, a diffuse tendency, or no specific tendency. Then, 29 bio-inspired algorithms are classified by each of these principles, and approaches grouped by each principle are experimentally compared.

The prior related work reviewed above indicates that the community widely acknowledges (with more emphasis in recent times) the need for properly organizing the plethora of bio- and nature-inspired algorithms in a coherent taxonomy. However, the majority of them are only focused on the natural inspiration of the algorithms for creating the taxonomy, not giving any attention to their behavior. This aspect is considered in \cite{Chu2018}, but does not propose a real taxonomy, but rather different independent design principles. On the contrary, as will be next described, our proposed taxonomies consider 1) the source of inspiration; and 2) the procedure by which new solutions are produced over the search process of every algorithm (\emph{behavior}). Furthermore, we note that efforts invested in this regard to date are not up to the level of ambition and thoroughness pursued in our study. In addition, no study published so far has been specifically devoted to unveiling structural similarities between classical and modern meta-heuristics. There lies the novelty and core contribution of our study, along with the aforementioned novel behavior-based taxonomy.

\section{Taxonomy by Source of Inspiration} \label{sec:taxonomy}

In this section, we propose a novel taxonomy based on the inspirational source in which nature- and bio-inspired algorithms are claimed to find their design rationale in the literature. This allows classifying the great amount and variety of contributions reported in related fora.

The proposed taxonomy presented in what follows was developed reviewing 518 papers over different years, starting from the most classical proposals in the late 80's (such as Simulated Annealing \cite{SA} or PSO \cite{PSO}) to more novel techniques appearing in the literature until 2024 \cite{CSA2}. Thus, to our knowledge, this exercise can be considered the most exhaustive review in the area to date.

Taking into account all the reviewed papers, we group the proposals therein in a hierarchy of categories. In the hierarchy, not all proposals of a category must fit in one of its subcategories. In our classification, categories lying at the same level are disjoint sets, which means that each proposed algorithm can be only a member of one of these categories. To this end, for each algorithm, we select the category considered to be most suitable considering the nuances of the algorithm that allow us to differentiate it from its remaining counterparts.

Methodologically, a classification of all nature- and bio-inspired algorithms that can be found in the literature can become complicated, considering the different sources of inspiration as biological, physical, human-being, ... In some papers, authors suggest a possible categorization of more well-established groups, but not in all of them. Also, its classification could not be more appropriate and become eventually obsolete, since the number of proposals -- and thereby, the diversity of sources of inspiration motivating them -- increases over time. Algorithms within each proposed category were selected by their relative importance in the field, considering the number of citations, the number of algorithmic variants that were inspired by that algorithm, and other similar factors.

When establishing a hierarchical classification, it is important to achieve a good trade-off between information and simplicity by the following criteria:
\begin{itemize}[leftmargin=*]
\item When to establish a new division of a category into subcategories: a coarse split criterion for the taxonomy can imply categories of little utility for the subsequent analysis, since in that case, the same category would group very different algorithms. On the other hand, a fine-grained taxonomy can produce very complex hierarchies and, therefore, with little usefulness. To keep the taxonomy simple yet informative for our analytical purposes, we decided that a category should have at least four algorithms in order to be kept in the taxonomy. Thus, a category is only decomposed into subcategories if each of them has coherence and a minimum representativeness (as per the number of algorithms it contains).
  
\item Which number of subcategories into which to divide a category: the criterion followed in this regard must produce meaningful subcategories. In order to ensure a reduced number of subcategories, we consider that not all algorithms inside one category must be a member of one of its subcategories. In that way, we avoid introducing mess categories that lack cohesion.
\end{itemize}

\begin{figure}[!htp]
	\centering
\resizebox{0.6\textwidth}{!}{\begin{forest}
  forked edges,
  for tree={font=\sffamily, rounded corners, minimum width=13em, top color=gray!5, bottom
    color=gray!10, edge+={darkgray, line width=1pt}, draw=darkgray, align=left,
    anchor=children,grow=east,s sep=1cm, l sep=1cm, align=left,
  anchor=west, anchor=base
      west},
  before packing={where n children=3{calign child=2, calign=child edge}{}},
  before typesetting nodes={where content={}{coordinate}{}},
  where level<=1{line width=2pt}{line width=1pt},
  [Nature and population
  -based\\Meta-heuristics (518: 100\%), blur shadow
    [Miscellaneous\\(52: 10.04\%)]
    [Plants Based (23: 4.44\%)]
    [Social Human Behavior\\Algorithms (57: 11.00\%)]
    [Physics and Chemistry\\Based (76: 14.67\%)
	  [Chemistry Based\\(14: 2.70\%)]
	  [Physics Based\\(62: 11.97\%)]
    ]
    [Swarm Intelligence\hspace{3em}\\(277: 53.48\% )
        [Others (36: 6.95\%)]
	    [Microorganisms\\(21: 4.05\%)]
	    [Flying animals\\(85: 16.41\%)]
	    [Terrestrial animals\\(95: 18.35\%)]
	    [Aquatic animals\\(40: 7.72\%)]
    ]
    [Breeding-based Evolution \\(33: 6.37\%)]
    ]
\end{forest}}
\caption{Classification of the reviewed papers using the \emph{inspiration source} based taxonomy.} 
\label{fig:clasi2}
\end{figure}

\begin{figure}[htp]
  \centering
  \includegraphics[width=0.7\textwidth]{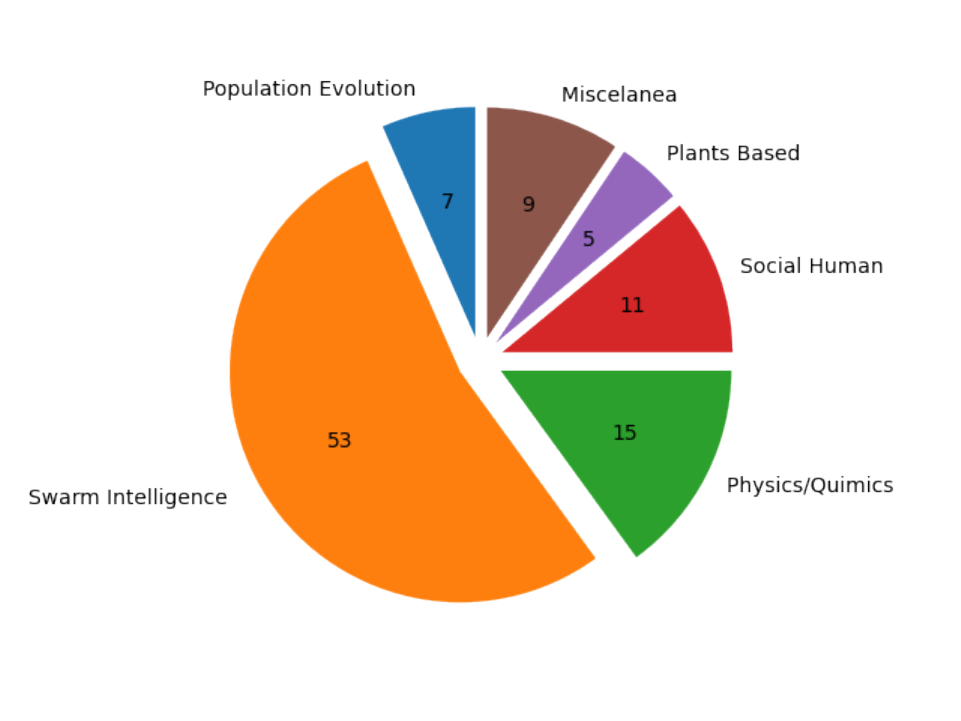}
  \caption{Ratio of reviewed algorithms by its category (first taxonomy).}
  \label{taxonomy_pie} 
\end{figure}

Figure \ref{fig:clasi2} depicts the classification we have reached, indicating, for the 518 reviewed algorithms, the number and ratio of proposals classified in each category and subcategory. It can be observed that the largest group of all is \textit{Swarm Intelligence} category (more than a half of the proposed, 53\%), inspired in the Swarm Intelligence concept \cite{SIBook}, followed by the \textit{Physics and Chemistry} category, inspired by different physical behaviors or chemical reactions (almost 15\% of proposals). Other relevant categories are \textit{Social Human Behavior Algorithms} (11\%), inspired by human aspects, and \textit{Breeding-based Evolution} (near 7\%), inspired by the Theory of Evolution over a population of individuals, that includes very renowned algorithms such as Genetic Algorithms. A new category emerges from our study -- \textit{Plants Based} -- which has not been included in other taxonomies. Nearly 10\% of the proposals are so different between them that they cannot be grouped in new categories. The percentage of classification of each category is visually displayed in Figure \ref{taxonomy_pie}.
 
For the sake of clarity regarding the classification criteria, in the next subsections, we provide a brief description of the different categories in this first taxonomy, including their main characteristics, an example, and a table listing the algorithms belonging to each category.

\subsection{Breeding-based Evolutionary Algorithms}\label{sec:ec}

This category comprises population-based algorithms inspired by the principles of Natural Evolution. Each individual in the population represents a solution of the problem and has an associated fitness value (namely, the value of the problem objective function for that solution). In these algorithms, a process of reproduction (also referred to breeding or crossover) and survival iterated over successive generations makes the population of solutions potentially evolve towards regions of higher optimality (as told by the best solution in the population). Thus, this category is characterized by the fact of being inspired by the concept of breeding-based evolution, without considering other operators performed on individuals than reproduction (e.g., mutation).

More in detail, in these algorithms we have a population with individuals that have the ability to breed and produce new offspring. Therefore, from the parents, we get children, which introduces some variety with respect to their parents. These characteristics allow them to adapt to the environment which, translated to the optimization realm, permits them to search more efficiently over the solution space of the problem at hand. By virtue of this mechanism, we have a population that evolves through generations and, when combined with a selection (survival) and -- possibly -- other operators, results are improved. Nevertheless, the breeding characteristic is what makes algorithms within this category unique with respect to those in other categories.
\begin{table}[htp]
	\begin{center}
    \caption{Nature- and bio-inspired meta-heuristics within the \textit{Breeding-based Evolution} category.}
		\label{fig:ec}
		\begin{tabular}{ l  l  l r}
			\toprule
			\multicolumn{4}{c}{{\Large \textbf{Breeding-based Evolution}}}\\
			\midrule
			Algorithm Name         &    Acronym & Year & Reference\\
			\midrule
			Artificial Ecosystem Algorithm             & AEA  & 2014 & \cite{artificialecosystem} \\
                Artificial Ecosystem Optimizer             & AEO  & 2020 & \cite{Zhao2020} \\
			Artificial Infections Disease Optimization & AIDO & 2016 & \cite{AIDO} \\  	
			Asexual Reproduction Optimization          & ARO  & 2010 & \cite{ARO}  \\  
			Biogeography Based Optimization            & BBO  & 2008 & \cite{BBO}  \\   
			Bird Mating Optimization                   & BMO  & 2014 & \cite{BMO}  \\  
			Bean Optimization Algorithm                & BOA  & 2011 & \cite{BOA}  \\  
                Coronavirus Mask Protection Algorithm      & CMPA & 2023 & \cite{Yuan2023} \\
                Coronavirus Disease Optimization Algorithm & COVIDOA & 2022 & \cite{Khalid2022} \\
			Coral Reefs Optimization                   & CRO  & 2014 & \cite{CRO}  \\  
			Dendritic Cells Algorithm				   & DCA  & 2005 & \cite{greensmith2005introducing} \\
			Differential Evolution                     & DE   & 1997 & \cite{DE}   \\  
			Ecogeography-Based Optimization            & EBO  & 2014 & \cite{EBO}  \\  
			Eco-Inspired Evolutionary Algorithm        & EEA  & 2011 & \cite{EEA}  \\  
			Earthworm Optimization Algorithm           & EOA  & 2018 & \cite{wang2018earthworm} \\
			Evolution Strategies					   & ES   & 2002 & \cite{Beyer2002} \\  
  		Genetic Algorithms                         & GA   & 1989 & \cite{GA}   \\  			
			Gene Expression                            & GE   & 2001 & \cite{GE}   \\  
			Hybrid Rice Optimization                   & HRO  & 2016 & \cite{HRO} \\
            Immune-Inspired Computational Intelligence & ICI  & 2008 & \cite{Cortes} \\
            Improved Genetic Immune Algorithm		   & IGIA & 2017 & \cite{tayeb2017research} \\
			Weed Colonization Optimization             & IWO  & 2006 & \cite{IWO}  \\  
			Marriage In Honey Bees Optimization        & MHBO & 2001 & \cite{MHBO}  \\ 	
			Mushroom Reproduction Optimization         & MRO  & 2018 & \cite{MRO}  \\
			Queen-Bee Evolution                        & QBE  & 2003 & \cite{QBE}  \\ 						
			SuperBug Algorithm						   & SuA  & 2012 & \cite{anandaraman2012new} \\
			Stem Cells Algorithm					   & SCA  & 2011 & \cite{Taherdangkoo2011394} \\
			Sheep Flock Heredity Model                 & SFHM & 2001 & \cite{Nara1999}  \\ 
			Swine Influenza Models Based Optimization  & SIMBO & 2013 & \cite{PATTNAIK2013628} \\
			Self-Organizing Migrating Algorithm        & SOMA & 2004 & \cite{Zelinka2004} \\
                T-Cell Immune Algorithm                    & TCIA & 2023 & \cite{Zhang2023} \\
			Variable Mesh Optimization                 & VMO  & 2012 & \cite{VMO}  \\ 			
			Virulence Optimization Algorithm		   & VOA  & 2016 & \cite{JADERYAN2016596} \\
			\bottomrule
		\end{tabular}
	\end{center}
\end{table}



Table \ref{fig:ec} compiles all reviewed algorithms that fall within this category. As could have been a priori expected, well-known classical Evolutionary Computation algorithms can be observed in this list, such as Genetic Algorithm (GA), Evolution Strategies (ES), and Differential Evolution (DE). However, other algorithms based on the reproduction of different biological organisms are also notable, such as queen bees and weeds.

\subsection{Swarm Intelligence based Algorithms} \label{sec:swarm}

Swarm Intelligence (SI) is already a consolidated term in the community, which was first introduced by Gerardo Beni and Jing Wang in 1989 \cite{SI}. It can be defined as the collective behavior of decentralized, self-organized systems, in either natural or artificial environments. The expression was proposed in the context of robotic systems, but has generalized over the years to denote the emergence of collective intelligence from a group of simple agents, governed by simple behavioral rules. Thus, bio-inspired meta-heuristics based on Swarm Intelligence are those inspired by the collective behavior of animal societies, such as insect colonies or bird flocks, wherein the collective intelligence emerging from the swarm permits to efficiently undertake optimization problems. The seminal bio-inspired algorithm relying on SI concepts was PSO \cite{PSO}, followed shortly thereafter by ACO \cite{ACO}. These early findings around SI concepts applied to optimization spurred many SI-based algorithms in subsequent years, such as ABC \cite{ABC} and more recently, Firefly Algorithm (FA, \cite{FA}) and Grasshopper Optimization Algorithm (GOA, \cite{GOA}).

Reviewed algorithms that fall under the Swarm Intelligence umbrella are shown in Tables \ref{fig:si1}, \ref{fig:si2}, \ref{fig:si3}, \ref{fig:si4}, \ref{fig:si5} and \ref{fig:si6}. This is the most populated category of all our study, characterized by a first category that relates to the type of animal that has inspired each algorithm: as such, we find i) \emph{flying animals}, namely, algorithms inspired in the flying movement of birds and flying animals like insects; ii) \emph{terrestrial animals}, inspired by the foraging and hunter mechanisms of land animals; iii) \emph{aquatic animals}, whose inspiration emerges from the movement of fish schools or other aquatic animals like dolphins; and iv) \emph{microorganisms}, inspired by virus, bacteria, algae and others alike.

Inside each subcategory, we have also distinguished whether they are inspired by
the foraging action of the inspired living creature -- \textit{Foraging-inspired} is in fact another popular term related to this type of inspiration \cite{Foraging2018} -- or, instead, by its movement patterns in general. When the behavior of the algorithm is inspired by both the movement and the foraging behavior of the animal, it is cataloged as foraging inside our first taxonomy. We will next examine in depth each of these subcategories.

\begin{table}[htp]
	\begin{center}
    \caption{Nature- and bio-inspired meta-heuristics within the \textit{Swarm Intelligence} category (I).}
    \label{fig:si1}
\begin{tabular}{p{22em}llllr}
			\toprule
			\multicolumn{6}{c}{{\Large \textbf{Swarm Intelligence (I)}}}\\
			\midrule
			Algorithm Name                                & Acronym & Subcategory & Type & Year & Reference   \\
			\midrule
			Artificial Algae Algorithm                   & AAA     & \micro  & \move     & 2015 & \cite{AAA}  \\  
			Artificial Beehive Algorithm 				 & ABA     & \flying & \foraging & 2009 & \cite{Munoz20091080} \\
			Artificial Bee Colony                        & ABC     & \flying & \foraging & 2007 & \cite{ABC}  \\ 
			Animal Behavior Hunting 					 & ABH     & \others & \foraging & 2014 & \cite{doi:10.1080/00207543.2013.871389} \\
			African Buffalo Optimization                 & ABO     & \lands  & \foraging & 2016 & \cite{ABO}  \\ 
			Andean Condor Algorithm 					 & ACA     & \flying & \foraging & 2019 & \cite{Almonacid2019351} \\
\bottomrule
		\end{tabular}
	\end{center}
\end{table}

\begin{table}[htp]
	\begin{center}
		\caption{Nature- and bio-inspired meta-heuristics within the \textit{Swarm Intelligence} category (II).}
		\label{fig:si2}		
		\begin{tabular}{p{17em}llllr}
			\toprule
			\multicolumn{6}{c}{{\Large \textbf{Swarm Intelligence (II)}}}\\
			\midrule
			Algorithm Name                                       & Acronym & Subcategory & Type & Year & Reference    \\
			\midrule
			Bioluminiscent Swarm Optimization Algorithm  & BSO.1   & \flying & \foraging & 2011 & \cite{Oliveira11} \\
                Biological Survival Optimizer                & BSO.4   & \others & \move & 2023 & \cite{Wang2023} \\
			Bees Swarm Optimization Algorithm            & BSOA    & \flying & \foraging & 2005 & \cite{BSOA} \\   
   		Buzzard Optimization Algorithm               & BUZOA   & \flying & \foraging & 2019 & \cite{BUZOA} \\
			Black Widow Optimization Algorithm           & BWO     & \lands  & \move     & 2020 & \cite{BWO} \\
                Beluga Whale Optimization                    & BWO.1   & \water & \foraging & 2022 & \cite{Zhong2022} \\
\bottomrule
		\end{tabular}
	\end{center}
\end{table}

\begin{table}[htp]
	\begin{center}
		\caption{Nature- and bio-inspired meta-heuristics within the \textit{Swarm Intelligence} category (III).}
		\label{fig:si3}		
		\begin{tabular}{p{22em}llllr}
			\toprule
			\multicolumn{6}{c}{{\Large \textbf{Swarm Intelligence (III)}}}\\
			\midrule
			Algorithm Name           	               & Acronym & Subcategory & Type & Year & Reference    \\
			\midrule
			Emperor Penguin Optimizer                    & EPO     & \lands  & \move & 2018 & \cite{EPO} \\
			Eagle Strategy                               & ES.1    & \flying & \foraging & 2010 & \cite{ES}   \\ 					
			Elephant Search Algorithm                    & ESA     & \lands  & \foraging & 2015 & \cite{ESA}  \\
			Elephant Swarm Water Search Algorithm        & ESWSA   & \lands  & \move & 2018 & \cite{elephantdrought} \\   
   			Egyptian Vulture Optimization Algorithm    	 & EV 	   & \flying & \foraging & 2013 & \cite{EV}   \\  			 	
			Firefly Algorithm                            & FA      & \flying & \foraging & 2009 & \cite{FA}   \\
\bottomrule
		\end{tabular}
	\end{center}
\end{table}

\begin{table}[htp]
	\begin{center}
		\caption{Nature- and bio-inspired meta-heuristics within the \textit{Swarm Intelligence} category (IV).}
		\label{fig:si4}		
		\begin{tabular}{p{18em}llllr}
			\toprule
			\multicolumn{6}{c}{{\Large \textbf{Swarm Intelligence (IV)}}}\\
			\midrule
			Algorithm Name           	               & Acronym & Subcategory & Type & Year & Reference    \\
			\midrule
			Jaguar Algorithm                                     & JA    & \lands  & \foraging & 2015 & \cite{JA}    \\ 
            Jellyfish Search                                     & JS    & \water  & \move  & 2021 & \cite{Chou2021} \\ 
			Japanese Tree Frogs Calling Algorithm                & JTFCA & \lands  & \move & 2012 & \cite{hernandez2012distributed} \\
			Krill Herd                                	   	 	 & KH    & \water  & \foraging & 2012 & \cite{KH}  \\			
                Kookaburra Optimization Algorithm                    & KOA & \flying & \foraging & 2023 & \cite{Dehghani2023c} \\
			Krestrel Search Algorithm                            & KSA   & \flying & \foraging & 2016 & \cite{KSA} \\
\bottomrule
		\end{tabular}
	\end{center}
\end{table}

\begin{table}[htp]
	\begin{center}
		\caption{Nature- and bio-inspired meta-heuristics within the \textit{Swarm Intelligence} category (V).}
		\label{fig:si5}		
		\begin{tabular}{p{18em}llllr}
			\toprule
			\multicolumn{6}{c}{{\Large \textbf{Swarm Intelligence (V)}}}\\
			\midrule
			Algorithm Name           	               & Acronym & Subcategory & Type & Year & Reference    \\
			\midrule
			Pigeon Inspired Optimization						 & PIO   & \flying & \move     & 2014 & \cite{duan2014pigeon} \\
			Population Migration Algorithm 						 & PMA   & \others & \move     & 2009 & \cite{Zhang2009165} \\
                Puma Optimizer                          & PO.1 & \lands & \move & 2024 & \cite{Abdollahzadeh2024} \\
                Pelican Optimization Algorithm                & POA.1 & \flying & \foraging & 2022 & \cite{Trojovsky2023b} \\
                Pufferfish Optimization Algorithm             & POA.2 & \water & \move & 2024 & \cite{Albaik2024} \\
			Prey Predator Algorithm                              & PPA   & \others & \foraging & 2015 & \cite{PPA}   \\
\bottomrule
		\end{tabular}
	\end{center}
\end{table}

\begin{table}[htp]
	\begin{center}
		\caption{Nature- and bio-inspired meta-heuristics within the \textit{Swarm Intelligence} category (VI).}
		\label{fig:si6}		
		\begin{tabular}{p{18em}llllr}
			\toprule
			\multicolumn{6}{c}{{\Large \textbf{Swarm Intelligence (VI)}}}\\
			\midrule
			Algorithm Name           	               & Acronym & Subcategory & Type & Year & Reference    \\
			\midrule
			Squirrel Search Algorithm							 & SSA.1 & \flying & \move	   & 2019 & \cite{JAIN2019148} \\ 
			Salp Swarm Algorithm 								 & SSA.2 & \water  & \foraging & 2017 & \cite{MIRJALILI2017163} \\
            Sparrow Search Algorithm                             & SSA.3 & \flying & \foraging & 2020 & \cite{Jiankai2020} \\
            Sling-shot Spider Optimization                       & S$^2$SO & \lands & \foraging & 2023 & \cite{Adhi2023} \\
			Shark Smell Optimization                             & SSO   & \water  & \foraging & 2016 & \cite{SSO1}  \\ 
			Swallow Swarm Optimization                           & SSO.1 & \flying & \foraging & 2013 & \cite{SSO2}  \\
\bottomrule
		\end{tabular}
	\end{center}
\end{table}

\subsubsection{Subcategories of SI based algorithms by the environment}

As mentioned previously, the global set of Swarm Intelligence algorithms can be divided as a function of the type of animals. Between the possible categories stemming from this criteria, we have grouped them according to the environmental medium inhibited by the inspiring animal (aquatic, terrestrial or aerial). This criterion is not only very intuitive since it inherits a criterion already applied in animal taxonomies, but it also relies on the fact that these environments condition the movement and hunting mode of the different species. As such, the following subcategories have been established:
\begin{itemize}[leftmargin=*]
    \item \textbf{Flying animals:} This category comprises meta-heuristics based on the concept of Swarm Intelligence in which the trajectory of agents is inspired by flying movements, as those observed in birds, bats, or other flying insects. The most well-known algorithms in this subcategory are PSO \cite{PSO} and ABC \cite{ABC}.
    \item \textbf{Terrestrial animals:} Meta-heuristics in this category are inspired by foraging or movements in terrestrial animals. The most renowned approach within this category is the classical ACO meta-heuristic \cite{ACO}, which replicates the stigmergic mechanism used by ants to locate food sources and inform of the existence of their counterparts in the colony. This category also includes other popular algorithms like Grey Wolf Optimization (GWO, \cite{GWO}), inspired in the wild wolf hunting strategy, Lion Optimization Algorithm (LOA, \cite{LOA}), which imitates hunting methods used by these animals, or the more recent Grasshopper Optimization Algorithm (GOA, \cite{GOA}), which finds its motivation in the jumping motion of grasshoppers.
    \item \textbf{Aquatic animals:} This type of meta-heuristic algorithm focuses on aquatic animals. The aquatic ecosystem in which they live has inspired different exploration mechanisms. It contains popular algorithms such as Krill Herd (KH, \cite{KH}), Whale Optimization Algorithm (WOA, \cite{WOA}), and algorithms based on the echolocation used by dolphins to detect fish like Dolphin Partner Optimization (DPO, \cite{DPO}) and Dolphin Echolocation \cite{DE1}.
    \item \textbf{Microorganisms:} Meta-heuristics based on microorganisms are related with the food search performed by bacteria. A bacteria colony performs a movement to search for food. Once they have found and taken it, they split to search again in the environment. Other types of meta-heuristics that can be part of this category are the ones related with virus, which usually replicate the infection process of the cell by virus. The most known algorithm of this category is Bacterial Foraging Optimization Algorithm (BFOA, \cite{BFOA}).
\end{itemize}

\subsubsection{Subcategories of SI based algorithms by the inspirational behavior}

Another criterion to group SI based algorithms is the specific behavior of the animal that captured the attention of researchers and inspired the algorithm. This second criterion is also reflected in Tables \ref{fig:si1}-\ref{fig:si4}, classifying each algorithm as belonging to one of the following behavioral patterns:
\begin{itemize}[leftmargin=*]
\item \textbf{Movement:} We have considered that an algorithm belongs to the \textit{movement inspiration} subcategory if the biological inspiration resides mainly in the way the animal inspiring the algorithm regularly moves around its environment. As such, the differential aspect of the movement could hinge on the dynamics of the movement itself (e.g. the flying movement of birds in PSO \cite{PSO}, jumping actions as in Shuffled Frog-Leaping Algorithm, SFLA \cite{SFLA}, or by aquatic movements as in DPO \cite{DPO}), or by the movement of the population (correspondingly, swarming movements as in Bird Swarm Algorithm, BSA \cite{BSA}, the migration of populations like Population Migration Algorithm, PMA \cite{Zhang2009165}, or the migration of particular animals like salmon \cite{TGSR}, among others).
  
\item \textbf{Foraging:} Rather than the movement strategy, in some other algorithmic variants it is the mechanism used to obtain their food what drives the behavior of the animal and, ultimately, the design of the meta-heuristic algorithm. This foraging behavior can in turn be observed in many flavors, from the tactics used by the animal at hand to surround its food source (as in the aforementioned GWO \cite{GWO} and LA \cite{LA}), inspired in breeding nutrition (as Cuckoo Search \cite{CS,Yang2018}), inspired in hunting techniques observed in grey wolves and lions, respectively), particular mechanisms to locate food sources and communicate its existence to the rest of the swarm (as in ACO \cite{ACO}), or other exploration strategies such as the echolocation in dolphins \cite{DE1}, or the flashing attraction between partners observed in fireflies \cite{FA}. Sometimes, the movement of the animal also obeys to food search and retrieval. In this case, we consider that the algorithm belongs to the \emph{foraging} inspiration type, rather than to the movement type. Nowadays, inspiration by foraging mechanisms is becoming more and more consolidated in the research community, appearing explicitly in the name of several bio-inspired algorithms.
\end{itemize}

\subsection{Physics/Chemistry based Algorithms} \label{sec:phqm}

Algorithms under this category are characterized by the fact that they imitate the behavior of physical or chemical phenomena, such as gravitational forces, electromagnetism, electric charges and water movement (in relation to physics-based approaches), and chemical reactions and gases particles movement as for chemistry-based optimization algorithms.

The complete list of reviewed algorithms in this category is provided in Tables \ref{fig:pbased1} and \ref{fig:pbased2} (physics-based algorithms) and Table \ref{fig:cbased} (chemistry-based methods). In this category we can find some well-known algorithms reported in the last century such as Simulated Annealing \cite{SA}, or one of the most important algorithms in physics-based meta-heuristic optimization, Gravitational Search Algorithm, GSA \cite{GSA}. Interestingly, a variety of space-based algorithms are rooted in GSA, such as Black Hole optimization (BH, \cite{BH}) or Galaxy Based Search Algorithm (GBSA, \cite{GBSA}). Other algorithms such as Harmony Search (HS, \cite{HS}) relate to the music composition process, a human invention that has more in common with other physical algorithms in what refers to the usage of sound waves than with Social Human Behavior based algorithms, the category discussed in what follows.
\begin{table}[htp]
	\begin{center}
		\caption{Nature- and bio-inspired meta-heuristics within the \textit{Physics based} category (I).}
		\label{fig:pbased1}
		\begin{tabular}{ l  l  l r}
			\toprule
			\multicolumn{4}{c}{\Large \textbf{Physics based (I)}}\\
			\midrule
			Algorithm Name                          & Acronym  & Year & Reference                    \\
			\midrule
			Artificial Electric Field Algorithm			& AEFA & 2019 & \cite{ANITA201993} \\
                Archimedes Optimization Algorithm           & AOA.1 & 2021 & \cite{hashim2021archimedes} \\
			Artificial Physics Optimization   			& APO      & 2009 & \cite{Xie20091321}           \\
			Atom Search Optimization                & ASO.1    & 2019 & \cite{ASO1} \\
			Big Bang Big Crunch                     & BBBC     & 2006 & \cite{BBBC}                  \\ 
			Black Hole Optimization                 & BH       & 2013 & \cite{BH}                    \\ 
			Colliding Bodies Optimization           & CBO      & 2014 & \cite{CBO}                   \\ 
			Crystal Energy Optimization Algorithm   & CEO      & 2016 & \cite{CEO}                   \\ 
			Central Force Optimization              & CFO      & 2008 & \cite{CFO}                   \\ 			
			Charged Systems Search                  & CSS      & 2010 & \cite{CSS}                   \\ 
			Electromagnetic Field Optimization      & EFO      & 2016 & \cite{EFO}                   \\ 	
			Electromagnetism Mechanism Optimization & EMO      & 2003 & \cite{EMO}                   \\ 
			Electimize Optimization Algorithm       & EOA.1    & 2011 & \cite{EOA1} \\
			Electron Radar Search Algorithm         & ERSA     & 2020 & \cite{ERSA} \\
			Galaxy Based Search Algorithm            & GBSA     & 2011 & \cite{GBSA}                  \\
			Gravitational Clustering Algorithm 			& GCA      & 1999 & \cite{Kundu19991149}         \\
			Gravitational Emulation Local Search 		& GELS     & 2009 & \cite{Barzegar2009}          \\
			Gravitational Field Algorithm      			& GFA      & 2010 & \cite{Zheng2010}             \\
                Geyser Inspired Algorithm                   & GIA & 2023 & \cite{Ghasemi2023} \\
			Gravitational Interactions Algorithm 		& GIO      & 2011 & \cite{Flores2011226}         \\			
			General Relativity Search Algorithm     & GRSA     & 2015 & \cite{GRSA}                  \\ 
			Gravitational Search Algorithm          & GSA      & 2009 & \cite{GSA}                   \\ 
			Galactic Swarm Optimization             & GSO.2    & 2016 & \cite{GSO2}                  \\
			Hydrological Cycle Algorithm                & HCA   & 2017 & \cite{wedyan2017hydrological} \\
			Harmony Elements Algorithm         			& HEA      & 2009 & \cite{cui2010lambda}         \\
			Hysteresis for Optimization        			& HO       & 2002 & \cite{Zarand2002150201}      \\
			Hurricane Based Optimization Algorithm  & HO.2     & 2014 & \cite{HO}                    \\ 
			Harmony Search                          & HS       & 2005 & \cite{HS}                    \\ 
			Intelligent Gravitational Search Algorithm & IGSA & 2012 & \cite{intelligentGSA} \\
			Intelligence Water Drops Algorithm      & IWD      & 2009 & \cite{IWD}                   \\ 
			Light Ray Optimization             			& LRO      & 2010 & \cite{Shen2010154}           \\
			Lightning Search Algorithm              & LSA      & 2015 & \cite{LSA}                   \\ 
			Magnetic Optimization Algorithm   	 		& MFO.2    & 2008 & \cite{Tayarani20082659}      \\
			Method of Musical Composition      			& MMC      & 2014 & \cite{Mora-Gutierrez2014301} \\
			Melody Search                           & MS.1     & 2011 & \cite{Ashrafi2011109}        \\
			Multi-Verse Optimizer                   & MVO      & 2016 & \cite{MVO}                   \\
            Newton-Raphson-Based Optimizer          & NRBO & 2024 & \cite{Sowmya2024} \\
            Optics Inspired Optimization            & OIO      & 2015 & \cite{OIO}                   \\ 
			Particle Collision Algorithm       			& PCA      & 2007 & \cite{Sacco2007}             \\
			PopMusic Algorithm                 			& PopMusic & 2002 & \cite{Taillard2002613}       \\
			Quantum Superposition Algorithm		     & QSA & 2015 & \cite{saire2015approach} \\
			Rain-Fall Optimization Algorithm        & RFOA & 2017 & \cite{AGHAYKABOLI201731} \\
			Rain Water Algorithm                    & RWA & 2017 & \cite{biyantowater} \\
			River Formation Dynamics                & RFD      & 2007 & \cite{RFD}                   \\ 	
			Radial Movement Optimization            & RMO      & 2014 & \cite{RMO}                   \\ 
			Ray Optimization                        & RO       & 2012 & \cite{RO}                    \\ 		
			\bottomrule
		\end{tabular}
	\end{center}
\end{table}

\begin{table}[htp]
	\begin{center}
		\caption{Nature- and bio-inspired meta-heuristics within the \textit{Physics based} category (II).}
		\label{fig:pbased2}
		\begin{tabular}{ l  l  l r}
			\toprule
			\multicolumn{4}{c}{\Large \textbf{Physics based (II)}}\\
			\midrule
			Algorithm Name                          & Acronym  & Year & Reference                    \\
			\midrule
                Snow Ablation Optimizer                 & SAO & 2023 & \cite{Deng2023} \\
			Space Gravitational Algorithm      			& SGA      & 2005 & \cite{Hsiao20052323}         \\			
			Sonar Inspired Optimization				& SIO & 2017 & \cite{tzanetos2017new} \\
			States Matter Optimization Algorithm    & SMS      & 2014 & \cite{SMS}                   \\ 			
			Spiral Dynamics Optimization            & SO       & 2011 & \cite{SO}                    \\ 
			Spiral Optimization Algorithm      			& SPOA     & 2010 & \cite{jin2010nature}         \\
			Self-Driven Particles                   & SPP      & 1995 & \cite{SPP}                   \\      
                Solar System Algorithm & SSA.4 & 2021 & \cite{Zitouni2021} \\			
			Turbulent Flow of Water-based Optimization & TFWO & 2020 & \cite{TFWO} \\
			Vibrating Particle Systems Algorithm 		& VPO      & 2017 & \cite{VPO}                   \\
			Vortex Search Algorithm                 & VS       & 2015 & \cite{VS}                    \\ 			
			Water Cycle Algorithm                   & WCA.2    & 2012 & \cite{WCA2}                  \\ 
			Water Evaporation Optimization          & WEO      & 2016 & \cite{WEO}                   \\ 
			Water Flow-Like Algorithms         			& WFA      & 2007 & \cite{Yang2007475}           \\
			Water Flow Algorithm               			& WFA.1    & 2007 & \cite{Basu20071825}          \\
			Water-Flow Algorithm Optimization       & WFO      & 2011 & \cite{WFO}                   \\
			Water Wave Optimization Algorithm       & WWA      & 2015 & \cite{WWA}                   \\
			\bottomrule
		\end{tabular}
	\end{center}
\end{table}

\begin{table}[htp]
	\begin{center}
		\caption{Nature- and bio-inspired meta-heuristics within the \textit{Chemistry based} category.}
		\label{fig:cbased}
		\begin{tabular}{ l  l  l r}
			\toprule
			\multicolumn{4}{c}{{\Large \textbf{Chemistry based}}}\\
			\midrule
			Algorithm Name                                      & Acronym & Year & Reference               \\
			Artificial Chemical Process                         & ACP     & 2005 & \cite{Irizarry20055663} \\
			Artificial Chemical Reaction Optimization Algorithm & ACROA   & 2011 & \cite{ACROA}            \\ 
			Artificial Reaction Algorithm                       & ARA     & 2013 & \cite{Melin20133185}    \\							
			Chemical Reaction Optimization Algorithm            & CRO.1   & 2010 & \cite{CRO1}             \\ 
			Gases Brownian Motion Optimization                  & GBMO    & 2013 & \cite{GBMO}             \\ 
			Heat Transfer Search Algorithm                      & HTS    & 2015 & \cite{Patel2015HeatTS}  \\
			Ions Motion Optimization Algorithm                  & IMO     & 2015 & \cite{IMO}              \\ 
			Integrated Radiation Optimization                   & IRO     & 2007 & \cite{Chuang20073157}   \\
			Kinetic Gas Molecules Optimization                  & KGMO    & 2014 & \cite{KGMO}             \\
      Photosynthetic Algorithm                            & PA      & 1999 & \cite{Murase2000115}    \\
      Simulated Annealing                                 & SA.1    & 1989 & \cite{SA}               \\
      Synergistic Fibroblast Optimization                 & SFO     & 2017 & \cite{SFO}              \\ 
			Thermal Exchange Optimization                       & TEO     & 2017 & \cite{TEO}              \\ 	
			\bottomrule
		\end{tabular}
	\end{center}
\end{table}

\subsection{Social Human Behavior based Algorithms} \label{sec:socio}

Algorithms falling in this category are inspired by human social concepts, such as decision-making and ideas related to the expansion/competition of ideologies inside the society as ideology (Ideology Algorithm, IA, \cite{IA}), or political concepts such as the Imperialist Colony Algorithm (ICA, \cite{ICA}). This category also includes algorithms that emulate sports competitions such as the Soccer League Competition Algorithm (SLC, \cite{SLC}). Brainstorming processes have also laid the inspirational foundations of several algorithms such as Brain Storm Optimization algorithm (BSO.2, \cite{BSO}) and Global-Best Brain Storm Optimization algorithm (GBSO, \cite{GBSO}). The complete list of algorithms in this category is given in Table \ref{fig:social} and in Table \ref{fig:social2}.
\begin{table}[htp]
	\begin{center}
		\caption{Nature- and bio-inspired meta-heuristics within the \textit{Social Human Behavior} based category.}
		\label{fig:social}
		\begin{tabular}{ p{22em}  l  l r}
			\toprule
			\multicolumn{4}{c}{{\Large \textbf{Social Human Behavior (I)}}}\\
			\midrule
			Algorithm Name         &    Acronym & Year & Reference\\
			\midrule
			Adolescent Identity Search Algorithm & AISA & 2020 & \cite{AISA} \\
			Anarchic Society Optimization                  & ASO     & 2012 & \cite{ASOBueno}    \\
            Alpine Skiing Optimization                     & ASO.2   & 2022 & \cite{Yuan2022} \\
			Brain Storm Optimization Algorithm             & BSO.2   & 2011 & \cite{BSO}    \\ 	
			Bus Transportation Behavior 				   & BTA     & 2019 & \cite{Bodaghi2019} \\
			Collective Decision Optimization Algorithm     & CDOA    & 2017 & \cite{CDOA}   \\ 
			Cognitive Behavior Optimization Algorithm      & COA.3   & 2016 & \cite{COA1} \\ 			 
			Competitive Optimization Algorithm             & COOA    & 2016 & \cite{COOA} \\			
			Community of Scientist Optimization Algorithm  & CSOA    & 2012 & \cite{scientistopt} \\			
			Cultural Algorithms							   & CA      & 1999 & \cite{Jin19991672} \\
			Duelist Optimization Algorithm				   & DOA     & 2016 & \cite{biyanto2016duelist} \\
			Election Algorithm                             & EA      & 2015 & \cite{EAemami} \\
			Football Game Inspired Algorithms              & FCA.1   & 2009 & \cite{Fadakar20166} \\
			FIFA World Cup Competitions                    & FIFAAO  & 2016 & \cite{FIFAAO} \\ 
			Golden Ball Algorithm						   & GBA     & 2014 & \cite{Osaba2014} \\
			Global-Best Brain Storm Optimization Algorithm & GBSO    & 2017 & \cite{GBSO}   \\ 
			Group Counseling Optimization                  & GCO     & 2010 & \cite{GCO}    \\ 
			Group Leaders Optimization Algorithm 		   & GLOA    & 2011 & \cite{Daskin2011761} \\
			Greedy Politics Optimization Algorithm         & GPO     & 2014 & \cite{GPO}    \\ 		
                Gaining-sharing Knowledge                      & GSK     & 2023 & \cite{Kapoor2023} \\
			Group Teaching Optimization Algorithm          & GTOA    & 2020 & \cite{GTOA}   \\
			Human Evolutionary Model                       & HEM     & 2007 & \cite{HEM}    \\ 
			Human Group Formation						   & HGF     & 2010 & \cite{Thammano20101628} \\
			Human-Inspired Algorithms                      & HIA     & 2009 & \cite{HIA}    \\ 			
			Human Urbanization Algorithm                   & HUA     & 2020 & \cite{HUA}  \\
			Ideology Algorithm                             & IA      & 2016 & \cite{IA}     \\ 
			Imperialist Competitive Algorithm              & ICA     & 2007 & \cite{ICA}    \\ 
			Kho-Kho optimization Algorithm                 & KKOA    & 2020 & \cite{KKOA} \\
			League Championship Algorithm                  & LCA.1   & 2014 & \cite{LCA}    \\ 
			Life Choice Based Optimizer                    & LCBO    & 2020 & \cite{LCBO} \\
			Leaders and Followers Algorithm                & LFA     & 2015 & \cite{Gonzalez-Fernandez2015776} \\
			Old Bachelor Acceptance                        & OBA     & 1995 & \cite{OBA}    \\ 		
			Oriented Search Algorithm 					   & OSA     & 2008 & \cite{Zhang20082856}  \\
                Political Optimizer                            & PO      & 2020 & \cite{Askari2020} \\
			Parliamentary Optimization Algorithm           & POA     & 2008 & \cite{POA}    \\
			Poor and Rich Optimization Algorithm           & PRO     & 2019 & \cite{PRO} \\
			Queuing Search Algorithm					   & QSA.1     & 2018 & \cite{ZHANG2018464} \\
			Search and Rescue Algorithm                    & SAR     & 2019 & \cite{SAR} \\
			Social Behavior Optimization Algorithm         & SBO.1   & 2003 & \cite{SBO1}   \\ 
			Social Cognitive Optimization                  & SCO     & 2002 & \cite{SCOXie} \\
			Social Cognitive Optimization Algorithm		   & SCOA    & 2010 & \cite{Wei201011} \\
			Social Emotional Optimization Algorithm        & SEA     & 2010 & \cite{SEA}    \\
                Stock Exchange Trading Optimization            & SETO    & 2022 & \cite{Emami2022stock} \\
			Stochastic Focusing Search					   & SFS     & 2008 & \cite{Weibo2008583} \\
			Soccer Game Optimization                       & SGO     & 2012 & \cite{SGO}    \\ 
			Soccer League Competition                      & SLC     & 2014 & \cite{SLC}    \\ 
			Student Psychology Optimization Algorithm      & SPBO & 2020 & \cite{SPBO} \\
			\bottomrule
		\end{tabular}
	\end{center}
\end{table}

\begin{table}[htp]
	\begin{center}
		\caption{Nature- and bio-inspired meta-heuristics within the \textit{Social Human Behavior} based category.}
		\label{fig:social2}
		\begin{tabular}{ p{22em}  l  l r}
			\toprule
			\multicolumn{4}{c}{{\Large \textbf{Social Human Behavior (II)}}}\\
			\midrule
			Algorithm Name         &    Acronym & Year & Reference\\
			\midrule
                Stadium Spectators Optimizer                   & SSO.3 & 2024 & \cite{Nemati2024} \\
			Tiki-Taka Algorithm                            & TTA & 2020 &  \cite{TTA} \\
			Team Game Algorithm                            & TGA     & 2018 & \cite{teamgame} \\
			Teaching-Learning Based Optimization Algorithm & TLBO    & 2011 & \cite{TLBO} \\ 
			Thieves and Police Optimization Algorithm      & TPOA    & 2021 & \cite{TPOA} \\
                Tactical Unit Algorithm                        & TUA & 2024 & \cite{Li2024} \\
			Tug Of War Optimization                        & TWO     & 2016 & \cite{TWO}    \\ 
			Unconscious Search							   & US      & 2012 & \cite{Ardjmand2012233} \\
			Volleyball Premier League Algorithm            & VPL     & 2017 & \cite{VPL}    \\
			Wisdom of Artificial Crowds 				   & WAC     & 2011 & \cite{Yampolskiy2011358} \\
			\bottomrule
		\end{tabular}
	\end{center}
\end{table}
\subsection{Plants based Algorithms} \label{sec:plants-based}

This category essentially gathers all optimization algorithms whose search process is inspired by plants. In this case, as opposed to other methods within the Swarm Intelligence category, there is no communication between agents. One of the most well-known is Forest Optimization Algorithms (FOA.1, \cite{FOA1}), inspired by the process of plant reproduction. Table \ref{fig:plants} details the specific algorithms classified in this category.
\begin{table}[htp]
	\begin{center}
		\caption{Nature- and bio-inspired meta-heuristics within the \textit{Plants based} category.}
		\label{fig:plants}
		\begin{tabular}{ l  l  l r}
			\toprule
			\multicolumn{4}{c}{{\Large \textbf{Plants based}}}\\
			\midrule
			Algorithm Name                                 & Acronym & Year & Reference           \\
			\midrule
			Artificial Flora Optimization Algorithm        & AFO     & 2018 & \cite{artificialflora} \\
			Artificial Plants Optimization Algorithm       & APO.1   & 2013 & \cite{APO}          \\ 
			BrunsVigia Flower Optimization Algorithm       & BVOA    & 2018 & \cite{BVOA} \\
                Carnivorous Plant Algorithm                    & CPA     & 2021 & \cite{OngMeng2021} \\
                Discrete Mother Tree Optimization              & DMTO    & 2020 & \cite{Korani2019} \\
			Forest Optimization Algorithm                  & FOA.1   & 2014 & \cite{FOA1}         \\  	
			Flower Pollination Algorithm                   & FPA     & 2012 & \cite{FPA}          \\
                Lotus Effect Algorithm                         & LEA     & 2023 & \cite{Dalirinia2023} \\
			Natural Forest Regeneration Algorithm		   & NFR     & 2016 & \cite{Moez2016} \\	
			Plant Growth Optimization                      & PGO     & 2008 & \cite{PGOplants} \\
			Plant Propagation Algorithm                    & PPA.1   & 2009 & \cite{PPA1}         \\
			Paddy Field Algorithm                          & PFA     & 2009 & \cite{PFA}          \\  
			Root Growth Optimizer                          & RGO     & 2015 & \cite{plantroot} \\
			Root Tree Optimization Algorithm               & RTOA    & 2016 & \cite{yacineplants} \\
			Runner Root Algorithm                          & RRA     & 2015 & \cite{RRA}          \\ 
			Saplings Growing Up Algorithm                  & SGA.1   & 2007 & \cite{Karci2007450} \\
			Self-Defense Mechanism Of The Plants Algorithm & SDMA    & 2018 & \cite{SDMA}         \\
                Seasons Optimization                           & SO.1    & 2022 & \cite{Emami2022} \\
			Strawberry Plant Algorithm                     & SPA     & 2014 & \cite{Strawberryplant} \\
                Smart Root Search                              & SRS     & 2020 & \cite{Naseri2020} \\
			Tree Growth Algorithm						   & TGA.1   & 2019 & \cite{CHERAGHALIPOUR2018393} \\
			Tree Physiology Optimization                   & TPO     & 2018 & \cite{plantphysiology} \\
			Tree Seed Algorithm                            & TSA     & 2015 & \cite{KIRAN20156686} \\
			\bottomrule
		\end{tabular}
	\end{center}
\end{table}

\subsection{Algorithms with Miscellaneous Sources of Inspiration} \label{sec:miscelaneous}

In this category there are included the algorithms that do not fit in any of the previous categories, i.e., we can find algorithms of diverse characteristics such as the Ying-Yang Pair Optimization (YYOP, \cite{YYOP}). Although this defined category is heterogeneous and does not exhibit any uniformity among the algorithms it represents, its inclusion in the taxonomy serves as an exemplifying fact of the very different sources of inspiration existing in the literature. The ultimate goal of reflecting this miscellaneous set of algorithms is to spawn new categories once more algorithms are created by recreating similar inspirational concepts that the assorted ones already present in this category. 

The complete list of algorithms in this category is in Tables \ref{fig:misc2}  and \ref{fig:misc}. In this regard, we stress this pressing need for grouping assorted algorithms in years to come to give rise to new categories. Otherwise, if we just stockpile new algorithms without a clear correspondence to the aforementioned categories in this miscellaneous group, the overall taxonomy will not evolve and will eventually lack its main purpose: to systematically sort and ease the analysis of future advances and achievements in the field.

\begin{table}[!htp]
	\begin{center}
		\caption{Nature- and bio-inspired meta-heuristics within the \textit{Miscellaneous} category.}
		\label{fig:misc2}		
		\begin{tabular}{lllr}
			\toprule
			\multicolumn{4}{c}{{\Large \textbf{Miscellaneous (II)}}}\\
			\midrule
			Algorithm Name         &    Acronym & Year & Reference\\
			\midrule
			Scientifics Algorithms                         & SA.2  & 2014 & \cite{SA1}  \\ 
			Social Engineering Optimization				   & SEO   & 2017 & \cite{fardsocial} \\
			Stochastic Fractal Search                      & SFS.1 & 2015 & \cite{SFS}  \\ 
                Snow Flake Optimization Algorithm              & SFO.1 & 2023 & \cite{Toz2023re} \\
			Search Group Algorithm                         & SGA.2 & 2015 & \cite{SGA}  \\
			Simple Optimization 						   & SOPT  & 2012 & \cite{hasanccebi2012efficient} \\ 
                Ship Rescue Optimization                       & SRO   & 2024 & \cite{Chu2024} \\
			Small World Optimization                       & SWO   & 2006 & \cite{SWO1} \\ 
			The Great Deluge Algorithm 					   & TGD   & 1993 & \cite{Dueck199386} \\
			Wind Driven Optimization                       & WDO   & 2010 & \cite{WDO}   \\ 								
			Ying-Yang Pair Optimization                    & YYOP  & 2016 & \cite{YYOP} \\
			\bottomrule
		\end{tabular}
	\end{center}
\end{table}

\newpage

\begin{table}[!htp]
	\begin{center}
		\caption{Nature- and bio-inspired meta-heuristics within the \textit{Miscellaneous} category.}
		\label{fig:misc}		
		\begin{tabular}{lllr}
			\toprule
			\multicolumn{4}{c}{{\Large \textbf{Miscellaneous (I)}}}\\
			\midrule
			Algorithm Name         &    Acronym & Year & Reference\\
			\midrule
			Atmosphere Clouds Model                        & ACM   & 2013 & \cite{ACM}  \\ 
			Artificial Cooperative Search                  & ACS   & 2012 & \cite{ACS}  \\ 
			Innovative Gunner Algorithm                    & AIG   & 2019 & \cite{AIG} \\
			Across Neighbourhood Search                    & ANS   & 2016 & \cite{ANS}  \\ 
                Botox Optimization Algorithm                   & BOA.2 & 2024 & \cite{Hubalovska2024} \\
			Battle Royale Optimization Algorithm           & BRO   & 2020 & \cite{BRO} \\
			Bar Systems 								   & BS.2  & 2008 & \cite{DelAcebo200818} \\
			Backtracking Search Optimization               & BSO.3 & 2012 & \cite{BSO1} \\ 
			Cloud Model-Based Algorithm 				   & CMBDE & 2012 & \cite{Zhu201255} \\
			Chaos Optimization Algorithm			       & COA.4 & 1998 & \cite{Li1998409} \\ 
			Clonal Selection Algorithm                     & CSA.1   & 2000 & \cite{CSA}  \\ 
			COVID-19 Optimizer Algorithm                   & CVA   & 2020 & \cite{CVA} \\
			Dice Game Optimizer                            & DGO   & 2019 & \cite{DGO} \\
			Dialectic Search                               & DS    & 2009 & \cite{dialecticsearch} \\
			Differential Search Algorithm				   & DSA   & 2012 & \cite{DSA} \\ 
			Exchange Market Algorithm                      & EMA   & 2014 & \cite{EMA}  \\ 
			Extremal Optimization 						   & EO    & 2000 & \cite{Boettcher:1999:EOM:2933923.2934033} \\
                Equilibrium Optimizer                          & EO.1  & 2020 & \cite{Faramarzi2020} \\
			Fireworks Algorithm Optimization               & FAO   & 2010 & \cite{FAO}  \\ 
			Farmland Fertility Algorithm                   & FFA   & 2018 & \cite{FFA} \\
			Grenade Explosion Method                       & GEM   & 2010 & \cite{GEM}  \\ 		
			Golden Sine Algorithm						   & GSA.1 & 2017 & \cite{tanyildizi2017golden} \\
            Golf Sport Inspired Search                     & GSIS    & 2024 & \cite{Husseinzadeh2024} \\
			Heart Optimization                             & HO.1  & 2014 & \cite{HO1}  \\
			Hyper-parameter Dialectic Search               & HDS   & 2020 & \cite{hyperparamdialectic} \\
			Ideological Sublations                         & IS    & 2017 & \cite{hosseini2017ideological} \\
			Interior Search Algorithm                      & ISA   & 2014 & \cite{ISA}  \\ 
			Keshtel Algorithm                              & KA    & 2014 & \cite{KA}   \\ 	
			Kidney-Inspired Algorithm                      & KA.1  & 2017 & \cite{KA1} \\
			Kaizen Programming                             & KP    & 2014 & \cite{KP}   \\ 
                Liver Cancer Algorithm                         & LCA.2   & 2023 & \cite{Houssein2023} \\
                Literature Research Optimizer                  & LRO.1 & 2024 & \cite{NiLei2024} \\
			Membrane Algorithms							   & MA    & 2005 & \cite{Nishida200655} \\
			Mine Blast Algorithm                           & MBA   & 2013 & \cite{MBA}  \\ 
			Neuronal Communication Algorithm			   & NCA   & 2017 & \cite{asil2017new} \\
                Nizar Optimization Algorithm                   & NOA.1 & 2024 & \cite{Khouni2024} \\
                Plasma Generation Optimization                 & PGO.1 & 2020 & \cite{Kaveh2020} \\
			Pearl Hunting Algorithm						   & PHA   & 2012 & \cite{chan2012hyper} \\
			Passing Vehicle Search                         & PVS   & 2016 & \cite{PVS}  \\
			Artificial Raindrop Algorithm                  & RDA.1 & 2014 & \cite{RDA}  \\
			Reactive Dialectic Search                      & RDS   & 2017 & \cite{reactivedialectic} \\
			\bottomrule
		\end{tabular}
	\end{center}
\end{table}

\section{Taxonomy by Behavior for Population based  Nature- and Bio-inspired Optimization}
\label{sec:behavior}

We now proceed with our second proposed taxonomy for population-based metaheuristics. In this case, we sort the different algorithmic proposals reported by the community by their behavior, without any regard to their source of inspiration. To this end, a clear sorting criterion is needed that, while keeping itself agnostic with respect to its inspiration, could summarize as much as possible the different behavioral procedures characterizing the algorithms under review. The criterion adopted for this purpose is the mechanisms used for creating new solutions, or for changing existing solutions to the optimization problem. These are the main features that define the search process of each algorithm.

\begin{figure}[ht]
\centering
\includegraphics[width=\textwidth]{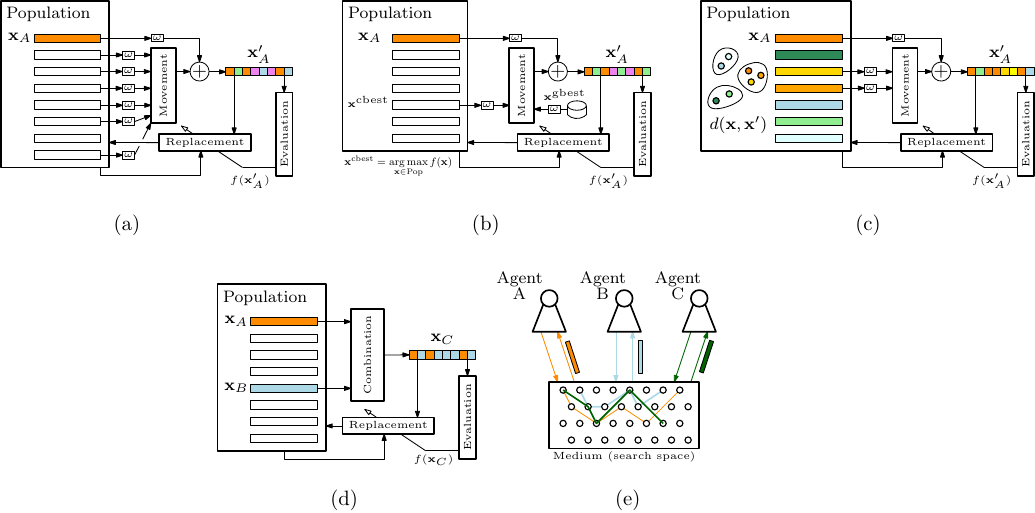}
\caption{\label{fig:behavior}Schematic diagrams of the different algorithmic behaviors on which our second taxonomy relies. The upper plots illustrate the process of generating new solutions by \emph{Differential Vector Movement} from a given solution $\mathbf{x}_A$, using (a) the entire population; (b) relevant individuals (in the example, the movement results from a weighted combination -- $\omega$-- of the current best solution in the population and the best solution found so far by the algorithm); and (c) neighboring solutions in the population to the reference individual. The lower plots show the same process using \emph{solution creation} by (d) combination; and (e) stigmergy.}
\end{figure}

First, we have divided the reviewed optimization algorithms into two categories:

\begin{itemize}[leftmargin=*]
\item \textbf{Differential Vector Movement}, in which new solutions are produced by a shift or a mutation performed onto a previous solution. The newly generated solution could compete against previous ones, or against other solutions in the population to achieve a space and remain therein in subsequent search iterations. This solution generation scheme implies selecting a solution as the reference, which is changed to explore the space of variables and, effectively, produce the search for the solution to the problem at hand. The most representative method of this category is arguably PSO \cite{PSO}, in which each solution evolves with a velocity vector to explore the search domain. Another popular algorithm with differential movement at its core is DE \cite{DEDas2}, in which new solutions are produced by adding differential vectors to existing solutions in the population. Once a solution is selected as the reference one, it is perturbed by adding the difference between other solutions. The decision as to which solutions from the population are influential in the movement is a decision that has an enormous influence on the behavior of the overall search. Consequently, we further divide this category by that decision. The movement -- thus, the search -- can be guided by i) all the population (Figure \ref{fig:behavior}.a); ii) only the significant/relevant solutions, e.g., the best and/or the worst candidates in the population (Figure \ref{fig:behavior}.b); or iii) a small group, which could stand for the neighborhood around each solution or, in algorithms with subpopulations, only the subpopulation to which each solution belongs (Figure \ref{fig:behavior}.c).

\item \textbf{Solution creation}, in which new solutions are not generated by mutation/movement of a single reference solution, but instead by combining several solutions (so there is not only a single parent solution), or other similar mechanism. Two approaches can be utilized for creating new solutions. The first one is by combination, or crossover of several solutions (Figure \ref{fig:behavior}.d). The classical GA \cite{GA} is the most straightforward example of this type. Another approach is by stigmergy (Figure \ref{fig:behavior}.e), in which there is indirect coordination between the different solutions or agents, usually using an intermediate structure, to generate better ones. A classical example of stigmergy for creating solutions is ACO \cite{ACOBook}, in which new solutions are generated by the trace of pheromones left by different agents on a graph representing the solution space of the problem under analysis.
\end{itemize}

Bearing the above criteria in mind, Figure \ref{fig:second_tax} shows the classification reached after our literature analysis. The plot indicates, for the 518 reviewed algorithms, the number and ratio of proposals classified in each category and subcategory. It can be observed that in most nature- and bio-inspired algorithms, new solutions are generated by differential vector movement over existing ones (69\% vs 31\%). Among them, the search process is mainly guided by representative solutions (almost 60\% in global, 86\% from this category), mainly the so-called current best solution (in a very similar fashion to the naive version of the PSO solver). Thus, the creation of new solutions by movement vectors oriented towards the best solution is the search mechanism found in more than half (almost 60\%) of all the 518 reviewed proposals.

\begin{figure}[H]
	\centering
	\begin{forest}
		forked edges,
		for tree={font=\sffamily, rounded corners, top color=gray!5, bottom
			color=gray!10, edge+={darkgray, line width=1pt}, draw=darkgray, align=left,
			anchor=children, s sep=1cm, l sep=1cm, align=left,
			anchor=west, anchor=base
			west},
		before packing={where n children=3{calign child=2, calign=child edge}{}},
		before typesetting nodes={where content={}{coordinate}{}},
		where level<=1{line width=1pt}{line width=1pt},
		[Nature and population-based\\Meta-heuristics (518: 100\%), blur shadow
		[Solution creation\\Based (158: 30.50\%)
		    [Combination \\(149: 28.76\%)]
		    [Stigmergy \\(9: 1.74\%)]
		]
		[Differential vector\\movement (360: 69.50\%)
		    [All Population\\(17: 3.28\%)]
            [Groups Based\\(33: 6.37\%)
		        [Subpopulation\\(27: 5.21\%)]
		        [Neighbourhood\\(6: 1.16\%)]    
		    ]
		    [Representative\\ Based (310: 59.85\%)]
		]
		]
	\end{forest}
	\caption{Classification of the reviewed papers using the \emph{behavior} taxonomy.} 
	\label{fig:second_tax}
\end{figure}

The following subsections provide a brief global view of the different categories introduced above. For each category, we describe its main characteristics, an example, and a table with the algorithms belonging to that category.

\subsection{Differential Vector Movement}
\label{sec:movements}

This category of our behavior-based taxonomy amounts up to 69\% of the analyzed algorithms. In all of them, new solutions are obtained by a movement departing from existing solutions. By using a solution as the reference, a differential vector is used to \emph{move} from the reference towards a new candidate, that could replace the previous one or instead compete to be included into the population.

The crucial decision in differential vector movement is how the differential vector (namely, the intensity and direction of the movement) is calculated. This differential vector could be calculated so as to move the reference solution to another solution (usually a better one), or as a lineal combination of other different solutions, allowing the combination of attraction vectors (toward the best solutions) with repulsion vectors (away from worse ones, or from other solutions, to enforce diversity). The mathematical nature of this operation usually restricts the domain of the representation to a numerical, usually real-valued representation. 

This category is further divided into subcategories as a function of the above decision, i.e. which solutions are considered to create the movement vector. It should be noted that some algorithms can be classified into more than one subcategory. For instance, a particle's update in the PSO solver is affected by the global best particle behavior and certain local best particle(s) behavior. The local best behavior can be either dependent on the particle's previous behavior or the behavior of some particles in its neighborhood. This makes PSO a possible member of two of the subcategories, namely, \emph{Differential Vector as a Function of Representative Solutions} and \emph{Differential Vector as a Function of a Group of Solutions}. Nevertheless, we have considered the classical PSO as a member of \textit{Representative Solutions} because the influence of the best algorithm is stronger than the influence of the neighborhood. In any case, following the above rationale, other PSO variants could fall within any other subcategory. We now describe each of such subcategories.

\subsubsection{Differential Vector as a Function of the Entire Population} \label{sec:mov_all}

One possible criterion is to use all the individuals in the population to generate the movement of each solution. In these algorithms, all individuals have a degree of influence on the movement of the other solutions. Such a degree is usually weighted according to the fitness difference and/or distance between solutions. A significant example is FA \cite{FA}, in which a solution suffers a moving force towards better solutions as a function of their distance. Consequently, solutions closer to the reference solution will have a stronger influence than more distant counterparts. As shown in Table \ref{fig:solmoveAll}, algorithms in this subcategory belong to different categories in the previous \emph{inspiration source} based taxonomy.
\begin{table}[htp]
	\begin{center}
		\caption{Nature- and bio-inspired meta-heuristics within the \emph{Differential Vector Movement} category, wherein the differential vector is influenced by the entire population.}
		\label{fig:solmoveAll}
		\begin{tabular}{ l  l  l r}
			\toprule
			\multicolumn{4}{c}{{\Large \textbf{Influenced by the entire population}}}\\
			\midrule
			Algorithm Name         &    Acronym & Year & Reference\\
			\midrule
			Artificial Electric Field Algorithm					& AEFA & 2019 & \cite{ANITA201993} \\
			Artificial Plants Optimization Algorithm       		& APO.1& 2013 & \cite{APO}  \\ 					
                Botox Optimization Algorithm                   & BOA.2 & 2024 & \cite{Hubalovska2024} \\
			Chaotic Dragonfly Algorithm                  		& CDA  & 2018 & \cite{CDA} \\ 
			Central Force Optimization                          & CFO & 2008 & \cite{CFO}   \\ 							
			Charged Systems Search                              & CSS & 2010 & \cite{CSS}   \\ 
            Dwarf Mongoose Optimization                  & DMO   & 2022 & \cite{Agushaka2022} \\
			Electromagnetism Mechanism Optimization             & EMO & 2003 & \cite{EMO}   \\
			Firefly Algorithm                            & FA    & 2009 & \cite{FA}   \\
			Gravitational Clustering Algorithm 				    & GCA & 1999 & \cite{Kundu19991149} \\			
			Group Counseling Optimization                  		& GCO  & 2010 & \cite{GCO}    \\ 			
			Gravitational Search Algorithm                      & GSA & 2009 & \cite{GSA}   \\ 			
			Human Group Formation						   		& HGF     & 2010 & \cite{Thammano20101628} \\			
			Hoopoe Heuristic Optimization                        & HHO.1  & 2012 & \cite{HHO}   \\ 			
			Intelligent Gravitational Search Algorithm & IGSA & 2012 & \cite{intelligentGSA} \\
			Integrated Radiation Optimization  					& IRO & 2007 & \cite{Chuang20073157} \\	
			Locust Swarms Search                 & LSS  & 2015 & \cite{locuscuevas} \\			
			\bottomrule
		\end{tabular}
	\end{center}
\end{table}

\subsubsection{Differential Vector as a Function of Representative Solutions} \label{sec:representatives}

In this group (the most populated in this second taxonomy), the different movement of each solution is only influenced by a small group of representative solutions. It is often the case that these representative solutions are selected to be the best solutions found by the algorithm (as per the objective of the problem at hand), being able to be guided only by e.g. the current best individual of the population.

Tables \ref{fig:solmoveRepre1}, \ref{fig:solmoveRepre2}, \ref{fig:solmoveRepre3}, \ref{fig:solmoveRepre4}, \ref{fig:solmoveRepre5}, \ref{fig:solmoveRepre6} and \ref{fig:solmoveRepre7} show the different algorithms in this subcategory. An exemplary algorithm of this category that has been a major meta-heuristic solver in the history of the field is PSO \cite{PSO}. In this solver, each solution or particle is guided by the global current best solution and the best solution obtained by that particle during the search. Another classical algorithm in this category is the majority of the family of DE approaches \cite{DEDas2}. In most of the variants of this evolutionary algorithm, the influence of the best solution(s) is hybridized with a differential vector that perturbs the new solution toward random individuals for the sake of increased diversity along the search. However, this subcategory also includes many other algorithms with differences in considering nearly better solutions (as in the Bat Inspired Algorithm \cite{BIA} or the Brain Storm Optimization Algorithm \cite{BSO}) or the worse solutions (to avoid less promising regions), as in the Grasshopper Optimization Algorithm (GOA, \cite{GOA}). More than half of all algorithmic proposals dwell in this subcategory, with a prominence of Swarm Intelligence solvers due to their behavioral inspiration in PSO and DE. We will revolve around these identified similarities in Section~\ref{sec:analysis-survey}. 

\begin{table}[htp]
	\begin{center}
		\caption{Nature- and bio-inspired meta-heuristics within the \emph{Differential Vector Movement} category, wherein the differential vector is influenced by representative solutions (I).}
		\label{fig:solmoveRepre1}
\begin{tabular}{ l  l  l r}
			\toprule
			\multicolumn{4}{c}{{\Large \textbf{Influenced by representative solutions (I)}}}\\
			\midrule
			Algorithm Name         &    Acronym & Year & Reference\\
			\midrule
			Artificial Algae Algorithm                   & AAA  & 2015 & \cite{AAA} \\ 
			Artificial Bee Colony                		 & ABC   & 2007 & \cite{ABC}  \\ 							
			Animal Behavior Hunting 					 & ABH   & 2014 & \cite{doi:10.1080/00207543.2013.871389} \\
			African Buffalo Optimization                 & ABO   & 2016 & \cite{ABO}  \\ 			
			Atmosphere Clouds Model                      & ACM   & 2013 & \cite{ACM}  \\ 	
                Artificial Ecosystem Optimizer             & AEO  & 2020 & \cite{Zhao2020} \\
\bottomrule
		\end{tabular}
	\end{center}
\end{table}

\begin{table}[htp]
	\begin{center}
		\caption{Nature- and bio-inspired meta-heuristics within the \emph{Differential Vector Movement} category, wherein the differential vector is influenced by representative solutions (II).}
		\label{fig:solmoveRepre2}
		\begin{tabular}{ l  l  l r}
			\toprule
			\multicolumn{4}{c}{{\Large \textbf{Influenced by representative solutions (II)}}}\\
			\midrule
			Algorithm Name         &    Acronym & Year & Reference\\
			\midrule
			Catfish Optimization Algorithm               & CAO   & 2011 & \cite{CAO}  \\ 				
			Cheetah Based Algorithm 					 & CBA    & 2018 & \cite{Klein2018685} \\
			Cricket Behavior-Based Algorithm 		     & CBBE  & 2016 & \cite{Canayaz2016362} \\
                Chaotic Crow Search Algorithm                & CCSA  & 2018 & \cite{CCSA} \\
			Collective Decision Optimization Algorithm   & CDOA  & 2017 & \cite{CDOA}   \\
			Camel Herd Algorithm                     & CHA & 2017 & \cite{camelherd} \\
\bottomrule
		\end{tabular}
	\end{center}
\end{table}

\begin{table}[htp]
	\begin{center}
		\caption{Nature- and bio-inspired meta-heuristics within the \emph{Differential Vector Movement} category, wherein the differential vector is influenced by representative solutions (III).}
		\label{fig:solmoveRepre3}
		\begin{tabular}{ l  l  l r}
			\toprule
			\multicolumn{4}{c}{{\Large \textbf{Influenced by representative solutions (III)}}}\\
			\midrule
			Algorithm Name         &    Acronym & Year & Reference\\
			\midrule
			Flock by Leader 							 & FL    & 2012 & \cite{Bellaachia2012117} \\	
               Frilled Lizard Optimization                      & FLO & 2024 & \cite{Falahah2024} \\
			Fruit Fly Optimization Algorithm             & FOA   & 2012 & \cite{FOA}  \\
			Falcon Optimization Algorithm                & FOA.2 & 2019 & \cite{FOA2} \\
			Flower Pollination Algorithm                 & FPA   & 2012 & \cite{FPA}  \\		   
			Fish-Swarm Algorithm                         & FSA  & 2002 & \cite{Li200232} \\
\bottomrule			
		\end{tabular}
	\end{center}
\end{table}

\begin{table}[htp]
	\begin{center}
		\caption{Nature- and bio-inspired meta-heuristics within the \emph{Differential Vector Movement} category, wherein the differential vector is influenced by representative solutions (IV).}
		\label{fig:solmoveRepre4}
		\begin{tabular}{ l  l  l r}
			\toprule
			\multicolumn{4}{c}{{\Large \textbf{Influenced by representative solutions (IV)}}}\\
			\midrule
			Algorithm Name         &    Acronym & Year & Reference\\
			\midrule
			Interior Search Algorithm                    & ISA   & 2014 & \cite{ISA}  \\ 	
			Jaguar Algorithm                             & JA    & 2015 & \cite{JA}    \\ 
            Jellyfish Search                             & JS    & 2021 & \cite{Chou2021} \\ 
			Kidney-Inspired Algorithm                    & KA.1  & 2017 & \cite{KA1} \\
			Kinetic Gas Molecules Optimization           & KGMO  & 2014 & \cite{KGMO}  \\ 			
			Krill Herd                                	 & KH    & 2012 & \cite{KH}  \\
\bottomrule			 									 	
		\end{tabular}
	\end{center}
\end{table}

\begin{table}[htp]
	\begin{center}
		\caption{Nature- and bio-inspired meta-heuristics within the \emph{Differential Vector Movement} category, wherein the differential vector is influenced by representative solutions (V).}
		\label{fig:solmoveRepre5}
		\begin{tabular}{ l  l  l r}
			\toprule
			\multicolumn{4}{c}{{\Large \textbf{Influenced by representative solutions (V)}}}\\
			\midrule
			Algorithm Name         &    Acronym & Year & Reference\\
			\midrule
			Oriented Search Algorithm 					  & OSA  & 2008 & \cite{Zhang20082856}  \\		
            Prairie Dog Optimization Algorithm             & PDO   & 2022 & \cite{Ezugu2022} \\
			Paddy Field Algorithm                          & PFA     & 2009 & \cite{PFA}     \\ 
			Pigeon Inspired Optimization						 & PIO & 2014 & \cite{duan2014pigeon} \\
			Population Migration Algorithm 			      & PMA  & 2009 & \cite{Zhang2009165} \\
                Political Optimizer                            & PO      & 2020 & \cite{Askari2020} \\
\bottomrule			 									 	
		\end{tabular}
	\end{center}
\end{table}

\begin{table}[htp]
	\begin{center}
		\caption{Nature- and bio-inspired meta-heuristics within the \emph{Differential Vector Movement} category, wherein the differential vector is influenced by representative solutions (VI).}
		\label{fig:solmoveRepre6}
		\begin{tabular}{ l  l  l r}
			\toprule
			\multicolumn{4}{c}{{\Large \textbf{Influenced by representative solutions (VI)}}}\\
			\midrule
			Algorithm Name         &    Acronym & Year & Reference\\
			\midrule

			Space Gravitational Algorithm      		& SGA  & 2005 & \cite{Hsiao20052323} \\			
			Soccer Game Optimization                       & SGO     & 2012 & \cite{SGO}    \\   
   		  Spotted Hyena Optimizer			 	    & SHO & 2017 & \cite{dhiman2017spotted} \\
			Selfish Herds Optimizer                 & SHO.1 & 2017 & \cite{SHO1} \\
                Sea-horse Optimizer                     & SHO.2 & 2023 & \cite{Zhao2023b} \\
			Swarm Inspired Projection Algorithm     & SIP  & 2009 & \cite{SIP}   \\
\bottomrule			 									 	
		\end{tabular}
	\end{center}
\end{table}

\begin{table}[htp]
	\begin{center}
		\caption{Nature- and bio-inspired meta-heuristics within the \emph{Differential Vector Movement} category, wherein the differential vector is influenced by representative solutions (VII).}
		\label{fig:solmoveRepre7}
		\begin{tabular}{ l  l  l r}
			\toprule
			\multicolumn{4}{c}{{\Large \textbf{Influenced by representative solutions (VII)}}}\\
			\midrule
			Algorithm Name         &    Acronym & Year & Reference\\
			\midrule   
			Thieves and Police Optimization Algorithm      & TPOA    & 2021 & \cite{TPOA} \\
                Tyrannosaurus Optimization Algorithm             & TROA & 2023 & \cite{Sahu2023} \\
			Tree Seed Algorithm                            & TSA     & 2015 & \cite{KIRAN20156686} \\
			Tunicate Swarm Algorithm                             & TSA.1 & 2020 & \cite{TSA1} \\
                Tangent Search Algorithm                             & TSA.2 & 2022 & \cite{Layeb2022tangent} \\
			Tiki-Taka Algorithm                     & TTA & 2020 &  \cite{TTA} \\
\bottomrule			 									 	
		\end{tabular}
	\end{center}
\end{table}

\subsubsection{Differential Vector as a Function of a Group of Solutions}
\label{sec:groups}

Algorithms within this category do not resort to representative solutions of the entire population (such as the current best), but they only consider solutions of a subset or group of the solutions in the population. When the differential movement considers both a group and a representative of all the population, the algorithm under analysis is considered to belong to the previous subcategory, because the representative has usually the strongest influence over the search. Two different subcategories hold when a group of solutions is used for computing the differential movement vector:
\begin{itemize}[leftmargin=*]
\item \textbf{Subpopulation based differential vector}: In algorithms belonging to this subcategory (listed in Table \ref{fig:solmoveGroupsSub}) the population is divided in several subpopulations, such that the movement of each solution is only affected by the other solutions in the same subpopulation. Examples of algorithms in this subcategory are LA \cite{LA} or the Monarch Butterfly Optimization algorithm (MBO, \cite{MBO}).
\begin{table}[htp]
	\begin{center}
		\caption{Nature- and bio-inspired meta-heuristics within the \emph{Differential Vector Movement} category, wherein the differential vector is influenced by subpopulations.}
		\label{fig:solmoveGroupsSub}
		\begin{tabular}{ l  l  l r}
			\toprule
			\multicolumn{4}{c}{{\Large \textbf{Influenced by subpopulations}}}\\
			\midrule
			Algorithm Name         &    Acronym & Year & Reference\\
			\midrule
			Artificial Chemical Process                         & ACP     & 2005 & \cite{Irizarry20055663} \\
			Artificial Cooperative Search                  & ACS   & 2012 & \cite{ACS}  \\ 			
			Artificial Physics Optimization   			& APO      & 2009 & \cite{Xie20091321} \\
			Bee Colony-Inspired Algorithm 				 & BCIA    & 2009 & \cite{Hackel} \\			
			Colliding Bodies Optimization                & CBO  & 2014 & \cite{CBO}   \\ 			
			Cuttlefish Algorithm                         & CFA   & 2013 & \cite{CFA}  \\  			
			Cuckoo Optimization Algorithm                & COA    & 2011 & \cite{COA}  \\
                Carnivorous Plant Algorithm                    & CPA     & 2021 & \cite{OngMeng2021} \\
			Chicken Swarm Optimization                   & CSO.1   & 2014 & \cite{CSO1} \\ 
			COVID-19 Optimizer Algorithm                   & CVA   & 2020 & \cite{CVA} \\
			Dice Game Optimizer                          & DGO   & 2019 & \cite{DGO} \\
			Exchange Market Algorithm                    & EMA   & 2014 & \cite{EMA}  \\ 			
			Greedy Politics Optimization Algorithm         & GPO     & 2014 & \cite{GPO}    \\ 		
                Gaining-sharing Knowledge                      & GSK     & 2023 & \cite{Kapoor2023} \\
			Group Search Optimizer                       & GSO.1  & 2009 & \cite{GSO1} \\
			Horse Optimization Algorithm                 & HOA  &  2020 & \cite{HOA} \\
			Hierarchical Swarm Model                     & HSM   & 2010 & \cite{HSM}   \\ 			
			Ions Motion Optimization Algorithm           & IMO & 2015 & \cite{IMO}   \\ 
			Life Choice Based Optimizer                  & LCBO    & 2020 & \cite{LCBO} \\
			Lion Optimization Algorithm                  & LOA  & 2016 & \cite{LOA}   \\			
			Monarch Butterfly Optimization               & MBO.1  & 2017 & \cite{MBO}   \\ 			
			Social Behavior Optimization Algorithm       & SBO.1   & 2003 & \cite{SBO1}   \\ 			
			Sperm Whale Algorithm                        & SWA  & 2016 & \cite{SWA}  \\ 			
			Thermal Exchange Optimization                & TEO  & 2017 & \cite{TEO}   \\
			Turbulent Flow of Water-based Optimization   & TFWO & 2020 & \cite{TFWO} \\
			Wisdom of Artificial Crowds 				 & WAC & 2011 & \cite{Yampolskiy2011358} \\			
			Worm Optimization                            & WO  & 2014 & \cite{WO}   \\			
			\bottomrule
		\end{tabular}
	\end{center}
\end{table}

\item \textbf{Neighborhood based differential vector}: In this subcategory, each solution is affected only by solutions in its local neighborhood. Table \ref{fig:solmoveGroupsNeigh} compiles all algorithms that are classified in this subcategory. A notable example in this list is BFOA \cite{BFOA}, in which all solutions in the neighborhood impact on the computation of the movement vector, either by attracting the solution (if the neighboring solution has better fitness than the reference solution) or in a repulsive way (when the neighboring solution is worse than the one to be moved). 
\begin{table}[htp]
	\begin{center}
		\caption{Nature- and bio-inspired meta-heuristics within the \emph{Differential Vector Movement} category, wherein the differential vector is influenced by neighborhoods.}
		\label{fig:solmoveGroupsNeigh}
		\begin{tabular}{ l  l  l r}
			\toprule
			\multicolumn{4}{c}{{\Large \textbf{Influenced by neighbourhoods}}}\\
			\midrule
			Algorithm Name                                                      &       Acronym     & Year      & Reference\\
			\midrule
			Bees Algorithm                                                      &       BA          & 2006      & \cite{BA}   \\ 			
			Biomimicry Of Social Foraging Bacteria for Distributed Optimization &       BFOA        & 2002 & \cite{BFOA} \\ 
		    Bacterial Foraging Optimization 			                        &       BFOA.1      & 2009 & \cite{BFADas} \\ 			
			Gravitational Emulation Local Search 		                        &       GELS        & 2009 & \cite{Barzegar2009}          \\ 
			Neuronal Communication Algorithm			   & NCA   & 2017 & \cite{asil2017new} \\			 Physarum-inspired Competition Algorithm              & PCA.1 & 2023 & \cite{Awad2023novel} \\
			\bottomrule
		\end{tabular}
	\end{center}
\end{table}

\end{itemize}

\subsection{Solution Creation} \label{sec:creation}

This category is composed of algorithms that explore the domain search by generating new solutions, not by moving existing ones. This group is a significant ratio (almost 31\%) of all proposals, and includes many classical algorithms like GA \cite{GA}. A very widely exploited advantage of these methods is the possibility to adapt the generation method to the particular problem, hence allowing for different possible representations and, therefore, easing its application to a wider range of problems. In the following, we describe the different subcategories that result from the diverse mechanisms by which solutions can be created.

\subsubsection{Creation by Combination}

The most common option to generate a new solution is to combine existing ones. In these algorithms, different solutions are selected and combined using a crossover operator or combining method to give rise to new solutions. The underlying idea is that by combining good solutions, even better solutions can be eventually generated. 

The combining method can be specific for the problem to be solved or instead, be conceived for a more general family of problems. In fact, combining methods are usually devised to be adaptable to many different solution representations. As mentioned before, the most popular algorithm in this category is GA \cite{GA}. However, many other bio-inspired algorithms exhibit a similar behavior when creating solutions, yet they are inspired by other phenomena, such as Cultural Optimization (CA, \cite{Jin19991672}) (in the Social Human Behavior category), LA \cite{LOA} (in the Swarm Intelligence category), Particle Collision Algorithm (PCA, \cite{Sacco2007}, in the chemistry-based category) or Light Ray Optimization (LRO, \cite{Shen2010154}, in the physics-based category). Tables \ref{fig:solcreationComb1}, \ref{fig:solcreationComb2}, \ref{fig:solcreationComb3}, and \ref{fig:solcreationComb4} show the algorithms that rely on combination when creating new solutions along their search.

\begin{table}[htp]
	\begin{center}
		\caption{Nature- and bio-inspired meta-heuristics within the \textit{Solution Creation - Combination} category (I).}
		\label{fig:solcreationComb1}
		\begin{tabular}{ l  l  l r}
			\toprule
			\multicolumn{4}{c}{{\Large \textbf{Creation-Combination category (I)}}}\\
			\midrule
			Algorithm Name         &    Acronym & Year & Reference\\
			\midrule
			Artificial Beehive Algorithm 				 & ABA   & 2009 & \cite{Munoz20091080} \\ 
			Andean Condor Algorithm						 & ACA   & 2019 & \cite{Almonacid2019351} \\
			Artificial Chemical Reaction Optimization Algorithm & ACROA & 2011 & \cite{ACROA}\\
			Artificial Ecosystem Algorithm             & AEA  & 2014 & \cite{artificialecosystem} \\
			Artificial Flora Optimization Algorithm        & AFO     & 2018 & \cite{artificialflora} \\	
			Artificial Infections Disease Optimization & AIDO  & 2016 & \cite{AIDO} \\
			Innovative Gunner Algorithm                    & AIG   & 2019 & \cite{AIG} \\
			Anglerfish Algorithm                         & AOA  & 2019 & \cite{AOA} \\
			Artificial Reaction Algorithm      		   & ARA   & 2013 & \cite{Melin20133185} \\
			Asexual Reproduction Optimization          & ARO   & 2010 & \cite{ARO}  \\
                American Zebra Optimization Algorithm        & AZOA  & 2023  & \cite{Mohapatra2023} \\
			Bacterial-GA Foraging 					   & BGAF  & 2007 & \cite{BGAF} \\ 
			Bumblebees                                   & BB   & 2009 & \cite{BB}   \\	
			Biogeography Based Optimization            & BBO   & 2008 & \cite{BBO}  \\   
			Bee Colony Optimization                    & BCO   & 2005 & \cite{BCO}  \\ 
			BeeHive Algorithm                            & BHA  & 2004 & \cite{BHA}  \\ 
			Bees Life Algorithm 					   & BLA   & 2018 & \cite{Bitam2018373} \\			
			Bird Mating Optimization   				   & BMO   & 2014 & \cite{BMO}  \\	
			Barnacles Mating Optimizer                   & BMO.1   &  2019  & \cite{BMO1} \\
			Bean Optimization Algorithm                & BOA   & 2011 & \cite{BOA}  \\ 
			Bull Optimization Algorithm                & BOA.1 & 2015 & \cite{bulloptimization} \\
			Bee System 								   & BS    & 1997 & \cite{Sato19973954}\\
			Bar Systems 							   & BS.2  & 2008 & \cite{DelAcebo200818} \\
			Backtracking Search Optimization           & BSO.3 & 2012 & \cite{BSO1} \\ 					
			Bees Swarm Optimization Algorithm          & BSOA   & 2005 & \cite{BSOA} \\					
			Bus Transportation Behavior 				   & BTA     & 2019 & \cite{Bodaghi2019} \\
			BrunsVigia Flower Optimization Algorithm       & BVOA    & 2018 & \cite{BVOA} \\
			Black Widow Optimization Algorithm           & BWO  & 2020 & \cite{BWO} \\
			Cultural Algorithms						   & CA    & 1999 & \cite{Jin19991672} \\			
			Cultural Coyote Optimization Algorithm	   & CCOA  & 2019 & \cite{Pierezan2019} \\
			Crystal Energy Optimization Algorithm      & CEO   & 2016 & \cite{CEO}   \\ 			
			Consultant Guide Search                    & CGS   & 2010 & \cite{CGS}\\
                Coronavirus Mask Protection Algorithm      & CMPA & 2023 & \cite{Yuan2023} \\
                Coronavirus Disease Optimization Algorithm & COVIDOA & 2022 & \cite{Khalid2022} \\
			Coral Reefs Optimization                   & CRO   & 2014 & \cite{CRO}  \\  			
			Chemical Reaction Optimization Algorithm   & CRO.1 & 2010 & \cite{CRO1}  \\ 			
			Cuckoo Search                                & CS  & 2009 & \cite{CS}   \\  
			Clonal Selection Algorithm                 & CSA.1   & 2000 & \cite{CSA}  \\ 	
			Dragonfly Swarm Algorithm                    & DA.1  & 2020 & \cite{bhardwaj2020dragonfly} \\
			Dendritic Cells Algorithm				   & DCA  & 2005 & \cite{greensmith2005introducing} \\
			Dolphin Echolocation                         & DE.1 & 2013 & \cite{DE1}  \\
                Discrete Mother Tree Optimization              & DMTO    & 2020 & \cite{Korani2019} \\
			Duelist Optimization Algorithm				   & DOA     & 2016 & \cite{biyanto2016duelist} \\
			Dialectic Search                               & DS    & 2009 & \cite{dialecticsearch} \\
			Election Algorithm                         & EA & 2015 & \cite{EAemami} \\
			Ecogeography-Based Optimization            & EBO   & 2014 & \cite{EBO}  \\ 				
			Eco-Inspired Evolutionary Algorithm        & EEA   & 2011 & \cite{EEA}  \\  
			Electromagnetic Field Optimization         & EFO   & 2016 & \cite{EFO}   \\
			\bottomrule
		\end{tabular}
	\end{center}
\end{table}

\begin{table}[htp]
	\begin{center}
		\caption{Nature- and bio-inspired meta-heuristics within the \textit{Solution Creation - Combination} category (II).}
		\label{fig:solcreationComb2}
		\begin{tabular}{ l  l  l r}
			\toprule
			\multicolumn{4}{c}{{\Large \textbf{Creation-Combination category (II)}}}\\
			\midrule
			Algorithm Name         &    Acronym & Year & Reference\\
			\midrule
    		Electric Fish Optimization                 & EFO.1 & 2020 & \cite{Yilmaz2020} \\
			Extremal Optimization 					   & EO    & 2000 & \cite{Boettcher:1999:EOM:2933923.2934033} \\		
                Equilibrium Optimizer                          & EO.1  & 2020 & \cite{Faramarzi2020} \\
			Earthworm Optimization Algorithm           & EOA  & 2018 & \cite{wang2018earthworm} \\
			Electimize Optimization Algorithm          & EOA.1    & 2011 & \cite{EOA1} \\
			Emperor Penguins Colony                    & EPC  & 2019 & \cite{penguinemperor} \\   
			Evolution Strategies					   & ES    & 2002 & \cite{Beyer2002} \\  
			Egyptian Vulture Optimization Algorithm    & EV    & 2013 & \cite{EV}   \\
			Frog Call Inspired Algorithm 	           & FCA   & 2009 & \cite{Mutazono5227977} \\
                Forest Optimization Algorithm              & FOA.1 & 2014 & \cite{FOA1} \\  				
                FOX-inspired Optimization Algorithm        & FOX   & 2023 & \cite{Mohammed2023fox} \\
			Genetic Algorithms                         & GA    & 1989 & \cite{GA}   \\
			Golden Ball Algorithm						   & GBA     & 2014 & \cite{Osaba2014} \\
			Galaxy Based Search Algorithm            & GBSA     & 2011 & \cite{GBSA}                  \\
			Gene Expression                            & GE    & 2001 & \cite{GE}   \\  
			Group Leaders Optimization Algorithm 	   & GLOA  & 2011 & \cite{Daskin2011761} \\
                Goat Search Algorithms                       & GSA.2 & 2022 & \cite{De2022} \\
                Golf Sport Inspired Search                     & GSIS    & 2024 & \cite{Husseinzadeh2024} \\
			Honey-Bees Mating Optimization Algorithm             & HBMO & 2006 & \cite{BOZORG} \\
			Hyper-parameter Dialectic Search               & HDS   & 2020 & \cite{hyperparamdialectic} \\
			Harmony Elements Algorithm         		   & HEA   & 2009 & \cite{cui2010lambda} \\		
			Human Evolutionary Model                   & HEM   & 2007 & \cite{HEM}    \\ 			
			Human-Inspired Algorithms                  & HIA   & 2009 & \cite{HIA}    \\			
			Hysteresis for Optimization        			 & HO     & 2002 & \cite{Zarand2002150201}      \\			
			Harmony Search                             & HS    & 2005 & \cite{HS}    \\
			Hypercube Natural Aggregation Algorithm				 & HYNAA & 2019 & \cite{Maciel2019} \\
			Japanese Tree Frogs Calling Algorithm                & JTFCA & 2012 & \cite{hernandez2012distributed} \\			
     		Immune-Inspired Computational Intelligence & ICI  & 2008 & \cite{Cortes} \\
            Improved Genetic Immune Algorithm		   & IGIA & 2017 & \cite{tayeb2017research} \\
			Improved Raven Roosting Algorithm					 & IRRO & 2018 & \cite{TORABI2018144} \\
			Invasive Tumor Optimization Algorithm      & ITGO  & 2015 & \cite{ITGO} \\					
			Weed Colonization Optimization             & IWO   & 2006 & \cite{IWO}  \\  
			Keshtel Algorithm                          & KA    & 2014 & \cite{KA}   \\ 	
			Kaizen Programming                             & KP    & 2014 & \cite{KP}   \\ 
			Lion Algorithm                             & LA    & 2012 & \cite{LA}   \\  
			Laying Chicken Algorithm 		 & LCA  & 2017 & \cite{hosseini2017laying} \\
                Liver Cancer Algorithm                         & LCA.2   & 2023 & \cite{Houssein2023} \\
			Lion Pride Optimizer                                 & LPO & 2012 & \cite{LPO} \\
			Light Ray Optimization             			& LRO      & 2010 & \cite{Shen2010154}           \\
			Migrating Birds Optimization               & MBO.2  & 2012 & \cite{MBO1}  \\			
                Migration-Crossover Algorithm               & MCA & 2024 & \cite{Kusuma2024migration} \\
			Mosquito Flying Optimization			  & MFO.1 & 2016 & \cite{alauddin2016mosquito} \\
			Marriage In Honey Bees Optimization        & MHBO  & 2001 & \cite{MHBO}  \\ 
			Method of Musical Composition      		   & MMC   & 2014 & \cite{Mora-Gutierrez2014301} \\		Mycorrhiza Optimization Algorithm          & MOA   & 2023 & \cite{Valdez2023} \\
			Mox Optimization Algorithm							 & MOX  & 2011 & \cite{MINHAS20114614} \\
			Melody Search 					   		   & MS.1  & 2011 & \cite{Ashrafi2011109} \\
			Natural Aggregation Algorithm 			   & NAA   & 2016 & \cite{Luo201694} \\		
			\bottomrule
		\end{tabular}
	\end{center}
\end{table}

\begin{table}[htp]
	\begin{center}
    \caption{Nature- and bio-inspired meta-heuristics within the \textit{Solution Creation - Combination} category (III).}
    \label{fig:solcreationComb3}
		\begin{tabular}{ l  l  l r}
			\toprule
			\multicolumn{4}{c}{{\Large \textbf{Creation-Combination category (III)}}}\\
			\midrule
			Algorithm Name         &    Acronym & Year & Reference\\
			\midrule
			Natural Forest Regeneration Algorithm		   & NFR     & 2016 & \cite{Moez2016} \\
			Old Bachelor Acceptance                        & OBA     & 1995 & \cite{OBA}    \\   
   			Photosynthetic Algorithm           		   & PA    & 1999 & \cite{Murase2000115}\\			
			Pity Beetle Algorithm 					 & PBA  & 2018 & \cite{Kallioras2018147} \\
			Polar Bear Optimization Algorithm          & PBOA & 2017 & \cite{PBOA} \\
			Particle Collision Algorithm       		   & PCA   & 2007 & \cite{Sacco2007} \\		
			Plant Growth Optimization                      & PGO     & 2008 & \cite{PGOplants} \\
                Plasma Generation Optimization                 & PGO.1 & 2020 & \cite{Kaveh2020} \\
			Pearl Hunting Algorithm						   & PHA   & 2012 & \cite{chan2012hyper} \\
			PopMusic Algorithm                 			& PopMusic & 2002 & \cite{Taillard2002613}       \\			
			Queen-Bee Evolution                        & QBE   & 2003 & \cite{QBE}  \\ 			
			Quantum Superposition Algorithm		     & QSA & 2015 & \cite{saire2015approach} \\
			Red Deer Algorithm									 & RDA   & 2016 & \cite{fard2016red} \\   
			Reactive Dialectic Search                      & RDS   & 2017 & \cite{reactivedialectic} \\			
			Rain-Fall Optimization Algorithm        & RFOA & 2017 & \cite{AGHAYKABOLI201731} \\
			Rhino Herd Behavior								 & RHB   & 2018 & \cite{wang2018novel} \\
                Rock Hyraxes Swarm Optimization               & RHSO  & 2021 & \cite{Belal2021} \\
			Raccoon Optimization Algorithm                        & ROA & 2018 & \cite{raccoonoptimization} \\			
			Reincarnation Concept Optimization Algorithm& ROA.1  & 2010 & \cite{ROA}   \\ 
			Ringed Seal Search                                   & RSS  & 2015 & \cite{ringedseal} \\				
			Shark Search Algorithm                     & SA    & 1998 & \cite{SA2}   \\
			Simulated Annealing                    & SA.1    & 1989 & \cite{SA}               \\	 		
			Scientifics Algorithms                     & SA.2  & 2014 & \cite{SA1}  \\ 							
			SuperBug Algorithm						   & SuA   & 2012 & \cite{anandaraman2012new} \\
			Simulated Bee Colony 					   & SBC   & 2009 & \cite{McCaffrey5211598} \\			
			Snap-Drift Cuckoo Search                   & SDCS  & 2016 & \cite{SDCS} \\
			Self-Defense Mechanism Of The Plants Algorithm & SDMA & 2018 & \cite{SDMA} \\ 			
			Social Engineering Optimization				   & SEO   & 2017 & \cite{fardsocial} \\
			Sheep Flock Heredity Model                 & SFHM  & 2001 & \cite{Nara1999}  \\ 							
			Shuffled Frog-Leaping Algorithm            & SFLA  & 2006 & \cite{SFLA} \\
                Snow Flake Optimization Algorithm              & SFO.1 & 2023 & \cite{Toz2023re} \\
			Saplings Growing Up Algorithm			   & SGA.1 & 2007 & \cite{Karci2007450} \\			
			Search Group Algorithm                     & SGA.2 & 2015 & \cite{SGA}  \\			
			Swine Influenza Models Based Optimization  & SIMBO & 2013 & \cite{PATTNAIK2013628} \\
			Sonar Inspired Optimization				& SIO & 2017 & \cite{tzanetos2017new} \\
            Seasons Optimization                           & SO.1    & 2022 & \cite{Emami2022} \\
			Self-Organizing Migrating Algorithm        & SOMA  & 2004 & \cite{Zelinka2004} \\		
			Simple Optimization 					   & SOPT  & 2012 & \cite{hasanccebi2012efficient} \\ 			
			Strawberry Plant Algorithm                     & SPA     & 2014 & \cite{Strawberryplant} \\		Smart Root Search                              & SRS     & 2020 & \cite{Naseri2020} \\
			Salp Swarm Algorithm								 & SSA.2 & 2017 & \cite{MIRJALILI2017163} \\
            Sling-shot Spider Optimization                       & S$^2$SO & 2023 & \cite{Adhi2023} \\
			Tree Growth Algorithm						   & TGA.1     & 2019 & \cite{CHERAGHALIPOUR2018393} \\
      		The Great Deluge Algorithm 					   & TGD   & 1993 & \cite{Dueck199386} \\ 
			Small World Optimization                       & SWO   & 2006 & \cite{SWO1} \\ 
			Virulence Optimization Algorithm		   & VOA  & 2016 & \cite{JADERYAN2016596} \\
			Virus Optimization Algorithm 			   & VOA.1   & 2009 & \cite{juarez2009virus} \\					
			Viral Systems Optimization                 & VSO   & 2008 & \cite{VSO}  \\ 					
\bottomrule
\end{tabular}
\end{center}
\end{table}

\begin{table}[htp]
	\begin{center}
    \caption{Nature- and bio-inspired meta-heuristics within the \textit{Solution Creation - Combination} category (IV).}
    \label{fig:solcreationComb4}
		\begin{tabular}{ l  l  l r}
			\toprule
			\multicolumn{4}{c}{{\Large \textbf{Creation-Combination category (IV)}}}\\
			\midrule
			Algorithm Name         &    Acronym & Year & Reference\\
			\midrule		
			Wasp Colonies Algorithm                     & WCA   & 1991 & \cite{WCA}   \\
			Water Flow-Like Algorithms         			& WFA      & 2007 & \cite{Yang2007475}           \\
			Water Flow Algorithm               		   & WFA.1 & 2007 & \cite{Basu20071825} \\		
			Wasp Swarm Optimization                    & WSO   & 2005 & \cite{WSO}   \\ 				
			Ying-Yang Pair Optimization                & YYOP  & 2016 & \cite{YYOP} \\			
\bottomrule
\end{tabular}
\end{center}
\end{table}
\subsubsection{Creation by Stigmergy}

Another popular option of creating new solutions relies on stigmergy, namely, an indirect communication and coordination between the different solutions or agents used to create new solutions. This communication is usually done using an intermediate structure, with information obtained from the different solutions, used to generate new solutions oriented towards more promising areas of the search space. This is indeed the search mechanism used in the most representative algorithm of this category, ACO \cite{ACOBook}, which is inspired by the foraging mechanism of ant colonies. Each ant of the colony describes a trajectory over a graph representation of the search space of the problem at hand, and leaves a trace of pheromone along its way whose intensity depends, in part, on the fitness value corresponding to the solution encoded by the trajectory of the ant. In subsequent iterations, new solutions are generated, dimension by dimension, considering the pheromones trail left by preceding ants, enforcing the search around the most promising values for each dimension.

Table \ref{fig:solcreationStimergy} lists the reviewed algorithms that employ stigmergy when creating new solutions. This is a reduced list when comparing with preceding categories, with the majority of the algorithms relying on Swarm Intelligence among insects (similarly to ACO). However, other algorithms inspired in physics have also a stigmertic behavior when producing new solutions, such as methods inspired by water flow dynamics \cite{WFO} and the natural formation of rivers \cite{RFD}. 
\begin{table}[htp]
	\begin{center}
		\caption{Nature- and bio-inspired meta-heuristics within the \emph{Solution Creation - Stigmergy} category.}
		\label{fig:solcreationStimergy}
		\begin{tabular}{ l  l  l r}
			\toprule
			\multicolumn{4}{c}{{\Large \textbf{Solution Creation - Stigmergy}}}\\
			\midrule
			Algorithm Name         &    Acronym & Year & Reference\\
			\midrule
			Ant Colony Optimization                      & ACO   & 1996 & \cite{ACO}  \\ 
			Bee System                                   & BS.1  & 2002 & \cite{BS}   \\ 
			Hammerhead Shark Optimization Algorithm      & HSOA & 2019 & \cite{HSOA} \\
			Intelligence Water Drops Algorithm           & IWD  & 2009 & \cite{IWD}   \\ 	
			River Formation Dynamics                     & RFD  & 2007 & \cite{RFD}   \\ 				
			Termite Hill Algorithm 						 & TA & 2012 & \cite{Zungeru20121901} \\			
			Virtual Ants Algorithm                       & VAA  & 2006 & \cite{VAA}   \\ 								
			Virtual Bees Algorithm                     	 & VBA & 2005 & \cite{VBA}   \\ 						
			Water-Flow Algorithm Optimization            & WFO  & 2011 & \cite{WFO}   \\								
			\bottomrule
		\end{tabular}
	\end{center}
\end{table}

\section{Taxonomies Analysis: Comparison and More Influential Algorithms} \label{sec:analysis-survey}

We now proceed by critically examining the reviewed literature as per the different taxonomies proposed in this overview. First, we are going to study the similarities between the results of the classifications following each taxonomy. Later, we identify the most influential algorithms over the rest, based on the behavior of the algorithms.

\subsection{Comparison Between both Taxonomies} \label{sec:double_tax}

Comparing the two taxonomies to each other and the algorithms falling into each of their categories, it can be observed that there is not a strong relationship between them. Interestingly, this unveils that features characterizing one algorithm are loosely associated with its inspirational model. For instance, algorithms inspired by very different concepts such as the gravitational forces (GFA, \cite{Zheng2010}) or animal evolution (ABO, \cite{ABO}) exhibit a significant similarity with PSO \cite{PSO}. This statement is supported by the fact that, in the second taxonomy, each category is composed by algorithms that, as per the first taxonomy, are inspired by diverse phenomena. The contrary also holds in general: proposals with very similar natural inspiration fall in the same category of the first taxonomy (as expected), but their search procedures differ significantly from each other, thereby being classified in different categories of the second taxonomy. An illustrative example is the Delphi Echolocation algorithm (DE, \cite{DE1}) and the Dolphin Partner Optimization \cite{DPO}. Both are inspired by the same animal (dolphin) and its mechanism to detect fishes (echolocation), but they are very different algorithms: the former creates new solutions by combination, whereas the latter resembles closely the movement performed in the PSO solver, mainly guided by the best solution.

In this same line of reasoning, the largest subcategory of the second taxonomy (Differential Vector Movements guided by representative solutions) not only contains more than half of the reviewed algorithms (almost 60\%), but it also comprises algorithms from all the different categories in the first taxonomy: Social Human Behavior (as Anarchic Society Optimization, ASO, \cite{ASOBueno}), microorganisms (Bacterial Colony Optimization, \cite{Niu2012501}), Physics/Chemistry category (correspondingly, Fireworks Algorithm Optimization, FAO, \cite{FAO}), Breeding-based Evolution (as Variable Mesh Optimization, VMO \cite{VMO}), or even Plants-Based (such as Flower Pollination Algorithm, FPA \cite{FPA}). This inspirational diversity is not exclusive to this subcategory. Others, such as Solution Creation, also include algorithms relying on the heterogeneity of natural concepts.

Considering the previous examples, it is clear that the real behavior of the algorithm is much more informative than its natural or biological inspiration. Even more, we have observed that in our first proposed taxonomy, built upon the review of 518 proposals, the huge diversity of inspirational sources does not correspond with the lower number of algorithmic behaviors on which our second taxonomy is based. This observation is in accordance with previous works in the literature, which have put to question whether the novelty in the natural inspiration of the algorithm actually yields different algorithms that could produce competitive results \cite{metaphor,review11}.

We further elaborate on the above statement: our literature analysis revealed that the majority of proposals (more than a half, 60\%) generate new solutions based on differential vector forces over existing ones, as in the classical PSO or DE. A complementary analysis can be done by departing from this observation towards discriminating which of the classical algorithms (PSO, DE, GA, ACO, ABC or SA) can be declared to be most similar to modern approaches. The results of this analysis are conclusive: 23\% of all reviewed algorithms (122 out of 518) were found to be so influenced by classical algorithms that, without their biological inspiration, they could be regarded as incremental variants. The other 396 solvers (about 77\%) have enough differences to be considered a new proposal by themselves, instead of another version of existing classical algorithms. But, we must emphasize that in these new algorithms there exists a lack of originality or justification in a significant percentage of cases. We must emphasize that in these new algorithms there exists a lack of justification due to the lack of comparison with the state of the art and the lack of real interest in achieving reasonable levels of quality from the perspective of the optimization of well-known problems in recent competitions.

\begin{table}[htp]
	\begin{center}
		\caption{Percentages of similar algorithms in the reviewed literature.}
		\label{tbl:analysis}		
		\begin{tabular}{lrr}
      \toprule
			Classical algorithm & Number of papers with similar algorithms & Percentage over the total \\
      \midrule
      PSO & 57 & 11.00\% \\
      DE & 24 & 4.63\% \\
      GA & 24 & 4.63\% \\
      ACO & 7 & 1.35\% \\
      ABC & 7 & 1.35\% \\
      SA &  3 & 0.59\% \\
      \midrule
      Total  & 122 & 23.55\% \\
      \bottomrule
		\end{tabular}
	\end{center}
\end{table}

\subsection{Identification of the Most Influential Algorithms} \label{sec:influential}

\begin{figure}[htp]
\centering
	\includegraphics[width=.9\textwidth]{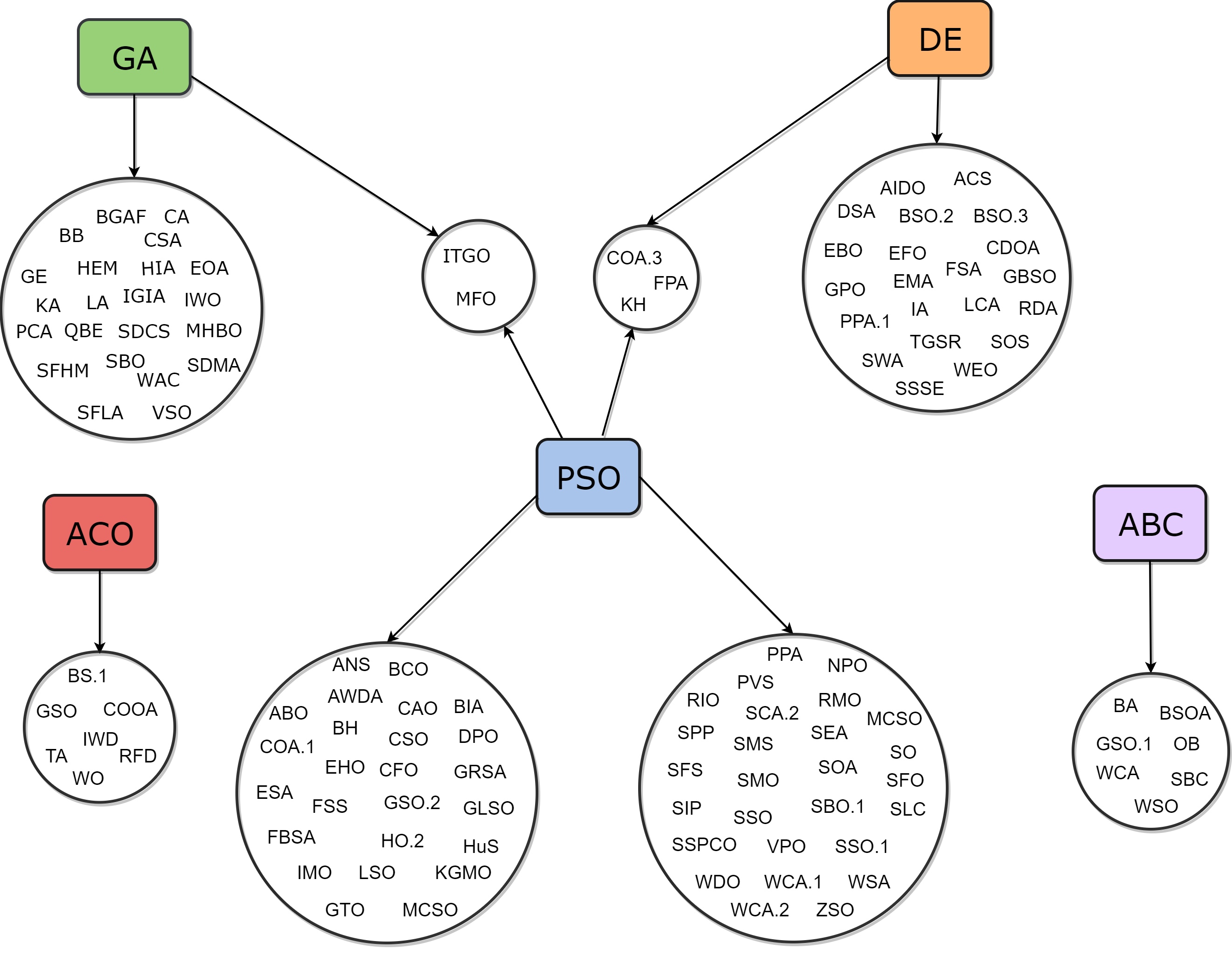}
	\label{fig:total}
	\caption{Classification of proposals by its original algorithm.} 
\end{figure}

In order to know which are the most influential reference algorithms used to design other bio-inspired algorithms, we have grouped together reviewed proposals that could be considered to be versions of the same classical algorithm. Figure \ref{fig:total} shows the classification of each algorithm based on its behavior, and the number of proposals in each classification are summarized in Table \ref{tbl:analysis}.

Very insightful conclusions can be drawn from this grouping. To begin with, in Table \ref{tbl:analysis} the most influential algorithm was identified to be PSO, appearing in 11\% of the reviewed literature (which corresponds to almost 47\% of the proposals that were clearly based on a previous algorithm). This bio-inspired solver is one of the most prominent and historically acknowledged algorithms in the Swarm Intelligence category and is the reference of many bio-inspired algorithms contributed since its inception. The simplicity of this algorithm and its ability to reach an optimum quickly -- as has been comparatively assessed in many application scenarios, see e.g. \cite{chouikhi2019bi,chouikhi2017pso} -- have inspired many authors to create new metaheuristics characterized by similar solution movement dynamics to those defined by PSO. Thus, many algorithms whose authors claim to simulate the behavior of a biological system eventually perform their search process through movements strongly influenced by PSO (in some cases, without any significant difference). 

The second and third most influential algorithms are GA, a very classic algorithm, and DE, a well-known algorithm whose natural inspiration resides only in the evolution of a population. Both have been used by around 5\% of all reviewed nature-inspired algorithms, and they are the most representative approach in the \textit{Evolutionary Algorithms} category. The search mechanism of GA is solution creation by combination, and the search mechanism of DE is to create new solutions with a linear combination of existing ones in the population, which is used by 5\% of all reviewed proposals, maybe by its superior performance reported for many optimization problems \cite{Molina2018Comp}. 

When inspecting the influential approaches from a higher perspective, two are the categories whose algorithms have been more frequently used to create new nature-based algorithms. The first one is \textit{Swarm Intelligence}: about 14\% of all studied nature-inspired algorithms are variations of SI algorithms (PSO, ACO, and ABC). The second one is shared between \textit{Evolutionary Algorithms} and GA, whose represented algorithms are both used in almost 5\% of the reviewed cases. It is noteworthy to highlight that it appears that the influence of more classic algorithms like GA and SA is declining when compared to other algorithms, such as DE and PSO. 

In summary, although in the last years many nature-inspired algorithms have been proposed by the community and their number grows steadily every year, more than half of the proposals reviewed in our work are incremental, minor versions of only three very classical algorithms (PSO, DE, and GA). We, therefore, conclude that a huge number of natural and biological sources of inspiration used so far to justify the design of new optimization solvers have not led to significantly disruptive algorithmic behaviors. This closing note will be at the heart of our critical analysis exposed in the next section.

\section{Learned Lessons and Recommendations from the Analysis of the Evolution of Bio-Inspired Optimization}\label{sec:lessons}

After reviewing the algorithms and both taxonomies, we have identified several key learned lessons which serve as recommendations to deal with in forthcoming years for that is working on nature- and bio-inspired optimization. The learned lessons gained from the taxonomies and research outlined in \cite{Molina2020comprehensive} form the foundation of this section. In the subsequent sections, we will further expand upon them to provide a more extended analysis. We next outline them in no particular order:

\begin{enumerate}[leftmargin=*]
\item \textbf{The behavior is more relevant than the natural inspiration:} As was exposed in Section \ref{sec:analysis-survey}, the current literature is flooded with a huge number of nature- and bio-inspired algorithms. However, as has been spotted by our proposed taxonomies, several algorithms belonging to categories with different sources of inspiration results are very similar in terms of behavior. This disparity is a controversial topic in recent years \cite{DELSER2019220,metaphor}. Therefore, we call for more research efforts around the design of optimization algorithms that focus on their behavior and properties (e.g., good performance, simplicity, ability to run it in parallel or their suitability to a specific type of problems) rather than on new inspiration sources.

\item \textbf{Nature-based terminology can make it more difficult to understand the proposal:} A great deal of papers presenting new bio-inspired solvers are difficult to understand and replicate due to the extended usage of vocabulary related to the natural source of inspiration. It is logical to use the semantic of the biological or natural domain, but to an extent. It would be desirable that the description of the algorithm could be defined in an inspiration-agnostic fashion, resorting to mathematical terms to describe each component, agent and/or phase of the optimization process (e.g. optimum/a, individuals, or solutions). Excessive usage of the domain terminology (without explicitly indicating the correspondences) could make it difficult to follow the details of the algorithm for researchers not acquainted with such a terminology. To overcome this issue, the correspondence between the domain terminology and the optimization terminology should be explicitly indicated.
  
\item \textbf{Good comparisons are crucial for new proposals:} The lack of fair comparisons is another important drawback of many proposals published to date. When new algorithms are proposed, unfortunately, many of them are only compared to very basic and classical algorithms (such as GA or PSO). These algorithms have been widely surpassed by more advanced versions over the years which, so obtaining better performance than naive version of classical algorithms is relatively easy to achieve, and it does not imply a competitive performance \cite{review11}. In some cases, the proposed algorithm is compared to similar algorithms but not with competitive algorithms outside that semantic niche \cite{review11,PerfTuned}. This methodological practice must be regarded as a very serious barrier for their application to real-world problems. We encourage researchers to increase the algorithms used in their experimental section, including more competitive or state-of-the-art algorithms: until they are proven to be competitive in respect to the state of the art, new nature- and bio-inspired solvers will not be used in practice either will attract enough attention of the research community.

\item \textbf{Many proposals have a very limited influence:} By examining in depth the historical trajectory followed by each reviewed algorithm, an intriguing trend is revealed: a fraction of the proposals have a very limited influence in new papers after the original publication. For them, there are almost no new papers with improved versions, or applying it to new problems. Fortunately, other algorithms have a stronger influence. In view of this dichotomy, the researchers should evaluate their proposals for diverse problems, including widely acknowledged benchmark functions and real-world practical problems, to grasp the interest of the community in considering their proposed algorithms for tackling other applications.

\item \textbf{The interest of making source code available:} Related to the previous one, it is very interesting, in order to gain more visibility, to make the source code of the proposed algorithm available for the community. It is true that the paper presenting the new algorithm should be detailed enough to allow for a clean implementation of the proposal from the provided specification. However, it is widely acknowledged that, in many occasions, there are important details that even though they have a strong influence on the results, are not remarked in the description \cite{Bosman2018,Biedrzycki2019247}. A publicly available reference implementation could not only improve its visibility, but could also offer other researchers the chance to undertake more thorough performance comparisons. In addition, there are a huge number of software frameworks for Evolutionary Computation and Swarm Intelligence programmed in different languages (such as C++, Java, Matlab, or Python), some of them very popular in the current research landscape. To cite a few: Evolutionary Computation Framework (ECF)\footnote{\url{http://ecf.zemris.fer.hr/}} and ParadisEO \cite{ParadisEO} in C++; jMetal \cite{jMetal} and MOEA \footnote{\url{http://moeaframework.org/}} in Java; NiaPy\cite{NiaPyJOSS2018}, jMetalPy \cite{BENITEZHIDALGO2019100598} and PyGMO\footnote{\url{http://esa.github.io/pygmo/index.html}} in Python; or PlatEMO \cite{PlatEMO} in Matlab, among others. Each of them implements the most popular algorithms (GA, DE, PSO, ABC, ...). A reference implementation could also favor the inclusion of the proposal in frameworks as the ones exemplified previously. Otherwise, different implementations of the allegedly same algorithm could produce diverging results from the original proposal (in part due to the ambiguity of the description).

\item \textbf{The role of bio-inspired algorithms in competitions:} Finally, we also stress on the fact that metaheuristic algorithms that have scored best in many competitions are far from being biologically inspired, although some of them retain their nature-inspired roots (mostly, DE) \cite{Molina2018Comp}. This fact was expected for the lack of good methodological practices when comparing nature- and bio-inspired algorithms, which was pointed out previously in our analysis. This issue has not encouraged participants in competitions to embrace them as reference algorithms to design better solvers. The rising trend of the community to generate an ever-growing number of bio-inspired proposals can be counterproductive and deviate efforts towards the development of a reduced number of proposals but with a better performance. 
\end{enumerate}

\section{A Short Reflection on The \textit{Good}, the \textit{Bad} and the \textit{Ugly}} \label{sec:criticalanalysis}

This section corresponds to the integration and extension of Section 3 of the article published in \cite{Molina2022nature} within this report. In Section \ref{sec:good}, we extend the original analysis on the importance of applications, stressing on the numerous applications that leverage results from this research area (the \textit{good}). In section \ref{sec:bad}, we have also extended the original content to more studies based on the recent problems of the area, namely, the lack of algorithmic innovation in algorithms inspired by novel metaphors and good comparisons between algorithms (the \textit{bad}). Section \ref{sec:ugly} remains as in the original work, underscoring the poor practices experienced by the area in recent times (the \textit{ugly}).

\subsection{The \textit{Good}: A Present and Future Plenty of Exciting Applications} \label{sec:good}

An undeniable fact is that nature- and bio-inspired optimization algorithms have been applied to a great variety of optimization problems emerging in different disciplines. We distinguish among three different horizons of applications, without being exhaustive, nor entering into the recent horizons of general-purpose AI that we will mention in the conclusions. They are outlined shortly as follows:
\begin{itemize}[leftmargin=*]
    \item \textbf{Real-world engineering applications}: We can find many examples regarding the usage of bio-inspired techniques to solve real-world engineering processes \cite{valadi2014applications}. Furthermore, structural design and civil engineering have also largely embraced the benefits of nature and bio-inspired solvers to assorted problems, including the multi-criteria design of structures \cite{gandomi2013metaheuristic}, logistics and supply chain management \cite{griffis2012metaheuristics}, to cite a few. The application of Evolutionary Algorithms (EAs) has reached many areas, including works from these human competitions for the design of breakwaters \cite{Starodubcev2022}, the evolution of antennas for Space Mission of the NASA \cite{Lohn2005evolved}, and also the discovery of new formulas in the field of physics \cite{Schmit2009}, among many other important applications.
    \item \textbf{Academic competitions}: From the research perspective, several worldwide competitions have developed over the years to test new proposals in an unbiased and replicable way. In such competitions, DE has created a great impact as the core meta-heuristic algorithm of winning competitors in the global optimization competitions held in renowned conferences (GECCO and CEC) over the last decade \cite{Molina2018Comp}. The family of EAs has attracted the interest of researchers by participating in genetic and evolutionary competitions with prizes \footnote{\url{https://human-competitive.org/}} (Annual "Humies" Awards For Human-Competitive Results), and others such as GECCO and CEC previously annotated \cite{review11}. 
    \item \textbf{Going deeper into the creation of Machine Learning (ML) and Deep Learning (DL) models}: Although most algorithms have been developed in recent years, the impact of EAs, a classical family of algorithms, has risen in the last few years. Their use in ML has been widely studied both for the design of models \cite{Linan2023} and also as a support for the optimization of those models \cite{Song2019}.  These algorithms have gained momentum under the evidence reported around their usage to evolve and improve other AI techniques: most notably, the optimization of the structure and training parameters of deep neural networks \cite{MARTINEZ2021161}, or the creation of new data-based models from scratch (i.e. by evolving very essential data processing primitives) that has been presented in the groundbreaking work by Google \cite{automlzero}. With this ongoing development, the research trend of Neural Architecture Search has emerged as another important area full of EAs applications \cite{Yuqiao2023}, which mainly focuses on the construction of the DL model via the evolution of block of layers \cite{Dufourq2017,Camacho2020,Assunccao2019}. Recently, we have witnessed the use of EAs to model more AI models, as in the case of POET \cite{Wang2019} where more environments are generated to learn from the diversity created, with the merging of EAs with Large Language Model (LLM) \cite{Akiba2024}, and with other areas such as Automated Machine Learning \cite{He2021}, Reinforcement Learning and robotics \cite{JasonMa2023}, and Multi-task Learning \cite{Zhao2023}. In recent years, an interesting synergy between bio-inspired optimization and modern ML systems has been observed in the literature, in particular General-Purpose Artificial Intelligence Systems (GPAIS), as we will highlight later in the report.   
\end{itemize}

\subsection{The \textit{Bad}: Novel Metaphors Not Leading to Innovative Solvers} \label{sec:bad}

As previously mentioned, an ever-growing amount of new bio-inspired optimization techniques has been proposed in recent decades (see Figure \ref{fig:papers}). This overwhelming number of alternatives could make it difficult to choose an appropriate option for a given optimization problem. The vast number of proposals not only casts doubt on the convenience of choosing one or another algorithm but has also produced solvers that, even if relying on different metaphors, are mathematically too similar to already existing optimization algorithms. In other words, despite the diversity of methods considering their natural inspiration, such diversity does not hold as far as mathematical differences are concerned, as exposed by recent studies \cite{PIOTROWSKI2014191}. As we have mentioned in the introduction, this metaphor-driven research trend has been already denounced in several contributions \cite{review8,fister2016new}, which have unleashed hot debates around specific meta-heuristic schemes that remain unresolved to date \cite{weyland2015critical,saka2016metaheuristics}, and with a growing problem when important challenges are not addressed and if more and more biological inspirations are maintained as we can observe in 2024 with more than 500 proposals.

Particular reasons aside, some algorithms are not created to solve problems and provide a practical advantage, but mainly to be published and gain notoriety without any consideration for their lack of algorithmic novelty and innovation. Examples of this controversy can be found in \cite{Camacho2020}, as authors state this problem even in the title of the work. In the previous work, authors ``provide compelling evidence that the grey wolf, the firefly, and the bat algorithms are not novel, but a reiteration of ideas introduced first for particle swarm optimization and reintroduced years later using new natural metaphors''. Then, they rewrite these highly cited papers in terms of PSO, and conclude that ``they create confusion because they hide their strong similarities with existing PSO algorithms ... these three algorithms are unnecessary since they do not add anything new to the tools that can be used to tackle optimization problems''. 

More and more works lack variety in the field, as it was discussed in \cite{Tzanetos2021} (``Nature inspired optimization algorithms or simply variations of metaheuristics?''), authors discussed several matters listed as follows:
\begin{itemize}[leftmargin=*]
    \item \textbf{Does the physical analogue exist?}: The inspiration of several bio-inspired algorithms does not strictly follow the rules of a phenomenon. An example is Cat Swarm Optimization, in which cats form a swarm, but in real life, they do not seem to cooperate in any way. Authors show more examples (Coyote Optimization Algorithm, Dolphin Swarm Optimization Algorithm, among others), and claim that `` a significant number of these algorithms are very similar to other already existing ones''.
    \item \textbf{Similar inspiration or duplicate methods?}: Authors analyze several classes of bio-inspired algorithms such as those based on gravitational forces, water phenomena, bees, penguins, wolves, and bacteria, and conclude that not all the different variations are real contributions.
    \item \textbf{Do authors propose multiple techniques based on the same idea?}: Authors discuss the fact that `` several cases can be found where the same authors propose multiple algorithms, which are based on the same nature-inspired idea.'' They show various examples in which a research group has almost a dozen ``novel'' algorithms, with the same researchers at the front. Also, a relevant group of algorithms that are based and attraction and repulsion is full of works under the same researcher's name.
    \item \textbf{When should a new nature-inspired algorithm be introduced?}: The authors analyze the cases in which it is necessary to create novel algorithms. In their words, ``They could be used as global optimizers, while a heuristic algorithm could be added for acting as local search technique for the solutions provided by the nature-inspired method.'' They also annotate the ability of these algorithms as optimizers for Artificial Neural Networks and Support Vector Machines.
\end{itemize}

Due to ``useless metaphors'', ``lack of novelty'' and ``poor experimental validation and comparison'', in \cite{Aranha2022} authors took the decision in this letter to ``call upon all editors-in-chief in the field to adapt their editorial policies'' to reject the publication of \emph{novel} metaphor-based metaheuristics. More than 80 important researchers in the area signed this letter, and accept the publication of novel bio-inspired algorithms if and only if (1) present their method using the normal, standard optimization terminology; (2) show that the new method brings useful and novel concepts to the field; (3) motivate the use of the metaphor on a sound, scientific basis; and (4) present a fair comparison with other state-of-the-art methods using state-of-the-art practices for benchmarking algorithms.

In the following, we shortly describe the critical analysis that has recently been published in several articles that address this problem ``not leading to innovative solvers'':
\begin{itemize}[leftmargin=*]

\item In \cite{Piotrowski2018}, the authors argue that metaheuristics should be simplified by eliminating the unneeded elements as in the case of two winners of the CEC2016 competition, L-SHADE-EpSin and UMOEA-II. The authors conclude that these algorithms ``contain operators that structurally bias their search by favouring sampling from some parts of the decision space'' and ``other metaheuristics should be simplified as they contain unneeded or even harmful operators''. By doing so, metaheuristics will be easier to understand for other researchers. The authors simplify both algorithms by removing operators that are the main cause of structural bias and the experiments when testing against other metaheuristics reveal `` that simplification of some metaheuristics may not only make them more transparent and easier to use, but also improve their performance.'' 

\item In \cite{Camacho2018,Camacho2019intelligent}, the authors analyze the algorithm called Intelligent Water Drops, providing several proofs that `` all main algorithmic components of Intelligent Water Drops are simplifications or special cases of ant colony optimization (ACO)''. They also examine the natural metaphor of  ``water drops flowing in rivers removing the soil from the riverbed”, which is the source of inspiration for this algorithm. Authors conclude that it ``is unnecessary, misleading and based on unconvincing assumptions of river dynamics and soil erosion that lack a real scientific rationale''. 

\item In \cite{Camacho2022}, authors present an analysis of the Cuckoo Search, one of the most well-known algorithms in the literature. Their review of this algorithm based on its usefulness, novelty and sound motivation allow them to ``conclude that neither the metaphor nor the algorithm can be considered as part of the set of useful techniques in stochastic optimization''. The Cuckoo Search is just an evolutionary strategy with some parts of DE, algorithms from the last century. 

\item In \cite{Kudela2022}, authors perform a comparison between seven bio-inspired algorithms with various benchmarks and discovered that ``these (algorithms) contain a centre-bias operator that lets them find optima in the centre of the benchmark set with ease''. The conclusion is that making more ``comparison with other methods (that do not have a centre-bias) is meaningless''. This problem is similar to the appearance of harmful operators, which has already been discussed in \cite{Piotrowski2018}. Authors carry out experimentation with these algorithms against DE and PSO on shifted problems and encounter that ``the worst one performed barely better than a random search'', which is a very serious problem. 

\item  In \cite{Campelo2023}, authors discuss the possible causes of the exponential growth of nature-inspired algorithms and the negative consequences for the field. One cause is the pressure to ``publish or perish,” and authors argue that the ``publishing metaphor-based method is perceived as a low-effort, low-risk process with high potential rewards'' because there are authors that have built professional careers out of creating not one but often multiple metaphor-based methods. The other cause reflected by the authors is ``the lack of a well-established statistical tradition in the field compounds the problem, leading to generally poor practices by authors and, in many cases, an inability of reviewers to pick up on the main methodological problems of some papers''.

\item In \cite{Tzanetos2023}, authors aim to present some nature-inspired methods that contribute to achieving lifelike features of computing systems such as open-ended evolution, intelligence, emergence, resilience, and social awareness. In this work, authors select the algorithms of Big Bang–Big Crunch,  Mine Blast Algorithm, Lightning Search Algorithm, Water Wave Optimization, Gravitational Search Algorithm, Cat Swarm Optimization, Chicken Swarm Optimization, and Roach Infestation Optimization to ``investigate if the mechanisms being part of the algorithms produce qualities found in evolutionary, physical, or chemical analogues.'' The conclusion is that most nature-inspired algorithms ``do not contribute to achieving lifelike features'' and that ``the recent algorithms do not remain accurate to the behavior or the phenomenon on which they are based.''

\item In \cite{Camacho2023b}, the authors claim that grey wolf, moth-flame, whale, firefly, bat, and antlion algorithms are not novel algorithms, and their inspiration has been in the literature for years. To assert this, the authors present a rigorous, component-based analysis of each algorithm that reveals evidences about them: these algorithms are variants of PSO and evolutionary strategies.

\item In \cite{kudela2023evolutionary}, authors discuss the problem of centre-bias. 47 of 90 algorithms that were compared presented a centre-bias. Also, the authors conclude that Harmony Search (HS), Cuckoo Search Algorithm, Firefly Algorithm, Moth Flame Optimization, Ant Lion Optimizer (ALO) should not be used, due to similarities to other algorithms.

\end{itemize}

\subsection{The \textit{Ugly}: Poor Methodological Practices (Questionable Reproducibility and Comparability)} \label{sec:ugly}

An alarming issue that prevails in the area besides the number of metaphor-based proposals is the lack of a fair experimental study to prove their competitiveness when compared to existing solvers. In many research contributions, the newly introduced bio-inspired optimization algorithms are not compared to relevant techniques, but only to classical solvers already surpassed by more recent approaches. Therefore, improving their performance in a benchmark is not a reliable proof of performance competitiveness, but rather a convenient choice of comparison counterparts. Moreover, the experimental design is often not right: for example, the optima of the tested functions is often at the center of the domain search, which favors solvers that focus their search over this region of the solution space. In addition, the statistical significance of the performance gaps reported among algorithms is also frequently overlooked, despite the variability of the results imprinted by the stochastic nature of these algorithms. In this regard, in \cite{kudela2023evolutionary} the same controversy as shown in \cite{Kudela2022} is followed: several algorithms contain a centre-bias operator that makes them more suitable for certain fitness functions. As a result of such bias, these algorithms achieve better results than other algorithms in the appearance of this condition. Thus, they should not be recommended for real-world problems, because the experiments that showed their good performance are biased.

Another important concern in the area is the questionable reproducibility of published studies: the only proof that a proposal is competitive is done experimentally, so it is of utmost importance that results can be reproduced, checked, and verified by third parties, ideally by a different team to that proposing the new algorithm. Unfortunately, in the majority of cases, this is not possible because the implementation of the algorithms is not available, or because important information for the replicability of the experiments is missing or not reported whatsoever \cite{Lopez2021}.

More and more researchers are advocating that a novel metaphor is not enough for a new bio-inspired algorithm to be considered a step beyond the state of the art. Instead, several factors should be proven with empirical evidence, such as superior performance to the state of the art, innovation in the design of its mathematical components and operators, or non-functional benefits that make them more appropriate for real-world optimization problems when compared to other alternatives, e.g. less computational complexity, smaller memory footprint, or faster convergence properties \cite{OSABA2021}.

We strongly urge interested readers to embrace the methodological practices recommended in \cite{LATORRE2021}, considering proposals that have been tested against modern techniques, using standard benchmarks, and with adequate statistical testing to shed light on the relevance of performance gaps. Unfortunately, many recent proposals do not follow these guidelines, remaining as evidence of the ugly side that still prevails in this research area.

\section{Three Propositional Discussions about Nature- and Bio-Inspired Optimization} \label{sec:directions}

As we have mentioned in the introduction, we revisit a triple study of evolutionary and bio-inspired algorithms from a triple perspective, where we stand and what’s next from a perspective published in 2020, but still valid in terms of the need to address important problems and challenges in optimization for EAs and population-based optimization models, a prescription of methodological guidelines for comparing bio-inspired optimization algorithms, and a tutorial on the design, experimentation, and application of metaheuristic algorithms to real-world optimization problems. 

\subsection{Bio-inspired computation: Where we stand and what’s next} \label{sec:hopeful2}

\begin{figure}[H]
\centering
\includegraphics[width=0.7\textwidth]{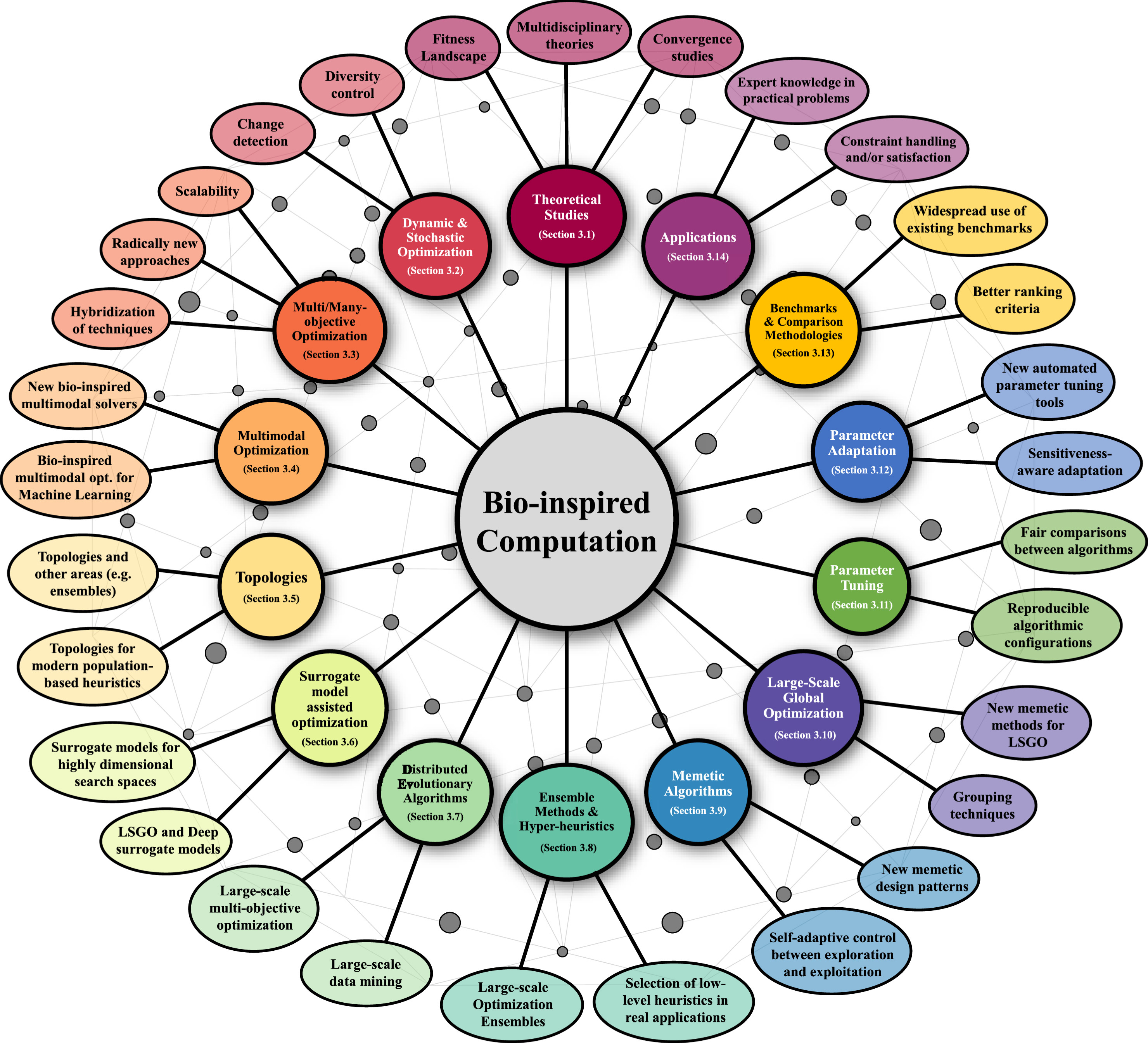}
\caption{\label{fig:tematicas}  Bio-inspired optimization areas and promising research lines. Image taken from \cite{DELSER2019220}.}
\end{figure}

We should pause and reflect on which research directions should be pursued in the future in regard to bio-inspired optimization and related areas, as there are other remarkable fields to be noted as direct applications for bio-inspired optimization. In \cite{DELSER2019220}, the authors show a full discussion of the status of the field from both descriptive (\textit{where we stand}) and prescriptive (\textit{what’s next}) points of view. Here, we describe the areas in which bio-inspired optimization algorithms are used, and research niches related to them, as shown in Figure \ref{fig:tematicas}. The areas and their main aspects that can be studied as promising research lines are:

\begin{itemize}
    \item Theoretical studies: By the hand of the fitness landscape for a better understanding of how a search algorithm can perform on a family of problem instances, of multidisciplinary theories to study the role of diversity and the balance of local search and global search required to undertake a certain problem efficiently, and of convergence, studies to identify the conditions for the convergence of the algorithm, its speed, fitness stability, and other characteristics.
    \item Dynamic and stochastic optimization: These areas need reliable modeling of real optimization scenarios where the characteristics of several of these problems hold (diversity control), and the development of change detection mechanisms relying on characteristics of the optimization algorithm (change detection).
    \item Multi/Many-objective optimization: These areas need new ideas regarding the design of multiobjective solvers because they usually use the main multi-objective solver (radically new approaches) or even the combination of different solvers to create new ones (hybridization of techniques). Another problem to be addressed is the scalability, as solvers do not scale properly with many objectives.
    \item Multimodal optimization: The incorporation of new bio-inspired multimodal solvers and the hybridization of new bio-inspired techniques with traditional strategies can contribute to the progress of this area.
    \item Topologies: A promising research direction is to jointly consider topologies and ensemble strategies to leverage the superior explorative/exploitative powers of ensembles and also topologies for population-based metaheuristics to achieve better solutions than other solvers.
    \item Surrogate model-assisted optimization: This area has promising research lines of investigation with highly dimensional search spaces and DL models, where there is a need to alleviate high computational efforts, with evaluation times that range from hours to days per experiment.
    \item Distributed EAs: These algorithms are needed in large-scale data mining to deal with expensive objective functions, which are common in real-world applications comprising multiple criteria, and also in large-scale multi-objective optimization for the development of asynchronous parallel multi-objective solvers.
    \item Ensemble methods and hyper-heuristics: Both areas have promising challenges in large-scale optimization to address problems such as the encoding strategy, the exploration capabilities of the algorithms, and the computational complexity of the proper ensembles. In real applications, these areas need to address the challenge of the appropriate selection of their low-level composing pieces.
    \item Memetic algorithms: Although these algorithms have shown great results, researchers need to investigate their hybridization with other bio-inspired optimization algorithms for the design of new algorithms, and also the derivation of self-adaptive mechanisms to tune the balance between exploration and exploitation
    \item Large-scale global optimization: For this area,  it will be interesting to develop new techniques that automatically infer relationships among variables (grouping variables) that could be optimized in isolation with the minimum loss of efficiency and to study new approaches of memetic computing with this area, due to the great results of DE and large-scale global optimization.
    \item Parameter tuning: The assignation of proper values to the parameters of bio-inspired algorithms is crucial for obtaining the best possible results for a given problem, so the development of parametric sensitivity analysis, or robustness studies, can be very useful to identify the relevant parameters to tune. Also, when an algorithm is compared to another, it will be necessary to perform a similar parameter-tuning process to make fair comparisons.
    \item Parameter adaptation: Further research about self-adaptation mechanisms for very sensible parameters is necessary, as it reduces the parameters to tune, but can also yield great improvements.
    \item Benchmarks and comparison methodologies: The development of a novel bio-inspired solver includes the comparison to other techniques with several fitness functions. To encourage better comparison methodologies, the most promising avenues are the use of existing benchmarks and also the creation of new ones based on real-world problems. Moreover, better comparison methodologies, including more attention to scalability and new statistical testing approaches such as the use of Bayesian tests, are needed. We delve deeper into this in Subsection \ref{sec:y}.
\end{itemize}

\subsection{Separating the Wheat from the Chaff: Fair and Right Comparisons} \label{sec:y}

One of the problems identified in this manuscript is the abundance of proposals with limited impact. A key aspect for these algorithms to show their strengths is the development of comparative best practices against more competitive algorithms and the state of the art.

To clarify and provide guidelines for a fair and effective comparison between bio-inspired proposals, an extended discussion of the various guidelines to be followed is presented in \cite{LATORRE2021} and here it is summarized as follows:

\begin{enumerate}
    \item \textbf{Benchmarks}: The choice of benchmarking in algorithm evaluation can vary between real-world scenarios and comparisons against existing algorithms. Selecting the right benchmark is crucial, as study conclusions heavily rely on the test environment. However, chosen benchmarks often exhibit biases that can unfairly advantage certain algorithms. Consequently, it is essential to analyze results considering the diverse characteristics of the test problems within the chosen benchmark to ensure fairness in subsequent comparisons.
    \item \textbf{Validation of the results}: Today, simply presenting raw results in extensive tables falls short. Validating results statistically is imperative, complementing tables with proper statistical analyses. It is crucial to not just employ statistical tests, but to ensure they are appropriate for the data at hand. Often, parametric tests are used without verifying if the underlying assumptions are met by the results. Moreover, the use of visualization techniques in comparative analysis is crucial, as these methods condense vast amounts of data into easily comprehensible representations, also aiding quick interpretation for readers.
    \item \textbf{Components analysis and parameter tuning of the proposal}: The hypotheses of the proposal should be explicitly outlined at the paper's outset and revisited upon validation of results. Furthermore, authors ought to undertake a comprehensive analysis of results, addressing key aspects such as search phase identification (balancing exploration and exploitation), component analysis (individually assessing each method component and its complexity), algorithm parameter tuning, and statistical comparison with state-of-the-art algorithms. This thorough examination ensures a robust evaluation of the proposed method and its performance relative to existing approaches.
    \item \textbf{Why is my algorithm useful?}: Prospective contributors must articulate why their proposed algorithm merits attention within the community. Several reasons are proposed to show why a new proposal constitutes an advancement in knowledge, such as its competitiveness against state-of-the-art methods or methodological contributions that foster additional research. This clarity aids in understanding the significance of the proposed algorithm and its potential impact on the field.
\end{enumerate}

\begin{figure}[H]
\centering
\includegraphics[width=0.8\textwidth]{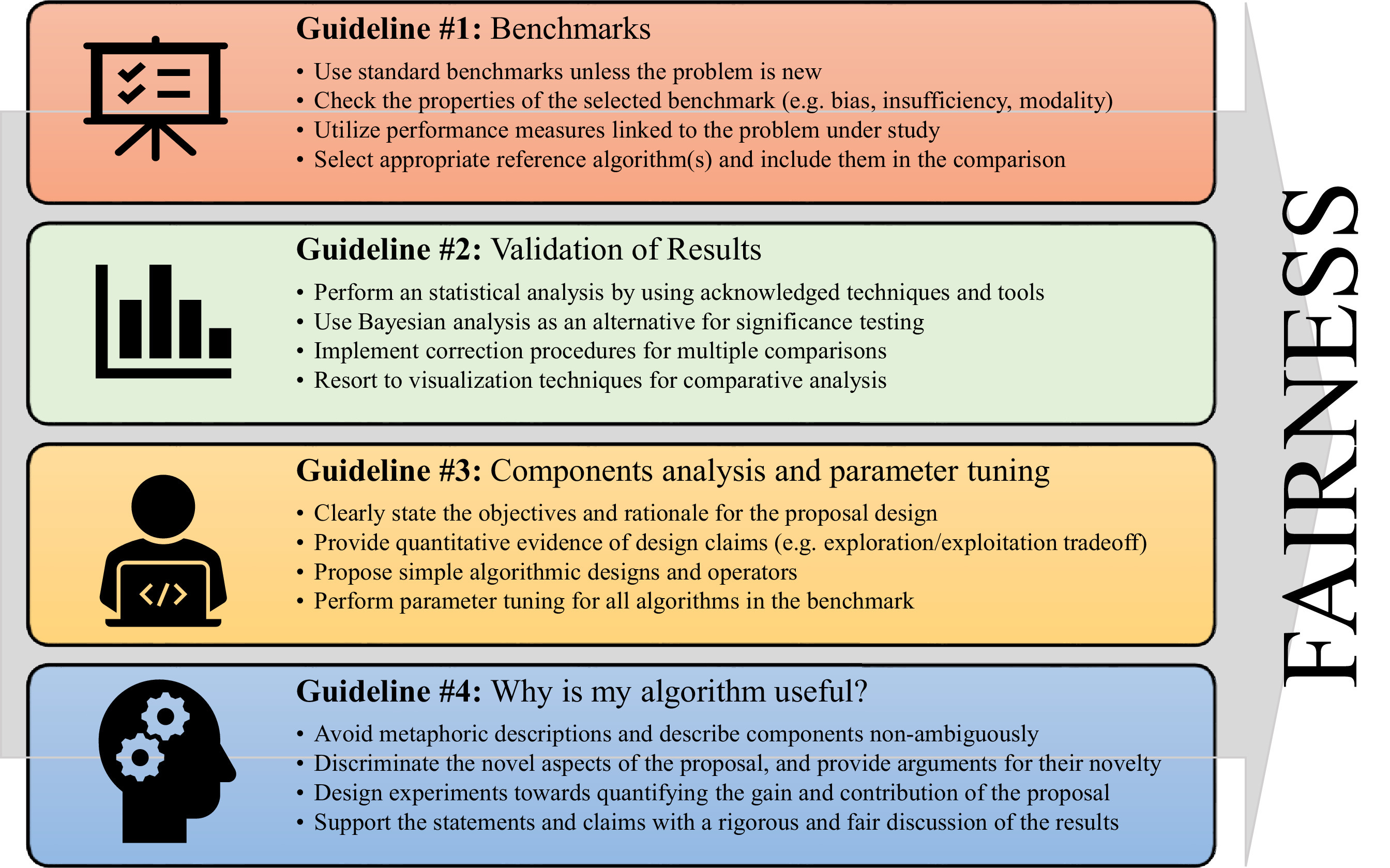}
\caption{\label{fig:guidelines} Summary of the guidelines for goods comparisons between bio-inspired optimization algorithms. Image taken from \cite{LATORRE2021}.}
\end{figure}

These four guidelines form the basis of their work and are discussed in more detail inside the paper published in \cite{LATORRE2021}. Finally, these guidelines are further detailed in Figure \ref{fig:guidelines}.

\subsection{Bridging Theory and Practice in Bio-inspired Optimization: Application to real-world Problems} \label{sec:z}

The correct design of a bio-inspired algorithm involves the execution of a series of steps in a conscientious and organized manner, both at the time of algorithm development and during subsequent experimentation and application to real-world optimization problems. In \cite{OSABA2021}, a complete tutorial on the design of new bio-inspired algorithms is presented, and in this work, we make a brief introduction to the phases that are necessary for quality research.

In such work, an analysis is conducted from a critical yet constructive point of view, aiming to correct misconceptions and bad methodological habits. Each phase of the analysis includes the prescription of application guidelines and recommendations intended for adoption by the community. These guidelines are intended to promote actionable metaheuristics designed and tested in a principled manner, to achieve valuable research results and ensure their practical use in real-world applications.

Other studies have standardized key optimization concepts, though often focusing narrowly on specific phases or domains. However, this tutorial addresses this gap by offering a comprehensive approach, covering all steps from problem modeling to algorithm validation and implementation. This analysis sheds light on different issues to be solved while designing new bio-inspired algorithms and, to prevent this difficulty, a list of the steps to be performed during the creation of the algorithm is presented, ranging from the early phase of problem modeling to the validation of the developed algorithm, as follows:
\begin{itemize}[leftmargin=*]
    \item \textbf{Problem Modeling and Mathematical Formulation}: Leading by a previous conceptualization of the problem, this phase entails the modeling and mathematical formulation of the optimization problem.
    \item \textbf{Algorithmic Design, Solution Encoding and Search Operators}: The goal of this phase is to design and implement the bio-inspired algorithm. In order to do so, we have to avoid the metaphor and align the design of the algorithm according to the constraints of the problem at hand.
    \item \textbf{Performance Assessment, Comparison and Replicability}: Certain aspects of correct evaluation, applicability, and consistency of the research are studied in this phase. It should be based on good practices as he published in \cite{LATORRE2021} which are resumed in the previous subsection.
    \item \textbf{Algorithmic Deployment for Real-World Applications}: This phase is focused on the study of the deployment of the algorithm in a real environment.
\end{itemize}

\begin{figure}[H]
\centering
\includegraphics[width=0.8\textwidth]{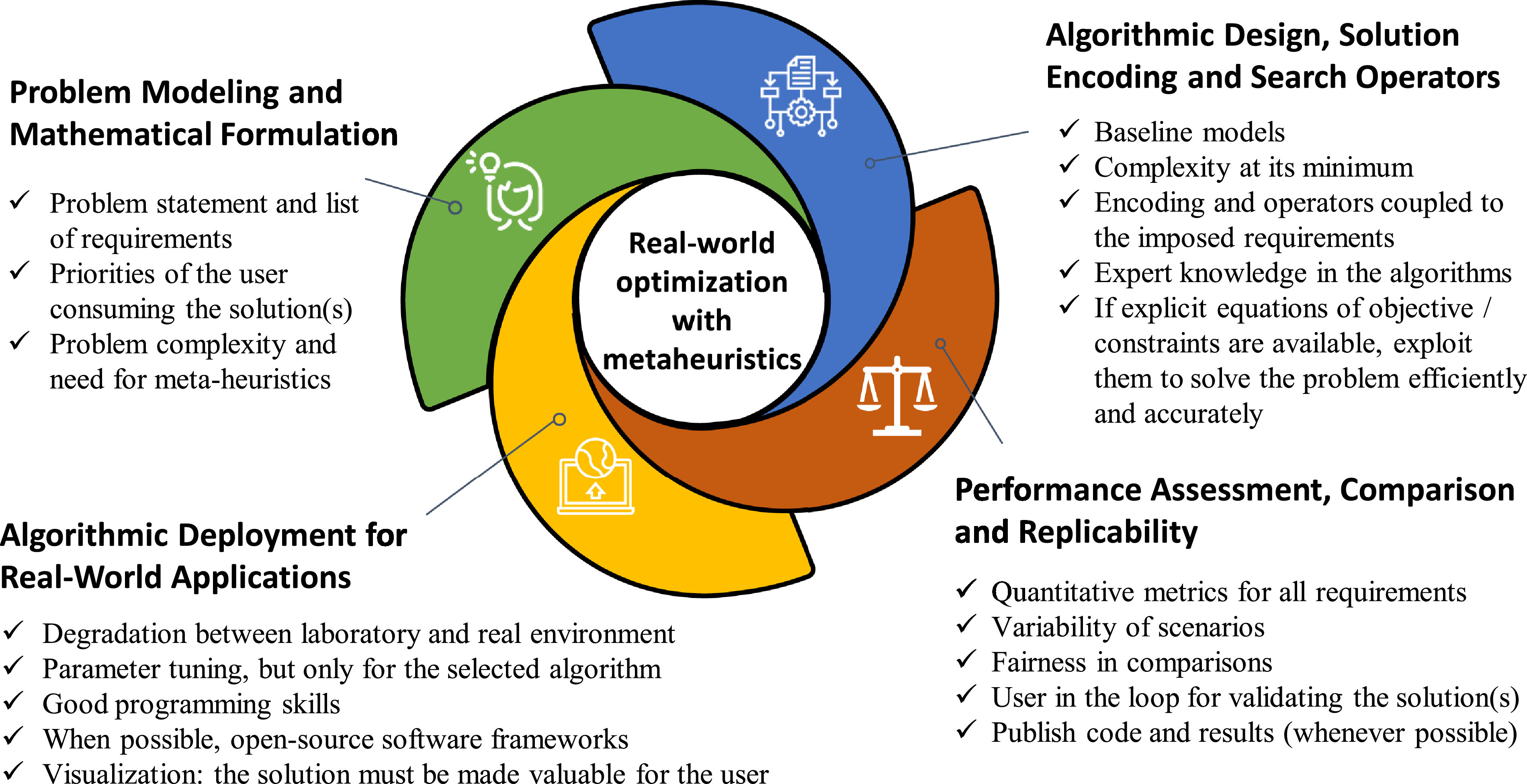}
\caption{\label{fig:tutorial} Summary of the recommendations in the four stages of the proposed methodology for the design of new bio-inspired algorithms. Image taken from \cite{OSABA2021}.}
\end{figure}

These phases are described in depth in the manuscript, but here we show in Figure \ref{fig:tutorial} a summary of the main recommendations for every phase of the proposed methodology for the design of new bio-inspired optimization algorithms.

\section{A Short Recent Literature Analysis: Good Practices, Taxonomies, Overviews, and General Approaches} \label{sec:othertax}

Since the initial version of this paper in 2020, the field of nature and bio-inspired optimization algorithms has continuously evolved. During these last years, the lack of novelty, and bad comparisons, among others, are described as problems that have to be solved to keep the field in progress. As a result, in Subsection \ref{sec:guidelines}, we show several studies and guidelines as good practices for designing metaheuristics. At the same time, researchers continue to work on new ways of classifying metaheuristics and publishing general studies on the topic. That is why in Subsection \ref{sec:otrastax} we consider, without the intention of being exhaustive with all the studies, a summary of almost a dozen of recent studies about taxonomies \cite{review2,Rajwar2023,Sarhani2023}, overviews \cite{Tzanetos2021,Ferrer2023,Sharma2024,Velasco2024,Brahim2024,Marti2024}, and general approaches \cite{Tang2024}. For each paper cited in both subsections, we provide its title, year of publication, and a very short summary of its main contents.

\subsection{Good Practices for Designing Metaheuristics} \label{sec:guidelines}

The constant evolution of the field leads to a significant issue: the lack of novelty in metaheuristics. However, researchers recognize the need to address this problem and have proposed methods to evaluate the novelty of new algorithms. This section shows different studies and guidelines to measure novelty, to design new metaheuristics, and to perform statistical tests between metaheuristics. We list these approaches as follows:
\begin{itemize}[leftmargin=*]
    \item \textbf{On detecting the novelties in metaphor-based algorithms - 2021  \cite{Fister2021}}: This work studies the comparison at the conceptual level using a mathematical formulation based on Markov's chains, and also at the experiment level using the Spearman correlation coefficient between the objective and the population diversity of the algorithms.
    \item \textbf{Similarity in metaheuristics: A gentle step towards a comparison methodology - 2022  \cite{DeArmas2022}}: This paper uses a pool template as a framework for decomposing and analyzing metaheuristics, inspired by another previous work. This template works as a framework for decomposing and analyzing metaheuristics based on these concepts explained in such work: generation method, pool of solutions, archive of solutions, selected pool of solutions, updating mechanism, updated pool, and the archiving and output functions. The authors provide some measures and methodologies to identify their similarities and novelties based on the updating mechanism component, similar to our second taxonomy. They review 15 metaheuristics and their insights confirm that many metaheuristics are special cases of others.
    \item \textbf{Metaheuristics ``In the Large” - 2022  \cite{Swan2022}}: The objective of this work is to provide a useful tool for researchers. To address the lack of novelty, the authors propose a new infrastructure to support the development, analysis, and comparison of new approaches. This framework is based on (1) the use of algorithm templates for reuse without modification, (2) white box problem descriptions that provide generic support for the injection of domain-specific knowledge, and (3) remotely accessible frameworks, components, and problems. This can be considered as a step towards the improvement of the reproducibility of results.
    \item \textbf{Designing new metaheuristics: Manual versus automatic approaches - 2023 \cite{Camacho2023}}: This study discusses two methods for the design of new metaheuristics, manual or automatic. Although authors give credit to the manual design of metaheuristics because this development is based on the designer’s \emph{intuition} and often involves looking for inspiration in other fields of knowledge, which is a positive aspect. However, they remark that this method could involve finding a good algorithm design in a large set of options through trial and error, possibly leading to eliminating designs that, based on their knowledge, they believe would not work for the problem at hand. For this reason, the authors assure the benefits of automatic design, which seeks to reduce human involvement in the design process by harnessing recent advances in automatic algorithm configuration methods. In this work, several automatic configuration methods and metaheuristic software frameworks from the literature are presented and analyzed, some of them already mentioned in section \ref{sec:lessons}, as steps towards better design of metaheuristics.    
    \item \textbf{Research orientation and novelty discriminant for new metaheuristic algorithms - 2024  \cite{Hu2024}}: This work proposes a discriminant method based on a mathematical formulation. It provides the division into root and homologous algorithms so that the former represent strongly innovative proposals due to the novelty of their reproduction operators, and the latter does not show any new combinatorial structure about their reproduction operator. This method shows that Harmony Search, Backtracking Search Optimization Algorithm, Grey Prediction Evolution algorithm, Grey Wolf Optimizer, and Gaining-sharing Knowledge-based algorithm are homologous to several classical algorithms. As a consequence, they develop a research orientation for homologous algorithms to transform the nature metaphor into new structures.
    \item \textbf{Guided learning strategy: A novel update mechanism for metaheuristic algorithms design and improvement - 2024  \cite{Jia2024}}: This work provides guidelines for improving the performance of metaheuristics. The authors have developed a strategy for recalling the algorithm's requirements based on the current population. Authors annotate that this novel mechanism is capable of checking that if the algorithm is biased towards exploration, it will shift towards exploitation in subsequent iterations and vice versa. This strategy obtains the dispersion degree of the population by calculating the standard deviation of the historical locations of individuals in recent generations and infers what guidance the algorithm currently needs. This method has been tested with nearly 60 algorithms, validating its effectiveness in improving performance.
    \item \textbf{A Simple statistical test against origin-biased metaheuristics - 2024  \cite{Walden2024}}: The authors have developed a test to determine algorithm bias. The test is based on the idea that an unbiased algorithm can choose either direction for one of two different local optima in a function. If there is a difference in behavior between independent runs, then the algorithm is likely biased. Algorithms that are biased in terms of the fitness function can lead to undesired behavior. This paper develops and applies a test to known algorithms, including Grey Wolf Optimizer, Whale Optimization, and Harris Hawk, which fail this test. However, algorithms such as DE, GA, and PSO pass the test. This test is a useful tool to solve the centre-bias problem that has already been studied in \cite{kudela2023evolutionary}.
\end{itemize}

\subsection{Latest Metaheuristics based Taxonomies, Overviews, and General Approaches}  \label{sec:otrastax}

This section aims to briefly analyze a summary of almost a dozen other taxonomies, overviews, and global approaches developed during these years. Sorted by year of publication, each work is shortly explained to summarize their messages and contribution to the community:
\begin{itemize}[leftmargin=*]
    \item \textbf{A new taxonomy of global optimization algorithms - 2020  \cite{review2}}: This work analyzes the four characteristic elements of optimization algorithms: how they initialize, generate, and select solutions, how these solutions are evaluated, and lastly, how these algorithms can be parametrized and controlled. By leveraging these elements, a generalized view of optimization algorithms can be created, just by identifying their specific components. The algorithms, according to those specific components, are classified in a taxonomy based on five categories: hill-climbing, trajectory, population-based, surrogate, or hybrid algorithms. Moreover, the study concludes that most algorithms and algorithm classes have a close connection and share similar components, operators, and a large part of their search strategies, which is the basis for the automated design of new algorithms. Lastly, this work provides a guide for algorithm selection, offering best practices for advanced practitioners when choosing optimization algorithms for new problems.
    \item \textbf{Nature inspired optimization algorithms or simply variations of metaheuristics? - 2021  \cite{Tzanetos2021}}: This overview focuses on the study of the frequency of new proposals that are no more than variations of old ones. The authors critique a large set of algorithms based on three criteria: (1) whether there is a physical analogy that follows the metaheuristic, (2) whether most algorithms are duplicates or similarly inspired, and (3) whether the authors propose different techniques based on the same idea. They then specify their criteria for introducing a new metaheuristic.    
    \item \textbf{An exhaustive review of the metaheuristic algorithms for search and optimization: taxonomy, applications, and open challenges - 2023  \cite{Rajwar2023}}: This taxonomy provides a large classification of metaheuristics based on the number of control parameters of the algorithm. In this work, the authors question the novelty of new proposals and discuss the fact that calling an algorithm new is often based on relatively minor modifications to existing methods. They highlight the limitations of metaheuristics, open challenges, and potential future research directions in the field.
    \item \textbf{Metaheuristics in a nutshell - 2023  \cite{Ferrer2023}}: The purpose of this overview is to define the main terms related to the concept of metaheuristic. The text does not provide an extensive taxonomy, but it clearly distinguishes between two classes of metaheuristics: trajectory and population algorithms. It describes the most well-known algorithms for both classes and for single and multi-objective problems. Finally, quality indicators and statistical analysis are explained as good practices. This overview serves as an introduction to the main concepts, algorithms, and methodologies that every metaheuristics researcher should know.
    \item \textbf{Initialization of metaheuristics: comprehensive review, critical analysis, and research directions - 2023  \cite{Sarhani2023}}: This review addresses a gap in the literature by developing a taxonomy of initialization methods for metaheuristics. This classification is based on the initialization of metaheuristics according to random techniques, learning methods (supervised learning, Markov models, opposition- and diversification-based learning), and other generic methods based on sampling, clustering, and cooperation. The review also examines the initialization of metaheuristics with local search approaches, offers guidance on designing a diverse and informative sequence of initial solutions, and provides insights that will help research in constrained and discrete optimization problems.
    \item \textbf{Metaheuristic optimization algorithms: a comprehensive overview and classification of benchmark test functions - 2024 \cite{Sharma2024}}: This work focuses on the practical scenario of developing a new metaheuristic. It reviews over 200 mathematical test functions and more than 50 real-world engineering design problems. For each function, it is described its variables and value range. Due to the design of a metaheuristic should be accompanied by a set of experiments, this paper provides researchers with a wide range of options to test the quality of their new developments.
    \item \textbf{A literature review and critical analysis of metaheuristics recently developed - 2024  \cite{Velasco2024}}: This review focuses on algorithms with titles containing words such as `new', `hybrid', or `improved', in response to the growing trend of nature-based approaches. After analyzing over 100 algorithms, it was found that a significant percentage of these algorithms outperform previous techniques. From the several analyses made in this review, it is noted that most new algorithms are an improved version of some established algorithm, which reveals that the trend is no longer to propose metaheuristics based on new analogies. Moreover, they compare Black Widow Optimization and Coral Reef Optimization, which are considered new frameworks. By analyzing the components of both metaheuristics, authors evident the lack of innovation, as the operators of such algorithms are merely a combination of other evolutionary operators.
    \item \textbf{Metaheuristic optimization algorithms: an overview - 2024 \cite{Brahim2024}}: This paper focuses on studying the main components and concepts of optimization. More specifically, the overview provides the advantages (agnostic to the problem being solved, gradient independence, global search capability, the capability of dealing with multi-objective optimization problems, balanced exploration and exploitation, configurability and tuning, practical problem-solving, and innovation) and the limitations (absence of global optimality guarantee, convergence speed, parameter tuning, and black-box nature) of metaheuristics. The authors specifically focus on the references used by the algorithms to guide the search, and on how to achieve a good balance between exploration and exploitation. Visual representations accompany the text to illustrate the behavior of a set of metaheuristics.
    \item \textbf{50 years of metaheuristics - 2024 \cite{Marti2024}}: This overview traces the last 50 years of the field, starting from the roots of the area to the latest proposals to hybridize metaheuristics with machine learning. The revision encompasses constructive (GRASP and ACO), local search (iterated local search, Tabu search, variable neighborhood search), and population-based heuristics (memetic algorithms, biased random-key genetic algorithms, scatter search, and path relinking). Each category presents its core characteristics and the description of the mentioned algorithms. This review presents metaheuristic frameworks to guide the design of heuristic optimization algorithms during the last 50 years. It discusses the role of the journal in which it is published in introducing solid heuristic papers. This work also recalls the maturity of the field, which leads to solving very complex problems, with a growing number of researchers applying them, as shown in the numerous conferences and related events. Also, they criticize the fragmentation as each group of research usually applies the same methods regardless of the type of problem being solved, the lack of theoretical foundations,  the limited analytical understanding of novel proposals, the problem-specific tuning of metaheuristics, the lack of standardized benchmarking protocols and the absence of general guidelines. Several research directions are also annotated for researchers to be applied in the future.
    \item \textbf{Learn to optimize – A brief overview - 2024 \cite{Tang2024}}: This paper discusses the concept of Learn to Optimize (L2O) and its application in accelerating the configuration process to obtain a good solver for unseen instances. The studies can be categorized into three main types: training a solver performance prediction model, training a single solver, and training a portfolio of solvers. The first category aims to connect problem instance features with solver performance, resulting in the selection of the best solver, as seen in Automated Algorithm Selection (AAS). The second category, training a single solver, involves finding the best solver for overall performance on the training instances, known as Automatic Algorithm Configuration (AAC). The last category, training a portfolio of solvers, is a more general case of the second category in which a set of solvers is trained, introducing a higher degree of freedom. L2O has achieved importance in general-purpose approaches and other problems like adversarial attacks.
\end{itemize}

\section{Conclusions} \label{sec:conclusions}

\indent Nature and biological organisms have been a source of inspiration for many optimization algorithms. During the last few years, this family of solvers has grown considerably in size, achieving unseen levels of diversity about their source of inspiration. This explosion of literature has made it difficult for the community to appraise the general trajectory followed by the field, which is a necessary step towards identifying research trends and challenges of scientific value and practical impact. Some efforts have been dedicated so far towards classifying the state of the art on nature- and bio-inspired optimization in a taxonomy with well-defined criteria, allowing researchers to classify existing algorithms and newly proposed schemes. 

We have reviewed 518 nature- and bio-inspired algorithms and grouped them into two taxonomies. The first taxonomy has considered the source of inspiration, while the second has discriminated algorithms based on their behavior in generating new candidate solutions. We have provided clear descriptions, examples, and an enumeration of the reviewed approaches within each taxonomy category. Our study has critically examined the reviewed literature and found that many algorithms claiming to be inspired by different natural and biological phenomena exhibit algorithmic similarities. Additionally, a significant percentage (24\%) of the reviewed proposals have been identified as versions of classical algorithms such as PSO, DE, or GA. These findings shed light on the ongoing debate within the nature- and bio-inspired community regarding the algorithmic contributions of recent advances in the field.

A critical point of reflection associated with this explosion of proposals has been that novel metaphors do not lead to new solvers, and that comparisons undergo serious methodological problems. Although there are increasingly more bio-inspired algorithms, many of them rely on so-claimed novel metaphors that do not create any innovative bio-inspired solvers. In addition, comparisons have been often inadequate, leading to problems of reproducibility and applicability. This problem has captured the interest of other researchers, leading to several papers on various aspects related to bad comparisons and the increasing number of unoriginal proposals, even to the point of not accepting completely new proposals with quality marks. As we have mentioned, we emphasize that in these new algorithms there exists a lack of justification together with the lack of comparison with the state of the art and the lack of real interest in achieving reasonable levels of quality from the perspective of the optimization of well-known problems in recent competitions. Good methodological practices must be followed in forthcoming studies when designing, describing, and comparing new algorithms. 

From a positive vision, bio-inspired algorithms have been regularly used in AI and real-world applications. These algorithms hold potential in new scientific avenues, contributing to recent advances in DL evolution \cite{MARTINEZ2021161}, the design of large language models (LLM) \cite{Wu2024evolutionary}, and more recently, the design and enrichment of GPAIS \cite{Triguero2024}. GPAIS (including DL evolution and generative AI, such as LLM) are capable of performing tasks beyond those for which they were originally designed. In this context, the paradigm of AI-powered AI, which involves utilizing AI algorithms to enhance other AI systems, assumes paramount importance. In this thrilling era of AI explosion and advancement, we are already witnessing the significant impact of bio-inspired algorithms on the improvement of AI systems through examples like POET \cite{Wang2019}, EUREKA \cite{JasonMa2023}, EvoPrompt \cite{Guo2024}, and EGANs \cite{Chen2023}, among others.

In the last update of this report, which is herein released 4 years after its original version, we note that there has been an evolution within the nature and bio-inspired optimization field. There is an excessive use of the biological approach as opposed to the real problem-solving approach to tackle real and complex optimization goals, as those discussed in Section \ref{sec:hopeful2}. This issue needs to be addressed in the future by following guidelines that will allow for the definition of metaheuristics in a way that is appropriate to current challenges. This is important for the constructive design and development of proposals in response to emerging problems. For this reason, the potential impact the emerging problems and GPAIS, population-based metaheuristics as nature and bio-inspired optimization algorithms are poised to shape the future of AI, contributing to the design of continuously emerging AI systems, and serving as an inspiration for the new era of innovation and progress in AI.

\bibliographystyle{vancouver}
\bibliography{paper_taxonomy}

\begin{thebibliography}{100}

\bibitem{Molina2020comprehensive}
Molina D, Poyatos J, Ser JD, Garc{\'\i}a S, Hussain A, Herrera F.
\newblock Comprehensive taxonomies of nature-and bio-inspired optimization:
  Inspiration versus algorithmic behavior, critical analysis recommendations.
\newblock Cognitive Computation. 2020;12; p. 897--939.

\bibitem{Molina2022nature}
Molina~Cabrera D, Poyatos~Amador J, Osaba~Icedo E, Del Ser~Lorente J,
  Herrera~Triguero F.
\newblock Nature-and bio-inspired optimization: the good, the bad, the ugly and
  the hopeful.
\newblock DYNA Ingenier{\'\i}a e Industria. 2022;97(2); p. 114--117.

\bibitem{DELSER2019220}
{Del Ser} J, Osaba E, Molina D, Yang XS, Salcedo-Sanz S, Camacho D, et~al.
\newblock Bio-inspired computation: Where we stand and what's next.
\newblock Swarm and Evolutionary Computation. 2019;48; p. 220--250.

\bibitem{LATORRE2021}
LaTorre A, Molina D, Osaba E, Poyatos J, {Del Ser} J, Herrera F.
\newblock A prescription of methodological guidelines for comparing
  bio-inspired optimization algorithms.
\newblock Swarm and Evolutionary Computation. 2021;67; p. 100973.

\bibitem{OSABA2021}
Osaba E, Villar-Rodriguez E, {Del Ser} J, Nebro AJ, Molina D, LaTorre A, et~al.
\newblock A Tutorial On the design, experimentation and application of
  metaheuristic algorithms to real-World optimization problems.
\newblock Swarm and Evolutionary Computation. 2021;64; p. 100888.

\bibitem{Pintea2014}
Pintea CM.
\newblock Bio-inspired Computing.
\newblock In: Advances in Bio-inspired Computing for Combinatorial Optimization
  Problems. Springer Berlin Heidelberg; 2014. p. 3--19.

\bibitem{mahdavi2015metaheuristics}
Mahdavi S, Shiri ME, Rahnamayan S.
\newblock Metaheuristics in large-scale global continues optimization: A
  survey.
\newblock Information Sciences. 2015;295; p. 407--428.

\bibitem{MARTINEZ2021161}
Martinez AD, {Del Ser} J, Villar-Rodriguez E, Osaba E, Poyatos J, Tabik S,
  et~al.
\newblock Lights and shadows in Evolutionary Deep Learning: Taxonomy, critical
  methodological analysis, cases of study, learned lessons, recommendations and
  challenges.
\newblock Information Fusion. 2021;67; p. 161--194.

\bibitem{review8}
Sörensen K.
\newblock Metaheuristics--the metaphor exposed.
\newblock International Transactions In Operational Research. 2015;22(1); p.
  3--18.

\bibitem{fister2016new}
Fister~Jr I, Mlakar U, Brest J, Fister I.
\newblock A new population-based nature-inspired algorithm every month: is the
  current era coming to the end.
\newblock In: Proceedings of the 3rd Student Computer Science Research
  Conference; 2016. p. 33--37.

\bibitem{weyland2015critical}
Weyland D.
\newblock A critical analysis of the harmony search algorithm -- how not to
  solve sudoku.
\newblock Operations Research Perspectives. 2015;2; p. 97--105.

\bibitem{saka2016metaheuristics}
Saka MP, Hasan{\c{c}}ebi O, Geem ZW.
\newblock Metaheuristics in structural optimization and discussions on harmony
  search algorithm.
\newblock Swarm and Evolutionary Computation. 2016;28; p. 88--97.

\bibitem{PIOTROWSKI2014191}
Piotrowski AP, Napiorkowski JJ, Rowinski PM.
\newblock How novel is the “novel” black hole optimization approach?
\newblock Information Sciences. 2014;267; p. 191--200.

\bibitem{Camacho2020}
Camacho~Villal{\'o}n CL, St{\"u}tzle T, Dorigo M.
\newblock Grey {W}olf, {F}irefly and {B}at {A}lgorithms: Three {W}idespread
  {A}lgorithms that {D}o {N}ot {C}ontain {A}ny {N}ovelty.
\newblock In: International Conference on Swarm Intelligence. vol. 12421; 2020.
  p. 121--133.

\bibitem{Tzanetos2021}
Tzanetos A, Dounias G.
\newblock Nature inspired optimization algorithms or simply variations of
  metaheuristics?
\newblock Artificial Intelligence Review. 2021;54(3); p. 1841--1862.

\bibitem{Aranha2022}
Aranha C, Camacho~Villal{\'o}n CL, Campelo F, Dorigo M, Ruiz R, Sevaux M,
  et~al.
\newblock Metaphor-based metaheuristics, a call for action: the elephant in the
  room.
\newblock Swarm Intelligence. 2022;16; p. 1--6.

\bibitem{Piotrowski2018}
Piotrowski AP, Napiorkowski JJ.
\newblock Some metaheuristics should be simplified.
\newblock Information Sciences. 2018;427; p. 32--62.

\bibitem{Camacho2018}
Camacho-Villal{\'o}n CL, Dorigo M, St{\"u}tzle T.
\newblock Why the {I}ntelligent {W}ater {D}rops {C}annot {B}e {C}onsidered as a
  {N}ovel {A}lgorithm.
\newblock In: Swarm Intelligence; 2018. p. 302--314.

\bibitem{Camacho2019intelligent}
Camacho-Villal{\'o}n CL, Dorigo M, St{\"u}tzle T.
\newblock The intelligent water drops algorithm: why it cannot be considered a
  novel algorithm: A brief discussion on the use of metaphors in optimization.
\newblock Swarm Intelligence. 2019;13; p. 173--192.

\bibitem{Camacho2022}
Camacho-Villalón CL, Dorigo M, Stützle T.
\newblock An analysis of why cuckoo search does not bring any novel ideas to
  optimization.
\newblock Computers \& Operations Research. 2022;142; p. 105747.

\bibitem{Kudela2022}
Kudela J.
\newblock A critical problem in benchmarking and analysis of evolutionary
  computation methods.
\newblock Nature Machine Intelligence. 2022;4(12); p. 1238--1245.

\bibitem{Campelo2023}
Campelo F, Aranha C.
\newblock {Lessons from the Evolutionary Computation Bestiary}.
\newblock Artificial Life. 2023;29(4); p. 421--432.

\bibitem{Tzanetos2023}
Tzanetos A.
\newblock {Does the Field of Nature-Inspired Computing Contribute to Achieving
  Lifelike Features?}
\newblock Artificial Life. 2023;29(4); p. 487--511.

\bibitem{Camacho2023b}
Camacho-Villalón CL, Dorigo M, Stützle T.
\newblock Exposing the grey wolf, moth-flame, whale, firefly, bat, and antlion
  algorithms: six misleading optimization techniques inspired by bestial
  metaphors.
\newblock International Transactions in Operational Research. 2023;30(6); p.
  2945--2971.

\bibitem{kudela2023evolutionary}
Kudela J. The {E}volutionary {C}omputation {M}ethods {N}o {O}ne {S}hould {U}se;
  2023.

\bibitem{Hu2024}
Hu Z, Zhang Q, Wang Y, Su Q, Xiong Z.
\newblock Research orientation and novelty discriminant for new metaheuristic
  algorithms.
\newblock Applied Soft Computing. 2024;157; p. 111521.

\bibitem{DeArmas2022}
de~Armas J, Lalla-Ruiz E, Tilahun SL, Vo{\ss} S.
\newblock Similarity in metaheuristics: A gentle step towards a comparison
  methodology.
\newblock Natural Computing. 2022;21; p. 265--287.

\bibitem{Swan2022}
Swan J, Adriaensen S, Brownlee AEI, Hammond K, Johnson CG, Kheiri A, et~al.
\newblock Metaheuristics {{"In the Large”}}.
\newblock European Journal of Operational Research. 2022;297(2); p. 393--406.

\bibitem{Camacho2023}
Camacho-Villal{\'o}n CL, St{\"u}tzle T, Dorigo M.
\newblock Designing {N}ew {M}etaheuristics: {M}anual {V}ersus {A}utomatic
  {A}pproaches.
\newblock Intelligent Computing. 2023;2; p. 0048.

\bibitem{Jia2024}
Jia H, Lu C.
\newblock Guided learning strategy: A novel update mechanism for metaheuristic
  algorithms design and improvement.
\newblock Knowledge-Based Systems. 2024;286; p. 111402.

\bibitem{Walden2024}
Walden A, Buzdalov M.
\newblock A {S}imple {S}tatistical {T}est {A}gainst {O}rigin-{B}iased
  {M}etaheuristics.
\newblock In: International Conference on the Applications of Evolutionary
  Computation; 2024. p. 322--337.

\bibitem{Fister2021}
Fister I, Fister I, Iglesias A, Galvez A.
\newblock On detecting the novelties in metaphor-based algorithms.
\newblock In: Proceedings of the Genetic and Evolutionary Computation
  Conference Companion; 2021. p. 71–72.

\bibitem{review2}
Stork J, Eiben AE, Bartz-Beielstein T.
\newblock A new taxonomy of global optimization algorithms.
\newblock Natural Computing. 2020;(21); p. 219--242.

\bibitem{Rajwar2023}
Rajwar K, Deep K, Das S.
\newblock An exhaustive review of the metaheuristic algorithms for search and
  optimization: Taxonomy, applications, and open challenges.
\newblock Artificial Intelligence Review. 2023;56; p. 13187--13257.

\bibitem{Sarhani2023}
Sarhani M, Vo{\ss} S, Jovanovic R.
\newblock Initialization of metaheuristics: comprehensive review, critical
  analysis, and research directions.
\newblock International Transactions in Operational Research. 2023;30(6); p.
  3361--3397.

\bibitem{Ferrer2023}
Ferrer J, Delgado-P{\'e}rez P.
\newblock Metaheuristics in a {N}utshell.
\newblock In: Optimising the Software Development Process with Artificial
  Intelligence. Springer; 2023. p. 279--307.

\bibitem{Sharma2024}
Sharma P, Raju S.
\newblock Metaheuristic optimization algorithms: A comprehensive overview and
  classification of benchmark test functions.
\newblock Soft Computing. 2024;28(4); p. 3123--3186.

\bibitem{Velasco2024}
Velasco L, Guerrero H, Hospitaler A.
\newblock A literature review and critical analysis of metaheuristics recently
  developed.
\newblock Archives of Computational Methods in Engineering. 2024;31; p.
  125--146.

\bibitem{Brahim2024}
Brahim B, Kobayashi M, Al~Ali M, Khatir T, Elmeliani MEAE.
\newblock Metaheuristic Optimization Algorithms: an overview.
\newblock HCMCOU Journal of Science--Advances in Computational Structures.
  2024;14(1); p. 1--28.

\bibitem{Marti2024}
Martí R, Sevaux M, Sörensen K.
\newblock 50 years of metaheuristics.
\newblock In Press European Journal of Operational Research. 2024;DOI:
  10.1016/j.ejor.2024.04.004.

\bibitem{Tang2024}
Tang K, Yao X.
\newblock Learn to Optimize-A Brief Overview.
\newblock National Science Review. 2024;p. 1--10.

\bibitem{KAR201620}
Kar AK.
\newblock Bio inspired computing – A review of algorithms and scope of
  applications.
\newblock Expert Systems with Applications. 2016;59; p. 20--32.

\bibitem{walkMH}
Xiong N, Molina D, Ortiz ML, Herrera F.
\newblock A walk into metaheuristics for engineering optimization: principles,
  methods and recent trends.
\newblock International Journal of Computational Intelligence Systems.
  2015;8(4); p. 606--636.

\bibitem{Molina2018Comp}
Molina D, LaTorre A, Herrera F.
\newblock An Insight into Bio-inspired and Evolutionary Algorithms for Global
  Optimization: Review, Analysis, and Lessons Learnt over a Decade of
  Competitions.
\newblock Cognitive Computation. 2018;10(4); p. 517--544.

\bibitem{Zavala2014}
Zavala GR, Nebro AJ, Luna F, Coello~Coello CA.
\newblock A survey of multi-objective metaheuristics applied to structural
  optimization.
\newblock Structural and Multidisciplinary Optimization. 2014;49(4); p.
  537--558.

\bibitem{BioTele}
Yang XS, Chien SF, Ting TO.
\newblock Bio-Inspired Computation in Telecommunications.
\newblock Morgan Kaufmann; 2015. p. 1--21.

\bibitem{SI}
Beni G, Wang J.
\newblock Swarm Intelligence in Cellular Robotic Systems.
\newblock In: Robots and Biological Systems: Towards a New Bionics?; 1993. p.
  703--712.

\bibitem{FONG2013385}
Fong S.
\newblock Opportunities and Challenges of Integrating Bio-Inspired Optimization
  and Data Mining Algorithms.
\newblock In: Swarm Intelligence and Bio-Inspired Computation. Elsevier; 2013.
  p. 385--402.

\bibitem{del2019bioinspired}
Del~Ser J, Osaba E, Sanchez-Medina JJ, Fister I.
\newblock Bioinspired computational intelligence and transportation systems: a
  long road ahead.
\newblock IEEE Transactions on Intelligent Transportation Systems. 2019;21(2);
  p. 466--495.

\bibitem{PSOPower}
del Valle Y, Venayagamoorthy GK, Mohagheghi S, Hernandez J, Harley RG.
\newblock Particle Swarm Optimization: Basic Concepts, Variants and
  Applications in Power Systems.
\newblock IEEE Transactions on Evolutionary Computation. 2008;12(2); p.
  171--195.

\bibitem{DRESSLER2010881}
Dressler F, Akan OB.
\newblock A survey on bio-inspired networking.
\newblock Computer Networks. 2010;54(6); p. 881 -- 900.

\bibitem{JOSEGARCIA2016192}
José-García A, Gómez-Flores W.
\newblock Automatic clustering using nature-inspired metaheuristics: A survey.
\newblock Applied Soft Computing. 2016;41; p. 192 -- 213.

\bibitem{BioFace}
Alsalibi B, Venkat I, Subramanian KG, Lutfi SL, Wilde PD.
\newblock The Impact of Bio-Inspired Approaches Toward the Advancement of Face
  Recognition.
\newblock ACM Computing Surveys. 2015;48(5); p. 1--33.

\bibitem{garcia2019bio}
Garc{\'\i}a-Godoy MJ, L{\'o}pez-Camacho E, Garc{\'\i}a-Nieto J, Del~Ser J,
  Nebro AJ, Aldana-Montes JF.
\newblock Bio-inspired optimization for the molecular docking problem: State of
  the art, recent results and perspectives.
\newblock Applied Soft Computing. 2019;79; p. 30--45.

\bibitem{KOLIAS2011625}
Kolias C, Kambourakis G, Maragoudakis M.
\newblock Swarm intelligence in intrusion detection: A survey.
\newblock Computers and Security. 2011;30(8); p. 625--642.

\bibitem{PSOReview}
Banks A, Vincent J, Anyakoha C.
\newblock A review of particle swarm optimization. Part I: background and
  development.
\newblock Natural Computing. 2007;6(4); p. 467--484.

\bibitem{Neri2010}
Neri F, Tirronen V.
\newblock Recent advances in differential evolution: a survey and experimental
  analysis.
\newblock Artificial Intelligence Review. 2010;33; p. 61--106.

\bibitem{DEDas1}
Das S, Suganthan PN.
\newblock Differential Evolution: A Survey of the State-of-the-Art.
\newblock IEEE Transactions on Evolutionary Computation. 2011;15(1); p. 4--31.

\bibitem{DEDas2}
Das S, Mullick SS, Suganthan PN.
\newblock Recent advances in differential evolution – An updated survey.
\newblock Swarm and Evolutionary Computation. 2016;27; p. 1--30.

\bibitem{Karaboga2014}
Karaboga D, Gorkemli B, Ozturk C, Karaboga N.
\newblock A comprehensive survey: artificial bee colony (ABC) algorithm and
  applications.
\newblock Artificial Intelligence Review. 2014;42; p. 21--57.

\bibitem{bereview}
Bitam S, Batouche M, Talbi E.
\newblock A survey on bee colony algorithms.
\newblock In: 2010 IEEE International Symposium on Parallel Distributed
  Processing, Workshops and Phd Forum (IPDPSW); 2010. p. 1--8.

\bibitem{BFADas}
Das S, Biswas A, Dasgupta S, Abraham A.
\newblock Bacterial Foraging Optimization Algorithm: Theoretical Foundations,
  Analysis, and Applications.
\newblock In: Foundations of Computational Intelligence Volume 3: Global
  Optimization. Springer Berlin Heidelberg; 2009. p. 23--55.

\bibitem{BatReview}
Yang XS, He X.
\newblock Bat Algorithm: Literature Review and Applications.
\newblock International Journal of Bio-Inspired Computation. 2013;5(3); p.
  141--149.

\bibitem{SIBook}
Bonabeau E, Dorigo M, Th{\'e}raulaz G.
\newblock Swarm intelligence: from natural to artificial systems.
\newblock Oxford University press; 1999.

\bibitem{naturebook}
Yang XS.
\newblock Nature-Inspired Optimization Algorithms.
\newblock Elsevier; 2014.

\bibitem{PSODEDas}
Das S, Abraham A, Konar A.
\newblock Particle Swarm Optimization and Differential Evolution Algorithms:
  Technical Analysis, Applications and Hybridization Perspectives.
\newblock In: Advances of Computational Intelligence in Industrial Systems.
  Springer Berlin Heidelberg; 2008. p. 1--38.

\bibitem{ELBELTAGI200543}
Elbeltagi E, Hegazy T, Grierson D.
\newblock Comparison among five evolutionary-based optimization algorithms.
\newblock Advanced Engineering Informatics. 2005;19(1); p. 43--53.

\bibitem{pazhaniraja2017}
Pazhaniraja N, Paul PV, Roja G, Shanmugapriya K, Sonali B.
\newblock A study on recent bio-inspired optimization algorithms.
\newblock In: 2017 Fourth International Conference on Signal Processing,
  Communication and Networking (ICSCN); 2017. p. 1--6.

\bibitem{BioGPU}
Kr{\"o}mer P, Plato{\v{s}} J, Sn{\'a}{\v{s}}el V.
\newblock Nature-Inspired Meta-Heuristics on Modern GPUs: State of the Art and
  Brief Survey of Selected Algorithms.
\newblock International Journal of Parallel Programming. 2014;42(5); p.
  681--709.

\bibitem{review10}
Piotrowski AP, Napiorkowski M, Napiorkowski JJ, Rowinski PM.
\newblock Swarm Intelligence and Evolutionary Algorithms: performance versus
  speed.
\newblock Information Sciencies. 2017;384; p. 34--85.

\bibitem{ForagingvsOther}
El-Abd M.
\newblock Performance assessment of foraging algorithms vs. evolutionary
  algorithms.
\newblock Information Sciences. 2012;182(1); p. 243--263.

\bibitem{chouikhi2019bi}
Chouikhi N, Ammar B, Hussain A, Alimi AM.
\newblock Bi-level multi-objective evolution of a Multi-Layered Echo-State
  Network Autoencoder for data representations.
\newblock Neurocomputing. 2019;341; p. 195--211.

\bibitem{chouikhi2017pso}
Chouikhi N, Ammar B, Rokbani N, Alimi AM.
\newblock PSO-based analysis of Echo State Network parameters for time series
  forecasting.
\newblock Applied Soft Computing. 2017;55; p. 211--225.

\bibitem{review_nature}
Fister~jr I, Yang XS, Fister I, Brest J, Fister D.
\newblock A Brief Review of Nature-Inspired Algorithms for Optimization.
\newblock Elektrotehniski Vestnik. 2013;80(3); p. 1--7.

\bibitem{Venga2015}
Baskaran A, Balaji N, Basha S, Vengattaraman T.
\newblock A survey of nature inspired algorithms.
\newblock International Journal of Applied Engineering Research. 2015;10; p.
  19313--19324.

\bibitem{Venga2016}
Rajakumar R, Dhavachelvan P, Vengattaraman T.
\newblock A survey on nature inspired meta-heuristic algorithms with its domain
  specifications.
\newblock In: 2016 International Conference on Communication and Electronics
  Systems (ICCES); 2016. p. 1--6.

\bibitem{review9}
Kumar~Kar A.
\newblock Bio inspired computing – A review of algorithms and scope of
  applications.
\newblock Expert Systems With Applications. 2016;59; p. 20--32.

\bibitem{Chu2018}
Chu X, Wu T, Weir JD, Shi Y, Niu B, Li L.
\newblock Learning--interaction--diversification framework for swarm
  intelligence optimizers: a unified perspective.
\newblock Neural Computing and Applications. 2020;(32); p. 1789–--1809.

\bibitem{SA}
Kirkpatrick S, Gelatt CD, P VM.
\newblock Optimization by Simulated Annealing.
\newblock Science. 1989;220(4598); p. 671--680.

\bibitem{PSO}
{Eberhart} R, {Kennedy} J.
\newblock A new optimizer using particle swarm theory.
\newblock In: MHS'95. Proceedings of the Sixth International Symposium on Micro
  Machine and Human Science; 1995. p. 39--43.

\bibitem{CSA2}
Braik MS.
\newblock Chameleon Swarm Algorithm: A bio-inspired optimizer for solving
  engineering design problems.
\newblock Expert Systems with Applications. 2021;174; p. 114685.

\bibitem{artificialecosystem}
Adham M, Bentley P.
\newblock An Artificial Ecosystem Algorithm applied to static and Dynamic
  Travelling Salesman Problems.
\newblock In: IEEE SSCI 2014 - 2014 IEEE Symposium Series on Computational
  Intelligence - IEEE ICES: 2014 IEEE International Conference on Evolvable
  Systems, Proceedings; 2015. p. 149--156.

\bibitem{Zhao2020}
Zhao W, Wang L, Zhang Z.
\newblock Artificial ecosystem-based optimization: a novel nature-inspired
  meta-heuristic algorithm.
\newblock Neural Computing and Applications. 2020;32; p. 9383--9425.

\bibitem{AIDO}
Huang G.
\newblock Artificial infectious disease optimization: A SEIQR epidemic dynamic
  model-based function optimization algorithm.
\newblock Swarm and Evolutionary Computation. 2016;27; p. 31--67.

\bibitem{ARO}
Farasat A, Menhaj MB, Mansouri T, Sadeghi~Moghadamd MR.
\newblock ARO: A new model-free optimization algorithm inspired from asexual
  reproduction.
\newblock Applied Soft Computing. 2010;10(4); p. 1284--1292.

\bibitem{BBO}
{Simon} D.
\newblock Biogeography-Based Optimization.
\newblock IEEE Transactions on Evolutionary Computation. 2008;12(6); p.
  702--713.

\bibitem{BMO}
Askarzadeh A.
\newblock Bird mating optimizer: An optimization algorithm inspired by bird
  mating strategies.
\newblock Communications in Nonlinear Science and Numerical Simulation.
  2014;19(4); p. 1213 -- 1228.

\bibitem{BOA}
Zhang X, Sun B, Mei T, Wang R.
\newblock Post-disaster restoration based on fuzzy preference relation and Bean
  Optimization Algorithm.
\newblock In: 2010 IEEE Youth Conference on Information, Computing and
  Telecommunications; 2010. p. 271--274.

\bibitem{Yuan2023}
Yuan Y, Shen Q, Wang S, Ren J, Yang D, Yang Q, et~al.
\newblock Coronavirus mask protection algorithm: A new bio-inspired
  optimization algorithm and its applications.
\newblock Journal of Bionic Engineering. 2023;20; p. 1747--1765.

\bibitem{Khalid2022}
Khalid AM, Hosny KM, Mirjalili S.
\newblock COVIDOA: a novel evolutionary optimization algorithm based on
  coronavirus disease replication lifecycle.
\newblock Neural Computing and Applications. 2022;34(24); p. 22465--22492.

\bibitem{CRO}
Salcedo-Sanz S, Del~Ser J, Landa-Torres I, Gil-L{\'o}pez S, Portilla-Figueras
  J.
\newblock The coral reefs optimization algorithm: a novel metaheuristic for
  efficiently solving optimization problems.
\newblock The Scientific World Journal. 2014;2014; p. 739768.

\bibitem{greensmith2005introducing}
Greensmith J, Aickelin U, Cayzer S.
\newblock Introducing dendritic cells as a novel immune-inspired algorithm for
  anomaly detection.
\newblock In: International Conference on Artificial Immune Systems; 2005. p.
  153--167.

\bibitem{DE}
Price K, Storn R.
\newblock A simple and efficient heuristic for global optimization over
  continuous spaces.
\newblock Journal of Global Optimization. 1997;11(4); p. 341--359.

\bibitem{EBO}
Zheng YJ, Ling HF, Xue JY.
\newblock Ecogeography-based optimization: Enhancing biogeography-based
  optimization with ecogeographic barriers and differentiations.
\newblock Computers and Operations Research. 2014;50; p. 115--127.

\bibitem{EEA}
{Parpinelli} RS, {Lopes} HS.
\newblock An eco-inspired evolutionary algorithm applied to numerical
  optimization.
\newblock In: 2011 Third World Congress on Nature and Biologically Inspired
  Computing; 2011. p. 466--471.

\bibitem{wang2018earthworm}
Wang GG, Deb S, dos Santos~Coelho L.
\newblock Earthworm optimisation algorithm: a bio-inspired metaheuristic
  algorithm for global optimisation problems.
\newblock IJBIC. 2018;12(1); p. 1--22.

\bibitem{Beyer2002}
Beyer HG, Schwefel HP.
\newblock Evolution strategies -- A comprehensive introduction.
\newblock Natural Computing. 2002;1; p. 3--52.

\bibitem{GA}
{Man} KF, {Tang} KS, {Kwong} S.
\newblock Genetic algorithms: concepts and applications [in engineering
  design].
\newblock IEEE Transactions on Industrial Electronics. 1996;43(5); p. 519--534.

\bibitem{GE}
Ferreira C.
\newblock Gene Expression Programming in Problem Solving.
\newblock In: Soft Computing and Industry: Recent Applications. Springer
  London; 2002. p. 635--653.

\bibitem{HRO}
Ye Z, Ma L, Chen H.
\newblock A hybrid rice optimization algorithm.
\newblock In: 2016 11th International Conference on Computer Science Education
  (ICCSE); 2016. p. 169--174.

\bibitem{Cortes}
Cort{\'e}s P, Garc{\'i}a JM, Onieva L, Mu{\~{n}}uzuri J, Guadix J.
\newblock Viral System to Solve Optimization Problems: An Immune-Inspired
  Computational Intelligence Approach.
\newblock In: Artificial Immune Systems; 2008. p. 83--94.

\bibitem{tayeb2017research}
Tayeb FBS, Bessedik M, Benbouzid M, Cheurfi H, Blizak A.
\newblock Research on permutation flow-shop scheduling problem based on
  improved genetic immune algorithm with vaccinated offspring.
\newblock Procedia Computer Science. 2017;112; p. 427--436.

\bibitem{IWO}
Mehrabian AR, Lucas C.
\newblock A novel numerical optimization algorithm inspired from weed
  colonization.
\newblock Ecological Informatics. 2006;1(4); p. 355--366.

\bibitem{MHBO}
{Abbass} HA.
\newblock MBO: marriage in honey bees optimization-a Haplometrosis polygynous
  swarming approach.
\newblock In: Proceedings of the 2001 IEEE Congress on Evolutionary
  Computation. vol.~1; 2001. p. 207--214.

\bibitem{MRO}
Bidar M, Kanan HR, Mouhoub M, Sadaoui S.
\newblock Mushroom Reproduction Optimization (MRO): A Novel Nature-Inspired
  Evolutionary Algorithm.
\newblock In: 2018 IEEE Congress on Evolutionary Computation (CEC); 2018. p.
  1--10.

\bibitem{QBE}
{Sung Hoon Jung}.
\newblock Queen-bee evolution for genetic algorithms.
\newblock Electronics Letters. 2003;39(6); p. 575--576.

\bibitem{anandaraman2012new}
Anandaraman C, Sankar AVM, Natarajan R.
\newblock A new evolutionary algorithm based on bacterial evolution and its
  application for scheduling a flexible manufacturing system.
\newblock Jurnal Teknik Industri. 2012;14(1); p. 1--12.

\bibitem{Taherdangkoo2011394}
Taherdangkoo M, Yazdi M, Bagheri MH.
\newblock Stem Cells Optimization Algorithm.
\newblock In: Bio-Inspired Computing and Applications; 2012. p. 394--403.

\bibitem{Nara1999}
{Nara} K, {Takeyama} T, {Hyungchul Kim}.
\newblock A new evolutionary algorithm based on sheep flocks heredity model and
  its application to scheduling problem.
\newblock In: IEEE SMC'99 Conference Proceedings. 1999 IEEE International
  Conference on Systems, Man, and Cybernetics. vol.~6; 1999. p. 503--508.

\bibitem{PATTNAIK2013628}
Pattnaik SS, Bakwad KM, Sohi BS, Ratho RK, Devi S.
\newblock Swine Influenza Models Based Optimization (SIMBO).
\newblock Applied Soft Computing. 2013;13(1); p. 628--653.

\bibitem{Zelinka2004}
Zelinka I.
\newblock SOMA --- Self-Organizing Migrating Algorithm.
\newblock In: New Optimization Techniques in Engineering. Springer Berlin
  Heidelberg; 2004. p. 167--217.

\bibitem{Zhang2023}
Zhang H, Zhang Y, Niu Y, He K, Wang Y.
\newblock T {C}ell {I}mmune {A}lgorithm: A {N}ovel {N}ature-{I}nspired
  {A}lgorithm for {E}ngineering {A}pplications.
\newblock IEEE Access. 2023;11; p. 95545--95566.

\bibitem{VMO}
Puris A, Bello R, Molina D, Herrera F.
\newblock Variable mesh optimization for continuous optimization problems.
\newblock Soft Computing. 2012;16(3); p. 511--525.

\bibitem{JADERYAN2016596}
Jaderyan M, Khotanlou H.
\newblock Virulence Optimization Algorithm.
\newblock Applied Soft Computing. 2016;43; p. 596--618.

\bibitem{ACO}
Dorigo M, Maniezzo V, Colorni A.
\newblock The Ant System: Optimization by a colony of cooperating agents.
\newblock IEEE Transactions on Systems, Man, and Cybernetics, Part B
  (Cybernetics). 1996;26(1); p. 29--41.

\bibitem{ABC}
Karaboga D, Basturk B.
\newblock A powerful and efficient algorithm for numerical function
  optimization: artificial bee colony (ABC) algorithm.
\newblock Journal of Global Optimization. 2007;39(3); p. 459--471.

\bibitem{FA}
Yang XS.
\newblock Firefly Algorithms for Multimodal Optimization.
\newblock In: Stochastic Algorithms: Foundations and Applications; 2009. p.
  169--178.

\bibitem{GOA}
Saremi S, Mirjalili S, Lewis A.
\newblock Grasshopper Optimisation Algorithm: Theory and application.
\newblock Advances in Engineering Software. 2017;105; p. 30--47.

\bibitem{Foraging2018}
Brabazon A, McGarraghy S.
\newblock Foraging-Inspired Optimisation Algorithms.
\newblock Natural Computing Series, Springer; 2018.

\bibitem{AAA}
Uymaz SA, Tezel G, Yel E.
\newblock Artificial algae algorithm (AAA) for nonlinear global optimization.
\newblock Applied Soft Computing. 2014;31; p. 153--171.

\bibitem{Munoz20091080}
Muñoz MA, López JA, Caicedo E.
\newblock An artificial beehive algorithm for continuous optimization.
\newblock International Journal of Intelligent Systems. 2009;24(11); p.
  1080--1093.

\bibitem{doi:10.1080/00207543.2013.871389}
Naderi B, Khalili M, Khamseh AA.
\newblock Mathematical models and a hunting search algorithm for the no-wait
  flowshop scheduling with parallel machines.
\newblock International Journal of Production Research. 2014;52(9); p.
  2667--2681.

\bibitem{ABO}
Odili JB, Mohmad~Kahar MN.
\newblock Solving the Traveling Salesman’s Problem Using the African Buffalo
  Optimization.
\newblock Computational Intelligence and Neuroscience. 2016;2016; p. 1510256.

\bibitem{Almonacid2019351}
Almonacid B, Soto R.
\newblock Andean Condor Algorithm for cell formation problems.
\newblock Natural Computing. 2019;18(2); p. 351--381.

\bibitem{AFB}
Lamy JB.
\newblock Artificial Feeding Birds (AFB): a new metaheuristic inspired by the
  behavior of pigeons.
\newblock In: Advances in nature-inspired computing and applications; 2019. p.
  43--60.

\bibitem{Zhao2022}
Zhao W, Wang L, Mirjalili S.
\newblock Artificial hummingbird algorithm: A new bio-inspired optimizer with
  its engineering applications.
\newblock Computer Methods in Applied Mechanics and Engineering. 2022;388; p.
  114194.

\bibitem{Zitouni2022}
Zitouni F, Harous S, Belkeram A, Hammou LEB.
\newblock The archerfish hunting optimizer: A novel metaheuristic algorithm for
  global optimization.
\newblock Arabian Journal for Science and Engineering. 2022;47(2); p.
  2513--2553.

\bibitem{AMO}
Li X, Zhang J, Yin M.
\newblock Animal migration optimization: an optimization algorithm inspired by
  animal migration behavior.
\newblock Neural Computing and Applications. 2014;24(7); p. 1867--1877.

\bibitem{Liu2022}
Liu R, Zhou N, Yao Y, Yu F.
\newblock An aphid inspired metaheuristic optimization algorithm and its
  application to engineering.
\newblock Scientific Reports. 2022;12; p. 18064.

\bibitem{ALO}
Mirjalili S.
\newblock The Ant Lion Optimizer.
\newblock Advances in Engineering Software. 2015;83; p. 80--98.

\bibitem{Abualigah2021b}
Abualigah L, Yousri D, {Abd Elaziz} M, Ewees AA, Al-qaness MAA, Gandomi AH.
\newblock Aquila {O}ptimizer: A novel meta-heuristic optimization algorithm.
\newblock Computers \& Industrial Engineering. 2021;157; p. 107250.

\bibitem{AOA}
Pook MF, Ramlan EI.
\newblock The Anglerfish algorithm: a derivation of randomized incremental
  construction technique for solving the traveling salesman problem.
\newblock Evolutionary Intelligence. 2019;12; p. 11--20.

\bibitem{Abualigah2021}
Abualigah L, Diabat A, Mirjalili S, {Abd Elaziz} M, Gandomi AH.
\newblock The Arithmetic Optimization Algorithm.
\newblock Computer Methods in Applied Mechanics and Engineering. 2021;376; p.
  113609.

\bibitem{Wang2022}
Wang L, Cao Q, Zhang Z, Mirjalili S, Zhao W.
\newblock Artificial rabbits optimization: A new bio-inspired meta-heuristic
  algorithm for solving engineering optimization problems.
\newblock Engineering Applications of Artificial Intelligence. 2022;114; p.
  105082.

\bibitem{Chen2009562}
{Chen} T, {Pang} L, {Jiang Du}, {Liu} Z, {Zhang} L.
\newblock Artificial Searching Swarm Algorithm for solving constrained
  optimization problems.
\newblock In: 2009 IEEE International Conference on Intelligent Computing and
  Intelligent Systems. vol.~1; 2009. p. 562--565.

\bibitem{Chen2012651}
Chen T, Wang Y, Li J.
\newblock Artificial tribe algorithm and its performance analysis.
\newblock Journal of Software. 2012;7(3); p. 651--656.

\bibitem{AWDA}
Subramanian C, Sekar ASS, Subramanian K.
\newblock A New Engineering Optimization Method African Wild Dog Algorithm.
\newblock International Journal of Soft Computing. 2013;8(3); p. 163--170.

\bibitem{Mohapatra2023}
Mohapatra S, Mohapatra P.
\newblock American zebra optimization algorithm for global optimization
  problems.
\newblock Scientific Reports. 2023;13; p. 5211.

\bibitem{BES}
Alsattar HA, Zaidan AA, Zaidan BB.
\newblock Novel meta-heuristic bald eagle search optimisation algorithm.
\newblock Artificial Intelligence Review. 2019;(53); p. 2237–2264.

\bibitem{BA}
Pham DT, Ghanbarzadeh A, Koç E, Otri S, Rahim S, Zaidi M.
\newblock The Bees Algorithm — A Novel Tool for Complex Optimisation
  Problems.
\newblock In: Intelligent Production Machines and Systems; 2006. p. 454--459.

\bibitem{BB}
Comellas F, Martinez-Navarro J.
\newblock Bumblebees: A Multiagent Combinatorial Optimization Algorithm
  Inspired by Social Insect Behaviour.
\newblock In: Proceedings of the First ACM/SIGEVO Summit on Genetic and
  Evolutionary Computation; 2009. p. 811--814.

\bibitem{Kazikova2019146}
Kazikova A, Pluhacek M, Senkerik R, Viktorin A.
\newblock Proposal of a new swarm optimization method inspired in bison
  behavior.
\newblock Advances in Intelligent Systems and Computing. 2019;837; p. 146--156.

\bibitem{Hackel}
H\"{a}ckel S, Dippold P.
\newblock The Bee Colony-inspired Algorithm (BCiA): A Two-stage Approach for
  Solving the Vehicle Routing Problem with Time Windows.
\newblock In: Proceedings of the 11th Annual Conference on Genetic and
  Evolutionary Computation; 2009. p. 25--32.

\bibitem{BCO}
Teodorović D, Dell'Orco M.
\newblock Bee colony optimization - A cooperative learning approach to complex
  transportation problems.
\newblock Advanced OR and AI Methods in Transportation. 2005;51; p. 51--60.

\bibitem{Niu2012501}
Niu B, Wang H.
\newblock Bacterial Colony Optimization: Principles and Foundations.
\newblock In: Emerging Intelligent Computing Technology and Applications; 2012.
  p. 501--506.

\bibitem{BCO1}
Müller SD, Marchetto J, Airaghi S, Koumoutsakos P.
\newblock Optimization Based on Bacterial Chemotaxis.
\newblock IEEE Transactions On Evolutionary Computation. 2002;6(1); p. 16--29.

\bibitem{BCO3}
Dutta T, Bhattacharyya S, Dey S, Platos J.
\newblock Border Collie Optimization.
\newblock IEEE Access. 2020;8; p. 109177--109197.

\bibitem{BFOA}
Lui Y, Passino KM.
\newblock Biomimicry of Social Foraging Bacteria for Distributed Optimization:
  Models, Principles, and Emergent Behaviors.
\newblock Journal of Optimization Theory and Applications. 2002;115(3); p.
  603--628.

\bibitem{BGAF}
{Chen} T, {Tsai} P, {Chu} S, {Pan} J.
\newblock A Novel Optimization Approach: Bacterial-GA Foraging.
\newblock In: Second International Conference on Innovative Computing,
  Informatio and Control (ICICIC 2007); 2007. p. 391--391.

\bibitem{BHA}
Wedde HF, Farooq M, Zhang Y.
\newblock BeeHive: An Efficient Fault-Tolerant Routing Algorithm Inspired by
  Honey Bee Behavior.
\newblock In: Ant Colony Optimization and Swarm Intelligence, Proceeding; 2004.
  p. 83--94.

\bibitem{Bitam2018373}
Bitam S, Zeadally S, Mellouk A.
\newblock Fog computing job scheduling optimization based on bees swarm.
\newblock Enterprise Information Systems. 2018;12(4); p. 373--397.

\bibitem{Malakooti20121071}
Malakooti B, Kim H, Sheikh S.
\newblock Bat intelligence search with application to multi-objective
  multiprocessor scheduling optimization.
\newblock International Journal of Advanced Manufacturing Technology.
  2012;60(9-12); p. 1071--1086.

\bibitem{BIA}
Yang XS.
\newblock A New Metaheuristic Bat-Inspired Algorithm.
\newblock In: Nature Inspired Cooperative Strategies for Optimization (NICSO
  2010). Springer; 2010. p. 65--74.

\bibitem{Zhang2019}
Zhang Q, Wang R, Yang J, Lewis A, Chiclana F, Yang S.
\newblock Biology migration algorithm: a new nature-inspired heuristic
  methodology for global optimization.
\newblock Soft Computing. 2019;23(16); p. 7333--7358.

\bibitem{BMO1}
Sulaiman MH, Mustaffa Z, Saari MM, Daniyal H, Mohamad AJ, Othman MR, et~al.
\newblock Barnacles Mating Optimizer Algorithm for Optimization.
\newblock In: Proceedings of the 10th National Technical Seminar on Underwater
  System Technology 2018; 2019. p. 211--218.

\bibitem{BNMR}
Taherdangkoo M, Shirzadi MH, Yazdi M, Bagher MH.
\newblock A robust clustering method based on blind, naked mole-rats (BNMR)
  algorithm.
\newblock Swarm and Evolutionary Computation. 2013;10; p. 1--11.

\bibitem{BO}
{Kumar} A, {Misra} RK, {Singh} D.
\newblock Butterfly optimizer.
\newblock In: 2015 IEEE Workshop on Computational Intelligence: Theories,
  Applications and Future Directions (WCI); 2015. p. 1--6.

\bibitem{BO1}
Das AK, Pratihar DK.
\newblock A new bonobo optimizer (BO) for real-parameter optimization.
\newblock In: 2019 IEEE Region 10 Symposium (TENSYMP); 2019. p. 108--113.

\bibitem{bulloptimization}
Findik O.
\newblock Bull optimization algorithm based on genetic operators for continuous
  optimization problems.
\newblock Turkish Journal of Electrical Engineering \& Computer Sciences.
  2015;23; p. 2225--2239.

\bibitem{Sato19973954}
Sato T, Hagiwara M.
\newblock Bee System: Finding solution by a concentrated search.
\newblock In: Proceedings of the IEEE International Conference on Systems, Man
  and Cybernetics. vol.~4; 1997. p. 3954--3959.

\bibitem{BS}
{Lucic} P, {Teodorovic} D.
\newblock Transportation modeling: an artificial life approach.
\newblock In: Proceedings of the 14th IEEE International Conference on Tools
  with Artificial Intelligence (ICTAI 2002); 2002. p. 216--223.

\bibitem{BSA}
Meng XB, Gao XZ, Lu L, Liu Y, Zhang H.
\newblock A new bio-inspired optimisation algorithm: Bird Swarm Algorithm.
\newblock Journal of Experimental and Theoretical Artificial Intelligence.
  2016;28(4); p. 673--687.

\bibitem{Akbari20103142}
Akbari R, Mohammadi A, Ziarati K.
\newblock A novel bee swarm optimization algorithm for numerical function
  optimization.
\newblock Communications in Nonlinear Science and Numerical Simulation.
  2010;15(10); p. 3142--3155.

\bibitem{Oliveira11}
de~Oliveira DR, Parpinelli RS, Lopes HS.
\newblock Bioluminescent Swarm Optimization Algorithm.
\newblock In: Evolutionary Algorithms. IntechOpen; 2011. .

\bibitem{Wang2023}
Wang L, Zhang Q, He X, Yang S, Jiang S, Dong Y.
\newblock Biological survival optimization algorithm with its engineering and
  neural network applications.
\newblock Soft Computing. 2023;27(10); p. 6437--6463.

\bibitem{BSOA}
Drias H, Sadeg S, Yahi S.
\newblock Cooperative Bees Swarm for Solving the Maximum Weighted
  Satisfiability Problem.
\newblock In: Computational Intelligence and Bioinspired Systems; 2005. p.
  318--325.

\bibitem{BUZOA}
Arshaghi A, Ashourian M, Ghabeli L.
\newblock Buzzard optimization algorithm: a nature-inspired metaheuristic
  algorithm.
\newblock Majlesi Journal of Electrical Engineering. 2019;13(3); p. 83--98.

\bibitem{BWO}
Hayyolalam V, {Pourhaji Kazem} AA.
\newblock Black Widow Optimization Algorithm: A novel meta-heuristic approach
  for solving engineering optimization problems.
\newblock Engineering Applications of Artificial Intelligence. 2020;87; p.
  103249.

\bibitem{Zhong2022}
Zhong C, Li G, Meng Z.
\newblock Beluga whale optimization: A novel nature-inspired metaheuristic
  algorithm.
\newblock Knowledge-Based Systems. 2022;251; p. 109215.

\bibitem{doi:10.1080/0305215X.2018.1463527}
K SR, Panwar L, Panigrahi BK, Kumar R.
\newblock Binary whale optimization algorithm: a new metaheuristic approach for
  profit-based unit commitment problems in competitive electricity markets.
\newblock Engineering Optimization. 2019;51(3); p. 369--389.

\bibitem{Cuevas2012}
Cuevas E, González M, Zaldivar D, Pérez-Cisneros M, García G.
\newblock An algorithm for global optimization inspired by collective animal
  behavior.
\newblock Discrete Dynamics in Nature and Society. 2012;2012; p. 638275.

\bibitem{Klein2018685}
Klein CE, Mariani VC, Coelho LDS.
\newblock Cheetah based optimization algorithm: A novel swarm intelligence
  paradigm.
\newblock In: ESANN 2018 - Proceedings, European Symposium on Artificial Neural
  Networks, Computational Intelligence and Machine Learning; 2018. p. 685--690.

\bibitem{CAO}
Shiqin Y, Jianjun J, Guangxing Y.
\newblock Improved binary particle swarm optimization using catfish effect for
  feature selection.
\newblock Expert Systems with Applications. 2011;38(10); p. 12699--12707.

\bibitem{Canayaz2016362}
Canayaz M, Karci A.
\newblock Cricket behaviour-based evolutionary computation technique in solving
  engineering optimization problems.
\newblock Applied Intelligence. 2016;44(2); p. 362--376.

\bibitem{Pierezan2019}
Pierezan J, Maidl G, Massashi~Yamao E, dos Santos~Coelho L, Cocco~Mariani V.
\newblock Cultural coyote optimization algorithm applied to a heavy duty gas
  turbine operation.
\newblock Energy Conversion and Management. 2019;199; p. 111932.

\bibitem{CCSA}
Rizk-Allah RM, Hassanien AE, Bhattacharyya S.
\newblock Chaotic crow search algorithm for fractional optimization problems.
\newblock Applied Soft Computing. 2018;71; p. 1161--1175.

\bibitem{CDA}
Sayed GI, Tharwat A, Hassanien AE.
\newblock Chaotic dragonfly algorithm: an improved metaheuristic algorithm for
  feature selection.
\newblock Applied Intelligence. 2019;49; p. 188--205.

\bibitem{CFA}
Eesa~Sabry A, Adbulazeez~Brifcani AM, Orman Z.
\newblock Cuttlefish Algorithm – A Novel Bio-Inspired Optimization Algorithm.
\newblock International Journal of Scientific and Engineering Research.
  2013;4(9); p. 1978--1986.

\bibitem{CGS}
Iordache S.
\newblock A Hierarchical Cooperative Evolutionary Algorithm.
\newblock In: Proceedings of the 12th Annual Conference on Genetic and
  Evolutionary Computation; 2010. p. 225--232.

\bibitem{camelherd}
Al-Obaidi A, Abdullah H, Othman Z.
\newblock Camel Herds Algorithm: a New Swarm Intelligent Algorithm to Solve
  Optimization Problems.
\newblock International Journal on Perceptive and Cognitive Computing.
  2017;3(1); p. 1--5.

\bibitem{Khishe2020}
Khishe M, Mosavi MR.
\newblock Chimp optimization algorithm.
\newblock Expert Systems with Applications. 2020;149; p. 113338.

\bibitem{COA}
Rajabioun R.
\newblock Cuckoo Optimization Algorithm.
\newblock Applied Soft Computing. 2011;11(8); p. 5508--5518.

\bibitem{COA2}
Ibrahim MK, Salim~Ali R.
\newblock Novel Optimization Algorithm Inspired by Camel Traveling Behavior.
\newblock Iraq Journal Electrical and Electronic Engineering. 2016;12(2); p.
  167--177.

\bibitem{Coyote2018}
{Pierezan} J, {Dos Santos Coelho} L.
\newblock Coyote Optimization Algorithm: A New Metaheuristic for Global
  Optimization Problems.
\newblock In: 2018 IEEE Congress on Evolutionary Computation (CEC); 2018. p.
  1--8.

\bibitem{Naruei2021}
Naruei I, Keynia F.
\newblock A new optimization method based on COOT bird natural life model.
\newblock Expert Systems with Applications. 2021;183; p. 115352.

\bibitem{Dehghani2023}
Dehghani M, Montazeri Z, Trojovská E, Trojovsk{\`y} P.
\newblock Coati Optimization Algorithm: A new bio-inspired metaheuristic
  algorithm for solving optimization problems.
\newblock Knowledge-Based Systems. 2023;259; p. 110011.

\bibitem{AbdelBasset2024}
Abdel-Basset M, Mohamed R, Abouhawwash M.
\newblock Crested {P}orcupine {O}ptimizer: A new nature-inspired metaheuristic.
\newblock Knowledge-Based Systems. 2024;284; p. 111257.

\bibitem{CS}
Yang X, Deb S.
\newblock Cuckoo Search via Lévy flights.
\newblock In: 2009 World Congress on Nature Biologically Inspired Computing
  (NaBIC); 2009. p. 210--214.

\bibitem{Askarzadeh20161}
Askarzadeh A.
\newblock A novel metaheuristic method for solving constrained engineering
  optimization problems: Crow search algorithm.
\newblock Computers and Structures. 2016;169; p. 1--12.

\bibitem{Qais2022}
Qais MH, Hasanien HM, Turky RA, Alghuwainem S, Tostado-V{\'e}liz M, Jurado F.
\newblock Circle search algorithm: A geometry-based metaheuristic optimization
  algorithm.
\newblock Mathematics. 2022;10(10); p. 1626.

\bibitem{CSO}
Chu SC, Tsai Pw, Pan JS.
\newblock Cat Swarm Optimization.
\newblock In: PRICAI 2006: Trends in Artificial Intelligence; 2006. p.
  854--858.

\bibitem{CSO1}
Meng X, Liu Y, Gao X, Zhang H.
\newblock A New Bio-inspired Algorithm: Chicken Swarm Optimization.
\newblock In: Advances in Swarm Intelligence; 2014. p. 86--94.

\bibitem{DA}
Mirjalili S.
\newblock Dragonfly algorithm: a new meta-heuristic optimization technique for
  solving single-objective, discrete, and multi-objective problems.
\newblock Neural Computing and Applications. 2016;27(4); p. 1053--1073.

\bibitem{bhardwaj2020dragonfly}
Bhardwaj S, Kim DS.
\newblock Dragonfly-based swarm system model for node identification in
  ultra-reliable low-latency communication.
\newblock Neural Computing and Applications. 2020;33; p. 1837--1880.

\bibitem{DE1}
Kaveh A, Farhoudi N.
\newblock A new optimization method: Dolphin echolocation.
\newblock Advances in Engineering Software. 2013;59; p. 53--70.

\bibitem{Ahmadi2023}
Ahmadi B, Giraldo JS, Hoogsteen G.
\newblock Dynamic {H}unting {L}eadership optimization: Algorithm and
  applications.
\newblock Journal of Computational Science. 2023;69; p. 102010.

\bibitem{DHOA}
Brammya G, Praveena S, Ninu~Preetha NS, Ramya R, Rajakumar BR, Binu D.
\newblock {Deer Hunting Optimization Algorithm: A New Nature-Inspired
  Meta-heuristic Paradigm}.
\newblock The Computer Journal. 2019;p. 1--20.

\bibitem{Agushaka2022}
Agushaka JO, Ezugwu AE, Abualigah L.
\newblock Dwarf {M}ongoose {O}ptimization {A}lgorithm.
\newblock Computer Methods in Applied Mechanics and Engineering. 2022;391; p.
  114570.

\bibitem{Zhao2022dandelion}
Zhao S, Zhang T, Ma S, Chen M.
\newblock Dandelion Optimizer: A nature-inspired metaheuristic algorithm for
  engineering applications.
\newblock Engineering Applications of Artificial Intelligence. 2022;114; p.
  105075.

\bibitem{Bairwa2021}
Bairwa AK, Joshi S, Singh D.
\newblock Dingo optimizer: a nature-inspired metaheuristic approach for
  engineering problems.
\newblock Mathematical Problems in Engineering. 2021;2021; p. 2571863.

\bibitem{DPO}
{Shiqin} Y, {Jianjun} J, {Guangxing} Y.
\newblock A Dolphin Partner Optimization.
\newblock In: 2009 WRI Global Congress on Intelligent Systems. vol.~1; 2009. p.
  124--128.

\bibitem{DTO}
Dehghani M, Mardaneh M, Malik OP, Nouraei~Pour SM.
\newblock DTO: Donkey Theorem Optimization.
\newblock In: 2019 27th Iranian Conference on Electrical Engineering (ICEE);
  2019. p. 1855--1859.

\bibitem{Kusuma2024}
Kusuma PD, Hasibuan FC.
\newblock Enriched {C}oati {O}sprey {A}lgorithm: A {S}warm-based
  {M}etaheuristic and {I}ts {S}ensitivity {E}valuation of {I}ts {S}trategy.
\newblock IAENG International Journal of Applied Mathematics. 2024;54(2).

\bibitem{Zhao2024}
Zhao W, Wang L, Zhang Z, Fan H, Zhang J, Mirjalili S, et~al.
\newblock Electric eel foraging optimization: A new bio-inspired optimizer for
  engineering applications.
\newblock Expert Systems with Applications. 2024;238; p. 122200.

\bibitem{Yilmaz2020}
Yilmaz S, Sen S.
\newblock Electric fish optimization: a new heuristic algorithm inspired by
  electrolocation.
\newblock Neural Computing and Applications. 2020;32(15); p. 11543--11578.

\bibitem{Wang20161}
{Wang} G, {Deb} S, d~S~{Coelho} L.
\newblock Elephant Herding Optimization.
\newblock In: 2015 3rd International Symposium on Computational and Business
  Intelligence (ISCBI); 2015. p. 1--5.

\bibitem{AlBetar2024}
Al-Betar MA, Awadallah MA, Braik MS, Makhadmeh S, Doush IA.
\newblock Elk herd optimizer: a novel nature-inspired metaheuristic algorithm.
\newblock Artificial Intelligence Review. 2024;57(3); p. 48.

\bibitem{Oyelade2022}
Oyelade ON, Ezugwu AES, Mohamed TIA, Abualigah L.
\newblock Ebola Optimization Search Algorithm: A New Nature-Inspired
  Metaheuristic Optimization Algorithm.
\newblock IEEE Access. 2022;10; p. 16150--16177.

\bibitem{penguinemperor}
Harifi S, Khalilian M, Mohammadzadeh J, Ebrahimnejad S.
\newblock Emperor Penguins Colony: a new metaheuristic algorithm for
  optimization.
\newblock Evolutionary Intelligence. 2019;12; p. 211--226.

\bibitem{EPO}
Dhiman G, Kumar V.
\newblock Emperor penguin optimizer: a bio-inspired algorithm for engineering
  problems.
\newblock Knowledge-Based Systems. 2018;159; p. 20--50.

\bibitem{ES}
Yang XS, Deb S.
\newblock Eagle Strategy Using L{\'e}vy Walk and Firefly Algorithms for
  Stochastic Optimization.
\newblock In: Nature Inspired Cooperative Strategies for Optimization (NICSO
  2010). Springer Berlin Heidelberg; 2010. p. 101--111.

\bibitem{ESA}
Deb S, Fong S, Tian Z.
\newblock Elephant Search Algorithm for optimization problems.
\newblock In: 2015 Tenth International Conference on Digital Information
  Management (ICDIM); 2015. p. 249--255.

\bibitem{elephantdrought}
Mandal S.
\newblock Elephant swarm water search algorithm for global optimization.
\newblock Sādhanā. 2018;43; p. 1--21.

\bibitem{EV}
Sur C, Sharma S, Shukla A.
\newblock Egyptian Vulture Optimization Algorithm -- A New Nature Inspired
  Meta-heuristics for Knapsack Problem.
\newblock In: The 9th International Conference on Computing and
  InformationTechnology (IC2IT2013); 2013. p. 227--237.

\bibitem{Cui2006505}
Cui X, Gao J, Potok TE.
\newblock A flocking based algorithm for document clustering analysis.
\newblock Journal of Systems Architecture. 2006;52(8-9); p. 505--515.

\bibitem{FBSA}
Chu Y, Mi H, Liao H, Ji Z, {Wu} QH.
\newblock A Fast Bacterial Swarming Algorithm for high-dimensional function
  optimization.
\newblock In: 2008 IEEE Congress on Evolutionary Computation (IEEE World
  Congress on Computational Intelligence); 2008. p. 3135--3140.

\bibitem{Mutazono5227977}
{Mutazono} A, {Sugano} M, {Murata} M.
\newblock Frog call-inspired self-organizing anti-phase synchronization for
  wireless sensor networks.
\newblock In: 2009 2nd International Workshop on Nonlinear Dynamics and
  Synchronization; 2009. p. 81--88.

\bibitem{Azizi2023fire}
Azizi M, Talatahari S, Gandomi AH.
\newblock Fire {H}awk {O}ptimizer: A novel metaheuristic algorithm.
\newblock Artificial Intelligence Review. 2023;56; p. 287--363.

\bibitem{Bellaachia2012117}
Bellaachia A, Bari A.
\newblock Flock by Leader: A Novel Machine Learning Biologically Inspired
  Clustering Algorithm.
\newblock In: Advances in Swarm Intelligence; 2012. p. 117--126.

\bibitem{Falahah2024}
Falahah IA, Al-Baik O, Alomari S, Bektemyssova G, Gochhait S, Leonova I, et~al.
\newblock Frilled {L}izard {O}ptimization: A {N}ovel {N}ature-{I}nspired
  {M}etaheuristic {A}lgorithm for {S}olving {O}ptimization {P}roblems.
\newblock Preprints. 2024;p. 1--44.

\bibitem{FOA}
Pan WT.
\newblock A new Fruit Fly Optimization Algorithm: Taking the financial distress
  model as an example.
\newblock Knowledge-Based Systems. 2012;26; p. 69--74.

\bibitem{FOA2}
de~{Vasconcelos Segundo} EH, Mariani VC, dos Santos~Coelho L.
\newblock Design of heat exchangers using Falcon Optimization Algorithm.
\newblock Applied Thermal Engineering. 2019;156; p. 119--144.

\bibitem{Mohammed2023fox}
Mohammed H, Rashid T.
\newblock {{FOX}}: a {{FOX}}-inspired optimization algorithm.
\newblock Applied Intelligence. 2023;53; p. 1030--1050.

\bibitem{Li200232}
Li XL, Shao ZJ, Qian JX.
\newblock Optimizing method based on autonomous animats: Fish-swarm Algorithm.
\newblock System Engineering Theory and Practice. 2002;22(11); p. 32--38.

\bibitem{FSA}
Tsai HC, Lin YH.
\newblock Modification of the fish swarm algorithm with particle swarm
  optimization formulation and communication behavior.
\newblock Applied Soft Computing. 2011;11(8); p. 5367 -- 5374.

\bibitem{FSS}
{Bastos Filho} CJA, {de Lima Neto} FB, {Lins} AJCC, {Nascimento} AIS, {Lima}
  MP.
\newblock A novel search algorithm based on fish school behavior.
\newblock In: 2008 IEEE International Conference on Systems, Man and
  Cybernetics; 2008. p. 2646--2651.

\bibitem{Dehghani2023green}
Dehghani M, Trojovsk{\`y} P, Malik OP.
\newblock Green Anaconda Optimization: A New Bio-Inspired Metaheuristic
  Algorithm for Solving Optimization Problems.
\newblock Biomimetics. 2023;8(1); p. 121.

\bibitem{Alsayyed2023}
Alsayyed O, Hamadneh T, Al-Tarawneh H, Alqudah M, Gochhait S, Leonova I, et~al.
\newblock Giant {A}rmadillo {O}ptimization: A {N}ew {B}io-{I}nspired
  {M}etaheuristic {A}lgorithm for {S}olving {O}ptimization {P}roblems.
\newblock Biomimetics. 2023;8(8); p. 619.

\bibitem{Min20116128}
{Min} H, {Wang} Z.
\newblock Design and analysis of Group Escape Behavior for distributed
  autonomous mobile robots.
\newblock In: 2011 IEEE International Conference on Robotics and Automation;
  2011. p. 6128--6135.

\bibitem{MohammadiBalani2021}
Mohammadi-Balani A, {Dehghan Nayeri} M, Azar A, Taghizadeh-Yazdi M.
\newblock Golden eagle optimizer: A nature-inspired metaheuristic algorithm.
\newblock Computers \& Industrial Engineering. 2021;152; p. 107050.

\bibitem{Chopra2023}
Chopra N, {Mohsin Ansari} M.
\newblock Golden jackal optimization: A novel nature-inspired optimizer for
  engineering applications.
\newblock Expert Systems with Applications. 2022;198; p. 116924.

\bibitem{HuGang2023}
Hu G, Guo Y, Wei G, Abualigah L.
\newblock Genghis {K}han shark optimizer: {A} novel nature-inspired algorithm
  for engineering optimization.
\newblock Advanced Engineering Informatics. 2023;58; p. 102210.

\bibitem{GLSO}
{Su} S, {Wang} J, {Fan} W, {Yin} X.
\newblock Good Lattice Swarm Algorithm for Constrained Engineering Design
  Optimization.
\newblock In: 2007 International Conference on Wireless Communications,
  Networking and Mobile Computing; 2007. p. 6421--6424.

\bibitem{Agushaka2023}
Agushaka JO, Ezugwu AE, Abualigah L.
\newblock Gazelle optimization algorithm: a novel nature-inspired metaheuristic
  optimizer.
\newblock Neural Computing and Applications. 2023;35; p. 4099--4131.

\bibitem{De2022}
De SK.
\newblock The goat search algorithms.
\newblock Artificial Intelligence Review. 2022;(56); p. 8265–8301.

\bibitem{GSO}
Zhou Y, Luo Q, Liu J.
\newblock Glowworm Swarm Optimization For Optimization Dispatching System Of
  Public Transit Vehicles.
\newblock Journal of Theoretical and Applied Information Technology. 2013;52;
  p. 205--210.

\bibitem{GSO1}
{He} S, {Wu} QH, {Saunders} JR.
\newblock Group Search Optimizer: An Optimization Algorithm Inspired by Animal
  Searching Behavior.
\newblock IEEE Transactions on Evolutionary Computation. 2009;13(5); p.
  973--990.

\bibitem{Wang20081437}
Wang J, Wang D.
\newblock Particle swarm optimization with a leader and followers.
\newblock Progress in Natural Science. 2008;18(11); p. 1437--1443.

\bibitem{Abdollahzadeh2021}
Abdollahzadeh B, Soleimanian~Gharehchopogh F, Mirjalili S.
\newblock Artificial gorilla troops optimizer: A new nature-inspired
  metaheuristic algorithm for global optimization problems.
\newblock International Journal of Intelligent Systems. 2021;36(10); p.
  5887--5958.

\bibitem{GWO}
Mirjalili S, Mirjalili SM, Lewis A.
\newblock Grey Wolf Optimizer.
\newblock Advances in Engineering Software. 2014;69; p. 46--61.

\bibitem{HBIA}
Morais RG, Nedjah N, Mourelle LM.
\newblock A novel metaheuristic inspired by Hitchcock birds’ behavior for
  efficient optimization of large search spaces of high dimensionality.
\newblock Soft Computing. 2020;24(8); p. 5633--5655.

\bibitem{BOZORG}
Bozorg-Haddad O, Afshar A, Mariño M.
\newblock Honey-Bees Mating Optimization (HBMO) Algorithm: A New Heuristic
  Approach for Water Resources Optimization.
\newblock Water Resources Management. 2006;20; p. 661--680.

\bibitem{YangYutao2021}
Yang Y, Chen H, Heidari AA, Gandomi AH.
\newblock Hunger games search: {V}isions, conception, implementation, deep
  analysis, perspectives, and towards performance shifts.
\newblock Expert Systems with Applications. 2021;177; p. 114864.

\bibitem{HEIDARI2019}
Heidari AA, Mirjalili S, Faris H, Aljarah I, Mafarja M, Chen H.
\newblock Harris hawks optimization: Algorithm and applications.
\newblock Future Generation Computer Systems. 2019;97; p. 849--872.

\bibitem{HHO}
El-Dosuky M, El-Bassiouny A, Hamza T, Rashad M.
\newblock New Hoopoe Heuristic Optimization.
\newblock International Journal of Science and Advanced Technology. 2012;2(9);
  p. 85--90.

\bibitem{Peraza2024}
Peraza-V{\'a}zquez H, Pe{\~n}a-Delgado A, Merino-Trevi{\~n}o M, Morales-Cepeda
  AB, Sinha N.
\newblock A novel metaheuristic inspired by horned lizard defense tactics.
\newblock Artificial Intelligence Review. 2024;57(3); p. 1--59.

\bibitem{HOA}
Moldovan D.
\newblock Horse Optimization Algorithm: A Novel Bio-Inspired Algorithm for
  Solving Global Optimization Problems.
\newblock In: Artificial Intelligence and Bioinspired Computational Methods;
  2020. p. 195--209.

\bibitem{HuS}
Oftadeh R, Mahjoob MJ, Shariatpanahi M.
\newblock A novel meta-heuristic optimization algorithm inspired by group
  hunting of animals: Hunting search.
\newblock Computers and Mathematics with Applications. 2010;60(7); p.
  2087--2098.

\bibitem{Quijano20073383}
{Quijano} N, {Passino} KM.
\newblock Honey Bee Social Foraging Algorithms for Resource Allocation, Part I:
  Algorithm and Theory.
\newblock In: 2007 American Control Conference; 2007. p. 3383--3388.

\bibitem{HSM}
Chen H, Zhu Y, Hu K, He X.
\newblock Hierarchical Swarm Model: A New Approach to Optimization.
\newblock Discrete Dynamics in Nature and Society. 2010;2010; p. 379649.

\bibitem{HSOA}
Ali A, Zafar K, Bakhshi T.
\newblock On Nature-Inspired Dynamic Route Planning: Hammerhead Shark
  Optimization Algorithm.
\newblock In: 2019 15th International Conference on Emerging Technologies
  (ICET); 2019. p. 1--6.

\bibitem{Anaraki2023}
Anaraki MV, Farzin S.
\newblock Humboldt {S}quid {O}ptimization {A}lgorithm {{(HSOA)}}: A {N}ovel
  {N}ature-{I}nspired {T}echnique for {S}olving {O}ptimization {P}roblems.
\newblock IEEE Access. 2023;11; p. 122069--122115.

\bibitem{Maciel2019}
Maciel O, Valdivia A, Oliva D, Cuevas E, Zald{\'i}var D, P{\'e}rez-Cisneros M.
\newblock A novel hybrid metaheuristic optimization method: hypercube natural
  aggregation algorithm.
\newblock Soft Computing. 2020;(24); p. 8823–8856.

\bibitem{TORABI2018144}
Torabi S, Safi-Esfahani F.
\newblock Improved Raven Roosting Optimization algorithm (IRRO).
\newblock Swarm and Evolutionary Computation. 2018;40; p. 144--154.

\bibitem{ITGO}
Tang D, Dong S, Jiang Y, Li H, Huang Y.
\newblock ITGO: Invasive tumor growth optimization algorithm.
\newblock Applied Soft Computing. 2015;36; p. 670--698.

\bibitem{JA}
{Chen} C, {Tsai} Y, {Liu} I, {Lai} C, {Yeh} Y, {Kuo} S, et~al.
\newblock A Novel Metaheuristic: Jaguar Algorithm with Learning Behavior.
\newblock In: 2015 IEEE International Conference on Systems, Man, and
  Cybernetics; 2015. p. 1595--1600.

\bibitem{Chou2021}
Chou JS, Truong DN.
\newblock A novel metaheuristic optimizer inspired by behavior of jellyfish in
  ocean.
\newblock Applied Mathematics and Computation. 2021;389; p. 125535.

\bibitem{hernandez2012distributed}
Hern{\'a}ndez H, Blum C.
\newblock Distributed graph coloring: an approach based on the calling behavior
  of Japanese tree frogs.
\newblock Swarm Intelligence. 2012;6(2); p. 117--150.

\bibitem{KH}
Hossein~Gandomi A, Hossein~Alavi A.
\newblock Krill herd: A new bio-inspired optimization algorithm.
\newblock Communications in Nonlinear Science and Numerical Simulation.
  2012;17(12); p. 4831--4845.

\bibitem{Dehghani2023c}
Dehghani M, Montazeri Z, Bektemyssova G, Malik OP, Dhiman G, Ahmed AEM.
\newblock Kookaburra {O}ptimization {A}lgorithm: A {N}ew {B}io-{I}nspired
  {M}etaheuristic {A}lgorithm for {S}olving {O}ptimization {P}roblems.
\newblock Biomimetics. 2023;8(6); p. 470.

\bibitem{KSA}
Agbehadji IE, Millham R, Fong S.
\newblock Kestrel-Based Search Algorithm for Association Rule Mining and
  Classification of Frequently Changed Items.
\newblock In: 2016 8th International Conference on Computational Intelligence
  and Communication Networks (CICN); 2016. p. 356--360.

\bibitem{biyanto2017killer}
Biyanto TR, Irawan S, Febrianto HY, Afdanny N, Rahman AH, Gunawan KS, et~al.
\newblock Killer whale algorithm: an algorithm inspired by the life of killer
  whale.
\newblock Procedia Computer Science. 2017;124; p. 151--157.

\bibitem{LA}
Rajakumar BR.
\newblock The Lion's Algorithm: A New Nature-Inspired Search Algorithm.
\newblock Procedia Technology. 2012;6; p. 126--135.

\bibitem{LBO}
Wang P, Zhu Z, Huang S.
\newblock Seven-Spot Ladybird Optimization: A Novel and Efficient Metaheuristic
  Algorithm for Numerical Optimization.
\newblock The Scientific World Journal. 2013;2013; p. 378515.

\bibitem{Dehghani2023b}
Dehghani M, Bektemyssova G, Montazeri Z, Shaikemelev G, Malik OP, Dhiman G.
\newblock Lyrebird {O}ptimization {A}lgorithm: A {N}ew {B}io-{I}nspired
  {M}etaheuristic {A}lgorithm for {S}olving {O}ptimization {P}roblems.
\newblock Biomimetics. 2023;8(6); p. 507.

\bibitem{hosseini2017laying}
Hosseini E.
\newblock Laying chicken algorithm: A new meta-heuristic approach to solve
  continuous programming problems.
\newblock Journal of Applied \& Computational Mathematics. 2017;6(1); p.
  344--351.

\bibitem{LOA}
Yazdani M, Jolai F.
\newblock Lion Optimization Algorithm (LOA): A nature-inspired metaheuristic
  algorithm.
\newblock Journal of Computational Design and Engineering. 2016;3(1); p.
  24--36.

\bibitem{LPO}
Wang B, Jin X, Cheng B.
\newblock Lion pride optimizer: An optimization algorithm inspired by lion
  pride behavior.
\newblock Science China Information Sciences. 2012;55(10); p. 2369--2389.

\bibitem{LSO}
Chen S.
\newblock An Analysis of Locust Swarms on Large Scale Global Optimization
  Problems.
\newblock In: Artificial Life: Borrowing from Biology; 2009. p. 211--220.

\bibitem{Rabie2023LSO}
Rabie AH, Mansour NA, Saleh AI.
\newblock Leopard seal optimization (LSO): A natural inspired meta-heuristic
  algorithm.
\newblock Communications in Nonlinear Science and Numerical Simulation.
  2023;125; p. 107338.

\bibitem{locuscuevas}
Cuevas E, Gonzalez A, Zaldivar D, Cisneros M.
\newblock An optimisation algorithm based on the behaviour of locust swarms.
\newblock International Journal of Bio-Inspired Computation. 2015;7; p.
  402--407.

\bibitem{MA1}
Zervoudakis K, Tsafarakis S.
\newblock A mayfly optimization algorithm.
\newblock Computers \& Industrial Engineering. 2020;145; p. 106559.

\bibitem{MBO3}
{Mo} H, {Xu} L.
\newblock Magnetotactic bacteria optimization algorithm for multimodal
  optimization.
\newblock In: 2013 IEEE Symposium on Swarm Intelligence (SIS); 2013. p.
  240--247.

\bibitem{MBO}
Wang GG, Deb S, Cui Z.
\newblock Monarch butterfly optimization.
\newblock Neural Computing and Applications. 2015;(31); p. 1995--2014.

\bibitem{MBO1}
Duman E, Uysal M, Alkaya AF.
\newblock Migrating Birds Optimization: A new metaheuristic approach and its
  performance on quadratic assignment problem.
\newblock Information Sciences. 2012;217; p. 65--77.

\bibitem{JAHANI2018987}
Jahani E, Chizari M.
\newblock Tackling global optimization problems with a novel algorithm –
  Mouth Brooding Fish algorithm.
\newblock Applied Soft Computing Journal. 2018;62; p. 987--1002.

\bibitem{Kusuma2024migration}
Kusuma PD, Kallista M.
\newblock Migration-{C}rossover {A}lgorithm: A {S}warm-based {M}etaheuristic
  {E}nriched with {C}rossover {T}echnique and {U}nbalanced {N}eighbourhood
  {S}earch.
\newblock International Journal of Intelligent Engineering \& Systems.
  2024;17(1).

\bibitem{MCS}
Walton S, Hassan O, Morgan K, Brown MR.
\newblock Modified cuckoo search: A new gradient free optimisation algorithm.
\newblock Chaos, Solitons and Fractals. 2011;44(9); p. 710--718.

\bibitem{MCSO}
Obagbuwa IC, Adewumi AO.
\newblock An Improved Cockroach Swarm Optimization.
\newblock ScientificWorld Journal. 2014;p. 375358.

\bibitem{MFO}
Mirjalili S.
\newblock Moth-flame optimization algorithm: A novel nature-inspired heuristic
  paradigm.
\newblock Knowledge-Based Systems. 2015;89; p. 228--249.

\bibitem{alauddin2016mosquito}
Alauddin M.
\newblock Mosquito flying optimization (MFO).
\newblock In: 2016 International Conference on Electrical, Electronics, and
  Optimization Techniques (ICEEOT); 2016. p. 79--84.

\bibitem{klein2018meerkats}
Klein CE, Coelho LDS.
\newblock Meerkats-inspired algorithm for global optimization problems; 2018.
  p. 679--684.

\bibitem{Valdez2023}
Valdez F, Carreon-Ortiz H, Castillo O.
\newblock Introduction to the {M}ycorrhiza {O}ptimization {A}lgorithm.
\newblock In: Mycorrhiza Optimization Algorithm. Springer Nature Switzerland;
  2023. p. 1--84.

\bibitem{MINHAS20114614}
ul~Amir Afsar~Minhas F, Arif M.
\newblock MOX: A novel global optimization algorithm inspired from Oviposition
  site selection and egg hatching inhibition in mosquitoes.
\newblock Applied Soft Computing. 2011;11(8); p. 4614--4625.

\bibitem{MPA}
Faramarzi A, Heidarinejad M, Mirjalili S, Gandomi AH.
\newblock Marine Predators Algorithm: A nature-inspired metaheuristic.
\newblock Expert Systems with Applications. 2020;152; p. 113377.

\bibitem{MS}
Mucherino A, Seref O.
\newblock Monkey search: a novel metaheuristic search for global optimization.
\newblock In: American Institute of Physics. vol. 953; 2007. p. 162--173.

\bibitem{Wang2018}
Wang GG.
\newblock Moth search algorithm: a bio-inspired metaheuristic algorithm for
  global optimization problems.
\newblock Memetic Computing. 2018;10(2); p. 151--164.

\bibitem{AbdelBasset2023}
Abdel-Basset M, Mohamed R, Zidan M, Jameel M, Abouhawwash M.
\newblock Mantis {S}earch {A}lgorithm: A novel bio-inspired algorithm for
  global optimization and engineering design problems.
\newblock Computer Methods in Applied Mechanics and Engineering. 2023;415; p.
  116200.

\bibitem{Luo201694}
{Luo} F, {Zhao} J, {Dong} ZY.
\newblock A new metaheuristic algorithm for real-parameter optimization:
  Natural aggregation algorithm.
\newblock In: 2016 IEEE Congress on Evolutionary Computation (CEC); 2016. p.
  94--103.

\bibitem{Salgotra2019}
Salgotra R, Singh U.
\newblock The naked mole-rat algorithm.
\newblock Neural Computing and Applications. 2019;(31); p. 8837–--8857.

\bibitem{AbdelBasset2023b}
Abdel-Basset M, Mohamed R, Jameel M, Abouhawwash M.
\newblock Nutcracker optimizer: A novel nature-inspired metaheuristic algorithm
  for global optimization and engineering design problems.
\newblock Knowledge-Based Systems. 2023;262; p. 110248.

\bibitem{Salih2019}
Salih SQ, Alsewari AA.
\newblock A new algorithm for normal and large-scale optimization problems:
  Nomadic People Optimizer.
\newblock Neural Computing and Applications. 2020;(32); p. 10359–10386.

\bibitem{OA}
Drias H, Drias Y, Khennak I.
\newblock A New Swarm Algorithm Based on Orcas Intelligence for Solving Maze
  Problems.
\newblock In: Trends and Innovations in Information Systems and Technologies;
  2020. p. 788--797.

\bibitem{OB}
{Maia} RD, d~{Castro} LN, {Caminhas} WM.
\newblock OptBees - A Bee-Inspired Algorithm for Solving Continuous
  Optimization Problems.
\newblock In: 2013 BRICS Congress on Computational Intelligence and 11th
  Brazilian Congress on Computational Intelligence; 2013. p. 142--151.

\bibitem{OFA}
Zhu GY, Zhang WB.
\newblock Optimal foraging algorithm for global optimization.
\newblock Applied Soft Computing. 2017;51; p. 294--313.

\bibitem{OOA}
de~{Vasconcelos Segundo} EH, Mariani VC, dos Santos~Coelho L.
\newblock Metaheuristic inspired on owls behavior applied to heat exchangers
  design.
\newblock Thermal Science and Engineering Progress. 2019;14; p. 100431.

\bibitem{Dehghani2023osprey}
Dehghani M, Trojovsk{\`y} P.
\newblock Osprey optimization algorithm: A new bio-inspired metaheuristic
  algorithm for solving engineering optimization problems.
\newblock Frontiers in Mechanical Engineering. 2023;8; p. 1126450.

\bibitem{Jiang2022}
Jiang Y, Wu Q, Zhu S, Zhang L.
\newblock Orca predation algorithm: A novel bio-inspired algorithm for global
  optimization problems.
\newblock Expert Systems with Applications. 2022;188; p. 116026.

\bibitem{Kallioras2018147}
Kallioras NA, Lagaros ND, Avtzis DN.
\newblock Pity beetle algorithm – A new metaheuristic inspired by the
  behavior of bark beetles.
\newblock Advances in Engineering Software. 2018;121; p. 147--166.

\bibitem{PBOA}
Połap D, Wozniak M.
\newblock Polar Bear Optimization Algorithm: Meta-Heuristic with Fast
  Population Movement and Dynamic Birth and Death Mechanism.
\newblock Symmetry. 2017;9(203); p. 1--20.

\bibitem{Awad2023novel}
Awad A, Coghill GM, Pang W.
\newblock A novel {P}hysarum-inspired competition algorithm for discrete
  multi-objective optimisation problems.
\newblock Soft Computing. 2023;p. 1--21.

\bibitem{Ezugu2022}
Ezugwu AE, Agushaka JO, Abualigah L, Mirjalili S, Gandomi AH.
\newblock Prairie {D}og {O}ptimization {A}lgorithm.
\newblock Neural Computing and Applications. 2022;34(22); p. 20017–20065.

\bibitem{duan2014pigeon}
Duan H, Qiao P.
\newblock Pigeon-inspired optimization: a new swarm intelligence optimizer for
  air robot path planning.
\newblock International Journal of Intelligent Computing and Cybernetics.
  2014;7(1); p. 24--37.

\bibitem{Zhang2009165}
Zhang W, Luo Q, Zhou Y.
\newblock A Method for Training RBF Neural Networks Based on Population
  Migration Algorithm.
\newblock In: Proceedings of the 2009 International Conference on Artificial
  Intelligence and Computational Intelligence. vol.~1; 2009. p. 165--169.

\bibitem{Abdollahzadeh2024}
Abdollahzadeh B, Khodadadi N, Barshandeh S, Trojovsk{\`y} P, Gharehchopogh FS,
  El-kenawy ESM, et~al.
\newblock Puma optimizer {{(PO)}}: A novel metaheuristic optimization algorithm
  and its application in machine learning.
\newblock Cluster Computing. 2024;p. 1--49.

\bibitem{Trojovsky2023b}
Trojovsk{\`y} P, Dehghani M.
\newblock Pelican {O}ptimization {A}lgorithm: A {N}ovel {N}ature-{I}nspired
  {A}lgorithm for {E}ngineering {A}pplications.
\newblock Sensors. 2022;22(3); p. 855.

\bibitem{Albaik2024}
Al-Baik O, Alomari S, Alssayed O, Gochhait S, Leonova I, Dutta U, et~al.
\newblock Pufferfish {O}ptimization {A}lgorithm: A {N}ew {B}io-{I}nspired
  {M}etaheuristic {A}lgorithm for {S}olving {O}ptimization {P}roblems.
\newblock Biomimetics. 2024;9(2).

\bibitem{PPA}
Tilahun SL, Choon~Ong H.
\newblock Prey-predator algorithm: A new metaheuristic algorithm for
  optimization problems.
\newblock International Journal of Information Technology and Decision Making.
  2015;14(6); p. 1331--1352.

\bibitem{PSOA}
Gheraibia Y, Moussaoui A.
\newblock Penguins Search Optimization Algorithm (PeSOA).
\newblock In: Recent Trends in Applied Artificial Intelligence; 2013. p.
  222--231.

\bibitem{Arora2019715}
Arora S, Singh S.
\newblock Butterfly optimization algorithm: a novel approach for global
  optimization.
\newblock Soft Computing. 2019;23(3); p. 715--734.

\bibitem{fard2016red}
Fard AF, Hajiaghaei-Keshteli M.
\newblock Red Deer Algorithm (RDA); a new optimization algorithm inspired by
  Red Deers’ mating.
\newblock In: International Conference on Industrial Engineering, IEEE.,(2016
  e); 2016. p. 33--34.

\bibitem{RFO}
Połap D, Woźniak M.
\newblock Red fox optimization algorithm.
\newblock Expert Systems with Applications. 2021;166; p. 114107.

\bibitem{wang2018novel}
Wang GG, Gao XZ, Zenger K, Coelho LdS.
\newblock A novel metaheuristic algorithm inspired by rhino herd behavior.
\newblock In: Proceedings of The 9th EUROSIM Congress on Modelling and
  Simulation, EUROSIM 2016, The 57th SIMS Conference on Simulation and
  Modelling SIMS 2016; 2018. p. 1026--1033.

\bibitem{Belal2021}
Khateeb BA, Ahmed K, Mahmood M, Le DN.
\newblock Rock {H}yraxes {S}warm {O}ptimization: A {N}ew {N}ature-{I}nspired
  {M}etaheuristic {O}ptimization {A}lgorithm.
\newblock Computers, Materials \& Continua. 2021;68(1); p. 643--654.

\bibitem{RIO}
Havens T, J~Spain C, G~Salmon N, M~Keller J.
\newblock Roach Infestation Optimization.
\newblock In: 2008 IEEE Swarm Intelligence Symposium, SIS 2008; 2008. p. 1--7.

\bibitem{raccoonoptimization}
Zangbari~Koohi S, Abdul~Hamid NAW, Othman M, Ibragimov G.
\newblock Raccoon Optimization Algorithm.
\newblock IEEE Access. 2019;7; p. 5383--5399.

\bibitem{ROA}
{Sharma} A.
\newblock A new optimizing algorithm using reincarnation concept.
\newblock In: 2010 11th International Symposium on Computational Intelligence
  and Informatics (CINTI); 2010. p. 281--288.

\bibitem{Rabie2023}
Rabie AH, Saleh AI, Mansour NA.
\newblock Red piranha optimization (RPO): a natural inspired meta-heuristic
  algorithm for solving complex optimization problems.
\newblock Journal of Ambient Intelligence and Humanized Computing. 2023;14(6);
  p. 7621--7648.

\bibitem{Givi2023}
Givi H, Dehghani M, Hub{\`a}lovsk{\`y} S.
\newblock Red {P}anda {O}ptimization {A}lgorithm: An {E}ffective
  {B}io-{I}nspired {M}etaheuristic {A}lgorithm for {S}olving {E}ngineering
  {O}ptimization {P}roblems.
\newblock IEEE Access. 2023;11; p. 57203--57227.

\bibitem{ravenroosting}
Brabazon A, Cui W, O'neill M.
\newblock The Raven Roosting Optimisation Algorithm.
\newblock Soft Computing. 2016;20(2); p. 525–545.

\bibitem{Ferahtia2023}
Ferahtia S, Houari A, Rezk H, Djerioui A, Machmoum M, Motahhir S, et~al.
\newblock Red-tailed hawk algorithm for numerical optimization and real-world
  problems.
\newblock Scientific Reports. 2023;13(1); p. 12950.

\bibitem{Abualigah2022}
Abualigah L, Elaziz MA, Sumari P, Geem ZW, Gandomi AH.
\newblock Reptile {S}earch {A}lgorithm {{(RSA)}}: {A} nature-inspired
  meta-heuristic optimizer.
\newblock Expert Systems with Applications. 2022;191; p. 116158.

\bibitem{Dhiman2021}
Dhiman G, Garg M, Nagar A, Kumar V, Dehghani M.
\newblock A novel algorithm for global optimization: rat swarm optimizer.
\newblock Journal of Ambient Intelligence and Humanized Computing. 2021;12; p.
  8457--8482.

\bibitem{ringedseal}
Saadi Y, Tri I, Yanto I, Herawan T, Balakrishnan V, Chiroma H, et~al.
\newblock Ringed Seal Search for Global Optimization via a Sensitive Search
  Model.
\newblock PLOS ONE. 2015;11(1); p. 1--31.

\bibitem{SA2}
Hersovici M, Jacovi M, Maarek YS, Pelleg D, Shtalhaim M, Ur S.
\newblock The shark-search algorithm. An application: tailored Web site
  mapping.
\newblock Computer Networks and ISDN Systems. 1998;30(1-7); p. 317--326.

\bibitem{Kusuma2024swarm}
Kusuma PD, Dinimaharawati A.
\newblock Swarm {B}ipolar {A}lgorithm: A {M}etaheuristic {B}ased on
  {P}olarization of {T}wo {E}qual {S}ize {S}ub {S}warms.
\newblock International Journal of Intelligent Engineering \& Systems.
  2024;17(2).

\bibitem{McCaffrey5211598}
{McCaffrey} JD.
\newblock Generation of pairwise test sets using a simulated bee colony
  algorithm.
\newblock In: 2009 IEEE International Conference on Information Reuse
  Integration; 2009. p. 115--119.

\bibitem{SBO}
Samareh~Moosavi SH, Khatibi~Bardsiri V.
\newblock Satin bowerbird optimizer: A new optimization algorithm to optimize
  ANFIS for software development effort estimation.
\newblock Engineering Applications of Artificial Intelligence. 2017;60; p.
  1--15.

\bibitem{SCA}
Mirjalili S.
\newblock SCA: A Sine Cosine Algorithm for solving optimization problems.
\newblock Knowledge-Based Systems. 2016;96; p. 120--133.

\bibitem{Seyyedabbasi2023}
Seyyedabbasi A, Kiani F.
\newblock Sand {C}at swarm optimization: A nature-inspired algorithm to solve
  global optimization problems.
\newblock Engineering with Computers. 2023;39; p. 2627--2651.

\bibitem{SDCS}
Rakhshani H, Rahati A.
\newblock Snap-drift cuckoo search: A novel cuckoo search optimization
  algorithm.
\newblock Applied Soft Computing. 2017;52; p. 771 -- 794.

\bibitem{SFLA}
Eusuff M, Lansey K, Pasha F.
\newblock Shuffled frog-leaping algorithm: a memetic metaheuristic for discrete
  optimization.
\newblock Engineering Optimization. 2006;38(2); p. 129--154.

\bibitem{dhiman2017spotted}
Dhiman G, Kumar V.
\newblock Spotted hyena optimizer: a novel bio-inspired based metaheuristic
  technique for engineering applications.
\newblock Advances in Engineering Software. 2017;114; p. 48--70.

\bibitem{SHO1}
Fausto F, Cuevas E, Valdivia A, González A.
\newblock A global optimization algorithm inspired in the behavior of selfish
  herds.
\newblock Biosystems. 2017;160; p. 39--55.

\bibitem{Zhao2023b}
Zhao S, Zhang T, Ma S, Wang M.
\newblock Sea-horse optimizer: a novel nature-inspired meta-heuristic for
  global optimization problems.
\newblock Applied Intelligence. 2023;53(10); p. 11833--11860.

\bibitem{SIP}
Su MC, Su SY, Zhao YX.
\newblock A swarm-inspired projection algorithm.
\newblock Pattern Recognition. 2009;42(11); p. 2764--2786.

\bibitem{MonismithJr.2008}
{Monismith} DR, {Mayfield} BE.
\newblock Slime Mold as a model for numerical optimization.
\newblock In: 2008 IEEE Swarm Intelligence Symposium; 2008. p. 1--8.

\bibitem{spermmotility}
Raouf O, M~Hezam I.
\newblock Sperm motility algorithm: a novel metaheuristic approach for global
  optimisation.
\newblock International Journal of Operational Research. 2017;28; p. 143.

\bibitem{SMO}
Chand~Bansal J, Sharma H, Singh~Jadon S, Clerc M.
\newblock Spider Monkey Optimization algorithm for numerical optimization.
\newblock Memetic Computation. 2014;6; p. 31--47.

\bibitem{Zamani2022}
Zamani H, Nadimi-Shahraki MH, Gandomi AH.
\newblock Starling murmuration optimizer: A novel bio-inspired algorithm for
  global and engineering optimization.
\newblock Computer Methods in Applied Mechanics and Engineering. 2022;392; p.
  114616.

\bibitem{SOA}
Dai C, Zhu Y, Chen W.
\newblock Seeker Optimization Algorithm.
\newblock In: Computational Intelligence and Security; 2007. p. 167--176.

\bibitem{SOA1}
Dhiman G, Kumar V.
\newblock Seagull optimization algorithm: Theory and its applications for
  large-scale industrial engineering problems.
\newblock Knowledge-Based Systems. 2019;165; p. 169--196.

\bibitem{SOA2}
Kaur A, Jain S, Goel S.
\newblock Sandpiper optimization algorithm: a novel approach for solving
  real-life engineering problems.
\newblock Applied Intelligence. 2020;50(2); p. 582--619.

\bibitem{SOA3}
Shadravan S, Naji HR, Bardsiri VK.
\newblock The Sailfish Optimizer: A novel nature-inspired metaheuristic
  algorithm for solving constrained engineering optimization problems.
\newblock Engineering Applications of Artificial Intelligence. 2019;80; p.
  20--34.

\bibitem{Dehghani2022serval}
Dehghani M, Trojovsk{\`y} P.
\newblock Serval {O}ptimization {A}lgorithm: {A} {N}ew {B}io-{I}nspired
  {A}pproach for {S}olving {O}ptimization {P}roblems.
\newblock Biomimetics. 2022;7(4); p. 204.

\bibitem{SOS}
Cheng MY, Prayogo D.
\newblock Symbiotic Organisms Search: A new metaheuristic optimization
  algorithm.
\newblock Computers and Structures. 2014;139; p. 98--112.

\bibitem{STOA}
Dhiman G, Kaur A.
\newblock STOA: A bio-inspired based optimization algorithm for industrial
  engineering problems.
\newblock Engineering Applications of Artificial Intelligence. 2019;82; p.
  148--174.

\bibitem{SSA}
Yu JJQ, Li VOK.
\newblock A social spider algorithm for global optimization.
\newblock Applied Soft Computing. 2015;30; p. 614--627.

\bibitem{JAIN2019148}
Jain M, Singh V, Rani A.
\newblock A novel nature-inspired algorithm for optimization: Squirrel search
  algorithm.
\newblock Swarm and Evolutionary Computation. 2019;44; p. 148--175.

\bibitem{MIRJALILI2017163}
Mirjalili S, Gandomi AH, Mirjalili SZ, Saremi S, Faris H, Mirjalili SM.
\newblock Salp Swarm Algorithm: A bio-inspired optimizer for engineering design
  problems.
\newblock Advances in Engineering Software. 2017;114; p. 163--191.

\bibitem{Jiankai2020}
Xue J, Shen B.
\newblock A novel swarm intelligence optimization approach: sparrow search
  algorithm.
\newblock Systems Science \& Control Engineering. 2020;8(1); p. 22--34.

\bibitem{Adhi2023}
R AL, S P.
\newblock Sling-shot spider optimization algorithm based packet length control
  in wireless sensor network and Internet of Things-based networks.
\newblock International Journal of Communication Systems. 2023;36(4); p. e5406.

\bibitem{SSO1}
Abedinia O, Amjady N, Ghasemi A.
\newblock A New Metaheuristic Algorithm Based on Shark Smell Optimization.
\newblock Complexity. 2016;21(5); p. 97--116.

\bibitem{SSO2}
Neshat M, Sepidnam G, Sargolzaei M.
\newblock Swallow swarm optimization algorithm: a new method to optimization.
\newblock Neural Computing and Applications. 2013;23(2); p. 429--454.

\bibitem{SSO}
Cuevas E, Cienfuegos M, Záldivar D, Pérez-Cisneros M.
\newblock A swarm optimization algorithm inspired in the behavior of the
  social-spider.
\newblock Expert Systems with Applications. 2013;40(16); p. 6374--6384.

\bibitem{spermfertilization}
Shehadeh H, Idris M, Ahmedy I, Ramli R, N~M N.
\newblock The Multi-Objective Optimization Algorithm Based on Sperm
  Fertilization Procedure (MOSFP) Method for Solving Wireless Sensor Networks
  Optimization Problems in Smart Grid Applications.
\newblock Energies. 2018;11; p. 97.

\bibitem{SSPCO}
{Omidvar} R, {Parvin} H, {Rad} F.
\newblock SSPCO Optimization Algorithm (See-See Partridge Chicks Optimization).
\newblock In: 2015 Fourteenth Mexican International Conference on Artificial
  Intelligence (MICAI); 2015. p. 101--106.

\bibitem{SSSE}
Haiyan Q, Xinling S.
\newblock A Surface-Simplex Swarm Evolution Algorithm.
\newblock Advances in Engineering Software. 2017;22; p. 38--50.

\bibitem{Trojovsky2023tiger}
Trojovsk{\`y} P, Dehghani M, Hanu{š} P.
\newblock Siberian Tiger Optimization: A New Bio-Inspired Metaheuristic
  Algorithm for Solving Engineering Optimization Problems.
\newblock IEEE Access. 2022;10; p. 132396--132431.

\bibitem{SWA}
Ebrahimi A, Khamehchi E.
\newblock Sperm whale algorithm: An effective metaheuristic algorithm for
  production optimization problems.
\newblock Journal of Natural Gas Science and Engineering. 2016;29; p. 211--222.

\bibitem{Abdel2023spider}
Abdel-Basset M, Mohamed R, Jameel M, Abouhawwash M.
\newblock Spider wasp optimizer: A novel meta-heuristic optimization algorithm.
\newblock Artificial Intelligence Review. 2023;56; p. 11675--–11738.

\bibitem{Zungeru20121901}
Zungeru AM, Ang LM, Seng KP.
\newblock Termite-hill: Performance optimized swarm intelligence based routing
  algorithm for wireless sensor networks.
\newblock Journal of Network and Computer Applications. 2012;35(6); p.
  1901--1917.

\bibitem{Majumder2023}
Majumder A.
\newblock Termite alate optimization algorithm: a swarm-based nature inspired
  algorithm for optimization problems.
\newblock Evolutionary Intelligence. 2023;16(3); p. 997--1017.

\bibitem{TCO}
{Hedayatzadeh} R, {Akhavan Salmassi} F, {Keshtgari} M, {Akbari} R, {Ziarati} K.
\newblock Termite colony optimization: A novel approach for optimizing
  continuous problems.
\newblock In: 2010 18th Iranian Conference on Electrical Engineering; 2010. p.
  553--558.

\bibitem{Dehghani2022}
Dehghani M, Hub{á}lovsk{ý} S, Trojovsk{\`y} P.
\newblock Tasmanian Devil Optimization: A New Bio-Inspired Optimization
  Algorithm for Solving Optimization Algorithm.
\newblock IEEE Access. 2022;10; p. 19599--19620.

\bibitem{Panteleev2022application}
Panteleev AV, Kolessa AA.
\newblock Application of the {T}omtit {F}lock {M}etaheuristic {O}ptimization
  {A}lgorithm to the {O}ptimal {D}iscrete {T}ime {D}eterministic {D}ynamical
  {C}ontrol {P}roblem.
\newblock Algorithms. 2022;15(9); p. 301.

\bibitem{TGSR}
Mozaffari A, Goudarzi AM, Fathi A.
\newblock Bio-inspired methods for fast and robust arrangement of
  thermoelectric modulus.
\newblock International Journal of Bio-Inspired Computation (IJBIC). 2013;5(1);
  p. 19--34.

\bibitem{Minh2023}
Minh HL, Sang-To T, Theraulaz G, {Abdel Wahab} M, Cuong-Le T.
\newblock Termite life cycle optimizer.
\newblock Expert Systems with Applications. 2023;213; p. 119211.

\bibitem{Sahu2023}
Sahu VSDM, Samal P, Panigrahi CK.
\newblock Tyrannosaurus optimization algorithm: A new nature-inspired
  meta-heuristic algorithm for solving optimal control problems.
\newblock e-Prime - Advances in Electrical Engineering, Electronics and Energy.
  2023;5; p. 100243.

\bibitem{TSA1}
Kaur S, Awasthi LK, Sangal AL, Dhiman G.
\newblock Tunicate Swarm Algorithm: A new bio-inspired based metaheuristic
  paradigm for global optimization.
\newblock Engineering Applications of Artificial Intelligence. 2020;90; p.
  103541.

\bibitem{Layeb2022tangent}
Layeb A.
\newblock Tangent search algorithm for solving optimization problems.
\newblock Neural Computing and Applications. 2022;34(11); p. 8853--8884.

\bibitem{VAA}
Yang X, Lees JM, Morley CT.
\newblock Application of Virtual Ant Algorithms in the Optimization of {CFRP}
  Shear Strengthened Precracked Structures.
\newblock In: Computational Science - {ICCS} 2006, 6th International
  Conference, Proceedings, Part {I}; 2006. p. 834--837.

\bibitem{VBA}
Yang XS.
\newblock Engineering Optimizations via Nature-Inspired Virtual Bee Algorithms.
\newblock In: Artificial Intelligence and Knowledge Engineering Applications: A
  Bioinspired Approach; 2005. p. 317--323.

\bibitem{VCS}
Li MD, Zhao H, Weng XW, Han T.
\newblock A novel nature-inspired algorithm for optimization: Virus colony
  search.
\newblock Advances in Engineering Software. 2016;92; p. 65--88.

\bibitem{juarez2009virus}
Juarez JRC, Wang HJ, Lai YC, Liang YC.
\newblock Virus Optimization Algorithm (VOA): A novel metaheuristic for solving
  continuous optimization problems.
\newblock In: Proceedings of the 2009 Asia Pacific Industrial Engineering and
  Management Systems Conference (APIEMS 2009); 2009. p. 2166--2174.

\bibitem{VSO}
Cortés P, García JM, Muñuzuri J, Onieva L.
\newblock Viral systems: A new bio-inspired optimisation approach.
\newblock Computers and Operations Research. 2008;35; p. 2840--2860.

\bibitem{WCA}
Theraulaz G, Goss S, Gervet J, Deneubourg JL.
\newblock Task differentiation in Polistes wasp colonies: a model for
  self-organizing groups of robots.
\newblock In: Proceedings of the First International Conference on Simulation
  of Adaptive Behavior: From Animals to Animates; 1991. p. 346--355.

\bibitem{WCA1}
Liu CY, Yan XH, Wu H.
\newblock The Wolf Colony Algorithm and Its Application.
\newblock Chinese Journal of Electronics. 2011;20; p. 212--216.

\bibitem{WO}
Arnaout JP.
\newblock Worm Optimization: A novel optimization algorithm inspired by C.
  Elegans.
\newblock In: Proceedings of the 2014 International Conference on Industrial
  Engineering and Operations Management; 2014. p. 2499--2505.

\bibitem{WOA}
Mirjalili S, Lewis A.
\newblock The Whale Optimization Algorithm.
\newblock Advances in Engineering Software. 2016;95; p. 51--67.

\bibitem{Trojovsky2023}
Trojovsk{\`y} P, Dehghani M.
\newblock A new bio-inspired metaheuristic algorithm for solving optimization
  problems based on walruses behavior.
\newblock Scientific Reports. 2023;13; p. 8775.

\bibitem{Yang:2007:AMH:1337686.1337957}
Yang C, Tu X, Chen J.
\newblock Algorithm of Marriage in Honey Bees Optimization Based on the Wolf
  Pack Search.
\newblock In: Proceedings of the The 2007 International Conference on
  Intelligent Pervasive Computing; 2007. p. 462--467.

\bibitem{WSA1}
Ting TO, Man KL, Guan SU, Nayel M, Wan K.
\newblock Weightless Swarm Algorithm (WSA) for Dynamic Optimization Problems.
\newblock In: Network and Parallel Computing, IFIP International Conference on
  Network and Parallel Computing; 2012. p. 508--515.

\bibitem{WSA}
Tang R, Fong S, Yang XS, Deb S.
\newblock Wolf search algorithm with ephemeral memory.
\newblock In: Seventh International Conference on Digital Information
  Management (ICDIM 2012); 2012. p. 165--172.

\bibitem{WSO}
Pinto P, Runkler TA, Sousa JM.
\newblock Wasp swarm optimization of logistic systems.
\newblock In: Adaptive and Natural Computing Algorithms; 2005. p. 264--267.

\bibitem{Braik2022}
Braik M, Hammouri A, Atwan J, Al-Betar MA, Awadallah MA.
\newblock White Shark Optimizer: A novel bio-inspired meta-heuristic algorithm
  for global optimization problems.
\newblock Knowledge-Based Systems. 2022;243; p. 108457.

\bibitem{YSGA}
Zald{\'\i}var D, Morales B, Rodr{\'\i}guez A, Valdivia-G A, Cuevas E,
  P{\'e}rez-Cisneros M.
\newblock A novel bio-inspired optimization model based on Yellow Saddle
  Goatfish behavior.
\newblock Biosystems. 2018;174; p. 1--21.

\bibitem{Trojovska2022}
Trojovská E, Dehghani M, Trojovsk{\`y} P.
\newblock Zebra Optimization Algorithm: A New Bio-Inspired Optimization
  Algorithm for Solving Optimization Algorithm.
\newblock IEEE Access. 2022;10; p. 49445--49473.

\bibitem{ZSO}
Nguyen HT, Bhanu B.
\newblock Zombie Survival Optimization: A Swarm Intelligence Algorithm Inspired
  By Zombie Foraging.
\newblock In: 21st International Conference on Pattern Recognition (ICPR 2012);
  2012. p. 987--990.

\bibitem{Yang2018}
Yang XS, Deb S, Mishra SK.
\newblock Multi-species Cuckoo Search Algorithm for Global Optimization.
\newblock Cognitive Computation. 2018;10(6); p. 1085--1095.

\bibitem{GSA}
Rashedi E, Nezamabadi-Pour H, Saryazdi S.
\newblock GSA: A Gravitational Search Algorithm.
\newblock Information Sciences. 2009;179(13); p. 2232--2248.

\bibitem{BH}
Hatamlou A.
\newblock Black hole: A new heuristic optimization approach for data
  clustering.
\newblock Information Sciences. 2013;222; p. 175--184.

\bibitem{GBSA}
Shah-Hosseini H.
\newblock Principal components analysis by the galaxy-based search algorithm: a
  novel metaheuristic for continuous optimisation.
\newblock International Journal of Computational Science and Engineering.
  2011;6(1-2); p. 132--140.

\bibitem{HS}
Lee KS, Geem ZW.
\newblock A new meta-heuristic algorithm for continuous engineering
  optimization: harmony search theory and practice.
\newblock Computer Methods in Applied Mechanics and Engineering. 2005;194; p.
  3902--3933.

\bibitem{ANITA201993}
Yadav A, Yadav A.
\newblock AEFA: Artificial electric field algorithm for global optimization.
\newblock Swarm and Evolutionary Computation. 2019;48; p. 93--108.

\bibitem{hashim2021archimedes}
Hashim FA, Hussain K, Houssein EH, Mabrouk MS, Al-Atabany W.
\newblock Archimedes optimization algorithm: a new metaheuristic algorithm for
  solving optimization problems.
\newblock Applied Intelligence. 2021;51; p. 1531--1551.

\bibitem{Xie20091321}
Xie L, Zeng J, Cui Z.
\newblock General framework of Artificial Physics Optimization Algorithm.
\newblock In: 2009 World Congress on Nature Biologically Inspired Computing
  (NaBIC); 2009. p. 1321--1326.

\bibitem{ASO1}
Zhao W, Wang L, Zhang Z.
\newblock Atom search optimization and its application to solve a hydrogeologic
  parameter estimation problem.
\newblock Knowledge-Based Systems. 2019;163; p. 283--304.

\bibitem{BBBC}
Erol OK, Eksin I.
\newblock A new optimization method: Big Bang–Big Crunch.
\newblock Advances in Engineering Software. 2006;37(2); p. 106--111.

\bibitem{CBO}
Kaveh A, Mahdavi VR.
\newblock Colliding bodies optimization: A novel meta-heuristic method.
\newblock Computers and Structures. 2014;139; p. 18--27.

\bibitem{CEO}
Feng X, Ma M, Yu H.
\newblock Crystal Energy Optimization Algorithm.
\newblock Computational Intelligence. 2016;32(2); p. 284--322.

\bibitem{CFO}
Formato RA.
\newblock Central Force Optimization: A New Nature Inspired Computational
  Framework for Multidimensional Search and Optimization.
\newblock In: Nature Inspired Cooperative Strategies for Optimization (NICSO
  2007). Springer Berlin Heidelberg; 2008. p. 221--238.

\bibitem{CSS}
Kaveh A, Talatahari S.
\newblock A novel heuristic optimization method: charged system search.
\newblock Acta Mechanica. 2010;213(3-4); p. 267--289.

\bibitem{EFO}
Abedinpourshotorban H, Shamsuddin SM, Beheshti Z, Jawawi DNA.
\newblock Electromagnetic field optimization: A physics-inspired metaheuristic
  optimization algorithm.
\newblock Swarm and Evolutionary Computation. 2016;26; p. 8--22.

\bibitem{EMO}
Ilker~Birbil S, Fang SC.
\newblock An Electromagnetism-like Mechanism for Global Optimization.
\newblock Journal of Global Optimization. 2003;25(3); p. 263--282.

\bibitem{EOA1}
Khalafallah A, Abdel-Raheem M.
\newblock Electimize: new evolutionary algorithm for optimization with
  application in construction engineering.
\newblock Journal of Computing in Civil Engineering. 2011;25(3); p. 192--201.

\bibitem{ERSA}
Rahmanzadeh S, Pishvaee MS.
\newblock Electron radar search algorithm: a novel developed meta-heuristic
  algorithm.
\newblock Soft Computing. 2020;24(11); p. 8443--8465.

\bibitem{Kundu19991149}
Kundu S.
\newblock Gravitational clustering: A new approach based on the spatial
  distribution of the points.
\newblock Pattern Recognition. 1999;32(7); p. 1149--1160.

\bibitem{Barzegar2009}
{Barzegar} B, {Rahmani} AM, {Far} KZ.
\newblock Gravitational emulation local search algorithm for advanced
  reservation and scheduling in grid systems.
\newblock In: 2009 First Asian Himalayas International Conference on Internet;
  2009. p. 1--5.

\bibitem{Zheng2010}
Zheng M, Liu Gx, Zhou Cg, Liang Yc, Wang Y.
\newblock Gravitation field algorithm and its application in gene cluster.
\newblock Algorithms for Molecular Biology. 2010;5(32); p. 1--11.

\bibitem{Ghasemi2023}
Ghasemi M, Zare M, Zahedi A, Akbari MA, Mirjalili S, Abualigah L.
\newblock Geyser inspired algorithm: A new geological-inspired meta-heuristic
  for real-parameter and constrained engineering optimization.
\newblock Journal of Bionic Engineering. 2023;p. 1--35.

\bibitem{Flores2011226}
Flores JJ, L{\'o}pez R, Barrera J.
\newblock Gravitational Interactions Optimization.
\newblock In: Learning and Intelligent Optimization; 2011. p. 226--237.

\bibitem{GRSA}
Beiranvand H, Rokrok E.
\newblock General Relativity Search Algorithm: A Global Optimization Approach.
\newblock International Journal of Computational Intelligence and Applications.
  2015;14(3); p. 1--29.

\bibitem{GSO2}
Muthiah-Nakarajan V, Noel MM.
\newblock Galactic Swarm Optimization: A new global optimization metaheuristic
  inspired by galactic motion.
\newblock Applied Soft Computing. 2016;38; p. 771--787.

\bibitem{wedyan2017hydrological}
Wedyan A, Whalley J, Narayanan A.
\newblock Hydrological Cycle Algorithm for Continuous Optimization Problems.
\newblock Journal of Optimization. 2017;2017; p. 1--25.

\bibitem{cui2010lambda}
Cui Y, Guo R, Guo D.
\newblock Lambda algorithm.
\newblock Journal of Uncertain Systems. 2010;4(1); p. 22--33.

\bibitem{Zarand2002150201}
Zar\'and G, P\'azm\'andi F, P\'al KF, Zim\'anyi GT.
\newblock Using Hysteresis for Optimization.
\newblock Physical Review Letters. 2002;89(15); p. 150201.

\bibitem{HO}
Rbouh I, El~Imrani AA.
\newblock Hurricane-based Optimization Algorithm.
\newblock AASRI Procedia. 2014;6; p. 26--33.

\bibitem{intelligentGSA}
Askari H, Zahiri SH.
\newblock Intelligent Gravitational Search Algorithm for optimum design of
  fuzzy classifier.
\newblock In: 2012 2nd International eConference on Computer and Knowledge
  Engineering, ICCKE 2012; 2012. p. 98--104.

\bibitem{IWD}
Shah-Hosseini H.
\newblock The intelligent water drops algorithm: a nature-inspired swarm-based
  optimization algorithm.
\newblock International Journal of Bio-inspired computation. 2009;1(1); p.
  71--79.

\bibitem{Shen2010154}
{Jihong Shen}, {Jialian Li}.
\newblock The principle analysis of Light Ray Optimization Algorithm.
\newblock In: 2010 Second International Conference on Computational
  Intelligence and Natural Computing. vol.~2; 2010. p. 154--157.

\bibitem{LSA}
Shareef H, Ibrahim AA, Mutlag AH.
\newblock Lightning search algorithm.
\newblock Applied Soft Computing. 2015;36; p. 315--333.

\bibitem{Tayarani20082659}
{Tayarani-N} MH, {Akbarzadeh-T} MR.
\newblock Magnetic Optimization Algorithms a new synthesis.
\newblock In: 2008 IEEE Congress on Evolutionary Computation (IEEE World
  Congress on Computational Intelligence); 2008. p. 2659--2664.

\bibitem{Mora-Gutierrez2014301}
Mora-Gutiérrez RA, Ramírez-Rodríguez J, Rincón-García EA.
\newblock An optimization algorithm inspired by musical composition.
\newblock Artificial Intelligence Review. 2014;41(3); p. 301--315.

\bibitem{Ashrafi2011109}
{Ashrafi} SM, {Dariane} AB.
\newblock A novel and effective algorithm for numerical optimization: Melody
  Search (MS).
\newblock In: 2011 11th International Conference on Hybrid Intelligent Systems
  (HIS); 2011. p. 109--114.

\bibitem{MVO}
Mirjalili S, Mirjalili SM, Hatamlou A.
\newblock Multi-Verse Optimizer: a nature-inspired algorithm for global
  optimization.
\newblock Neural Computing and Applications. 2016;27(2); p. 495--513.

\bibitem{Sowmya2024}
Sowmya R, Premkumar M, Jangir P.
\newblock Newton-{R}aphson-based optimizer: A new population-based
  metaheuristic algorithm for continuous optimization problems.
\newblock Engineering Applications of Artificial Intelligence. 2024;128; p.
  107532.

\bibitem{OIO}
Kashan AH.
\newblock A new metaheuristic for optimization: Optics inspired optimization
  (OIO).
\newblock Computers and Operations Research. 2015;55; p. 99--125.

\bibitem{Sacco2007}
Sacco WF, Filho HA, De~Oliveira CRE.
\newblock A populational particle collision algorithm applied to a nuclear
  reactor core design optimization.
\newblock In: Joint International Topical Meeting on Mathematics and
  Computations and Supercomputing in Nuclear Applications, 2007; 2007. p.
  1--10.

\bibitem{Taillard2002613}
Taillard {\'E}D, Voss S.
\newblock Popmusic --- Partial Optimization Metaheuristic under Special
  Intensification Conditions.
\newblock In: Essays and Surveys in Metaheuristics. Springer US; 2002. p.
  613--629.

\bibitem{saire2015approach}
Saire JEC, T{\'u}pac VYJ.
\newblock An approach to real-coded quantum inspired evolutionary algorithm
  using particles filter.
\newblock In: 2015 Latin America Congress on Computational Intelligence
  (LA-CCI); 2015. p. 1--6.

\bibitem{AGHAYKABOLI201731}
Kaboli SHA, Selvaraj J, Rahim NA.
\newblock Rain-fall optimization algorithm: A population based algorithm for
  solving constrained optimization problems.
\newblock Journal of Computational Science. 2017;19; p. 31--42.

\bibitem{biyantowater}
Biyanto TR, {Matradji}, Febrianto HY, Afdanny N, Rahman AH, Gunawan KS.
\newblock Rain Water Algorithm: {{Newton}}'s Law of Rain Water Movements during
  Free Fall and Uniformly Accelerated Motion Utilization.
\newblock AIP Conference Proceedings. 2019;2088(1); p. 020053.

\bibitem{RFD}
Rabanal P, Rodr{\'i}guez I, Rubio F.
\newblock Using River Formation Dynamics to Design Heuristic Algorithms.
\newblock In: Unconventional Computation; 2007. p. 163--177.

\bibitem{RMO}
Rahmani R, Yusof R.
\newblock A new simple, fast and efficient algorithm for global optimization
  over continuous search-space problems: Radial Movement Optimization.
\newblock Applied Mathematics and Computation. 2014;248; p. 287--300.

\bibitem{RO}
Kaveh A, Khayatazad M.
\newblock A new meta-heuristic method: Ray Optimization.
\newblock Computers and Structures. 2012;112--113; p. 283--294.

\bibitem{Deng2023}
Deng L, Liu S.
\newblock Snow ablation optimizer: A novel metaheuristic technique for
  numerical optimization and engineering design.
\newblock Expert Systems with Applications. 2023;225; p. 120069.

\bibitem{Hsiao20052323}
Hsiao YT, Chuang CL, Jiang JA, Chien CC.
\newblock A novel optimization algorithm: space gravitational optimization.
\newblock In: 2005 IEEE International Conference on Systems, Man and
  Cybernetics. vol.~3; 2005. p. 2323--2328.

\bibitem{tzanetos2017new}
Tzanetos A, Dounias G.
\newblock A new metaheuristic method for optimization: sonar inspired
  optimization.
\newblock In: International Conference on Engineering Applications of Neural
  Networks; 2017. p. 417--428.

\bibitem{SMS}
Cuevas E, Echavarría A, Ramírez-Ortegón MA.
\newblock An optimization algorithm inspired by the States of Matter that
  improves the balance between exploration and exploitation.
\newblock Applied Intelligence. 2014;40; p. 256--272.

\bibitem{SO}
Tamura K, Yasuda K.
\newblock Primary Study of Spiral Dynamics Inspired Optimization.
\newblock IEEJ Transactions On Electrical And Electronic Engineering. 2011;6;
  p. 98--100.

\bibitem{jin2010nature}
Jin GG, Tran TD.
\newblock A nature-inspired evolutionary algorithm based on spiral movements.
\newblock In: Proceedings of SICE Annual Conference 2010; 2010. p. 1643--1647.

\bibitem{SPP}
Vicsek T, Czir\'ok A, Ben-Jacob E, Cohen I, Shochet O.
\newblock Novel Type of Phase Transition in a System of Self-Driven Particles.
\newblock Physical Review Letters. 1995;75(6); p. 1226--1229.

\bibitem{Zitouni2021}
Zitouni F, Harous S, Maamri R.
\newblock The {S}olar {S}ystem {A}lgorithm: A {N}ovel {M}etaheuristic {M}ethod
  for {G}lobal {O}ptimization.
\newblock IEEE Access. 2021;9; p. 4542--4565.

\bibitem{TFWO}
Ghasemi M, Davoudkhani IF, Akbari E, Rahimnejad A, Ghavidel S, Li L.
\newblock A novel and effective optimization algorithm for global optimization
  and its engineering applications: Turbulent Flow of Water-based Optimization
  (TFWO).
\newblock Engineering Applications of Artificial Intelligence. 2020;92; p.
  103666.

\bibitem{VPO}
Kaveh A, Ilchi~Ghazaan M.
\newblock Vibrating particles system algorithm for truss optimization with
  multiple natural frequency constraints.
\newblock Acta Mechanica. 2017;228; p. 307--322.

\bibitem{VS}
Dogan B, Ölmez T.
\newblock A new metaheuristic for numerical function optimization: Vortex
  Search algorithm.
\newblock Information Sciences. 2015;293; p. 125--145.

\bibitem{WCA2}
Eskandar H, Sadollah A, Bahreininejad A, Hamdi M.
\newblock Water cycle algorithm – A novel metaheuristic optimization method
  for solving constrained engineering optimization problems.
\newblock Computers and Structures. 2012;110--111; p. 151--166.

\bibitem{WEO}
Kaveh A, Bakhshpoori T.
\newblock Water Evaporation Optimization: A novel physically inspired
  optimization algorithm.
\newblock Computers and Structures. 2016;167; p. 69--85.

\bibitem{Yang2007475}
Yang FC, Wang YP.
\newblock Water flow-like algorithm for object grouping problems.
\newblock Journal of the Chinese Institute of Industrial Engineers. 2007;24(6);
  p. 475--488.

\bibitem{Basu20071825}
Basu S, Chaudhuri C, Kundu M, Nasipuri M, Basu DK.
\newblock Text line extraction from multi-skewed handwritten documents.
\newblock Pattern Recognition. 2007;40(6); p. 1825--1839.

\bibitem{WFO}
Tran TH, Ng KM.
\newblock A water-flow algorithm for flexible flow shop scheduling with
  intermediate buffers.
\newblock Journal of Scheduling. 2011;14(5); p. 483--500.

\bibitem{WWA}
Zheng YJ.
\newblock Water wave optimization: A new nature-inspired metaheuristic.
\newblock Computers and Operations Research. 2015;55; p. 1--11.

\bibitem{Irizarry20055663}
Irizarry R.
\newblock A generalized framework for solving dynamic optimization problems
  using the artificial chemical process paradigm: Applications to particulate
  processes and discrete dynamic systems.
\newblock Chemical Engineering Science. 2005;60(21); p. 5663--5681.

\bibitem{ACROA}
Alatas B.
\newblock ACROA: Artificial Chemical Reaction Optimization Algorithm for global
  optimization.
\newblock Expert Systems with Applications. 2011;38; p. 13170--13180.

\bibitem{Melin20133185}
Melin P, Astudillo L, Castillo O, Valdez F, Garcia M.
\newblock Optimal design of type-2 and type-1 fuzzy tracking controllers for
  autonomous mobile robots under perturbed torques using a new chemical
  optimization paradigm.
\newblock Expert Systems with Applications. 2013;40(8); p. 3185--3195.

\bibitem{CRO1}
{Lam} AYS, {Li} VOK.
\newblock Chemical-Reaction-Inspired Metaheuristic for Optimization.
\newblock IEEE Transactions on Evolutionary Computation. 2010;14(3); p.
  381--399.

\bibitem{GBMO}
Abdechiri M, Meybodi MR, Bahrami H.
\newblock Gases Brownian Motion Optimization: an Algorithm for Optimization
  (GBMO).
\newblock Applied Soft Computing. 2013;13; p. 2932--2946.

\bibitem{Patel2015HeatTS}
Patel VK, Savsani VJ.
\newblock Heat transfer search (HTS): a novel optimization algorithm.
\newblock Information Science. 2015;324; p. 217--246.

\bibitem{IMO}
Javidy B, Hatamlou A, Mirjalili S.
\newblock Ions motion algorithm for solving optimization problems.
\newblock Applied Soft Computing. 2015;32; p. 72--79.

\bibitem{Chuang20073157}
Chuang CL, Jiang JA.
\newblock Integrated radiation optimization: inspired by the gravitational
  radiation in the curvature of space-time.
\newblock In: 2007 IEEE Congress on Evolutionary Computation; 2007. p.
  3157--3164.

\bibitem{KGMO}
Moein S, Logeswaran R.
\newblock KGMO: A swarm optimization algorithm based on the kinetic energy of
  gas molecules.
\newblock Information Sciences. 2014;275; p. 127--144.

\bibitem{Murase2000115}
Murase H.
\newblock Finite element inverse analysis using a photosynthetic algorithm.
\newblock Computers and Electronics in Agriculture. 2000;29(1-2); p. 115--123.

\bibitem{SFO}
Subashini P, Dhivyaprabha TT, Krishnaveni M.
\newblock Synergistic Fibroblast Optimization.
\newblock In: Artificial Intelligence and Evolutionary Computations in
  Engineering Systems; 2017. p. 285--294.

\bibitem{TEO}
Kaveh A, Dadras A.
\newblock A novel meta-heuristic optimization algorithm: Thermal exchange
  optimization.
\newblock Advances in Engineering Software. 2017;110; p. 69--84.

\bibitem{IA}
Huan TT, Kulkarni AJ, Kanesan J, Huang CJ, Abraham A.
\newblock Ideology algorithm: a socio-inspired optimization methodology.
\newblock Neural Computing and Applications. 2017;28(1); p. 845--876.

\bibitem{ICA}
{Atashpaz-Gargari} E, {Lucas} C.
\newblock Imperialist competitive algorithm: An algorithm for optimization
  inspired by imperialistic competition.
\newblock In: 2007 IEEE Congress on Evolutionary Computation; 2007. p.
  4661--4667.

\bibitem{SLC}
Moosavian N, Roodsari BK.
\newblock Soccer league competition algorithm: A novel meta-heuristic algorithm
  for optimal design of water distribution networks.
\newblock Swarm and Evolutionary Computation. 2014;17; p. 14--24.

\bibitem{BSO}
Shi Y.
\newblock Brain Storm Optimization Algorithm.
\newblock In: Advances in Swarm Intelligence; 2011. p. 303--309.

\bibitem{GBSO}
El-Abd M.
\newblock Global-best brain storm optimization algorithm.
\newblock Swarm and Evolutionary Computation. 2017;37; p. 27 -- 44.

\bibitem{AISA}
Bogar E, Beyhan S.
\newblock Adolescent Identity Search Algorithm (AISA): A novel metaheuristic
  approach for solving optimization problems.
\newblock Applied Soft Computing. 2020;95; p. 106503.

\bibitem{ASOBueno}
{Ahmadi-Javid} A.
\newblock Anarchic Society Optimization: A human-inspired method.
\newblock In: 2011 IEEE Congress of Evolutionary Computation (CEC); 2011. p.
  2586--2592.

\bibitem{Yuan2022}
Yuan Y, Ren J, Wang S, Wang Z, Mu X, Zhao W.
\newblock Alpine skiing optimization: A new bio-inspired optimization
  algorithm.
\newblock Advances in Engineering Software. 2022;170; p. 103158.

\bibitem{Bodaghi2019}
Bodaghi M, Samieefar K.
\newblock Meta-heuristic bus transportation algorithm.
\newblock Iran Journal of Computer Science. 2019;2; p. 23--32.

\bibitem{CDOA}
Zhang Q, Wang R, Yang J, Ding K, Li Y, Hu J.
\newblock Collective decision optimization algorithm: A new heuristic
  optimization method.
\newblock Neurocomputing. 2017;221; p. 123--137.

\bibitem{COA1}
Li M, Zhao H, Weng X, Han T.
\newblock Cognitive behavior optimization algorithm for solving optimization
  problems.
\newblock Applied Soft Computing. 2016;39; p. 199--222.

\bibitem{COOA}
Sharafi Y, Khanesar MA, Teshnehlab M.
\newblock COOA: Competitive optimization algorithm.
\newblock Swarm and Evolutionary Computation. 2016;30; p. 39--63.

\bibitem{scientistopt}
Milani A, Santucci V.
\newblock Community of scientist optimization: An autonomy oriented approach to
  distributed optimization.
\newblock AI Communications. 2012;25; p. 157--172.

\bibitem{Jin19991672}
{Xidong Jin}, {Reynolds} RG.
\newblock Using knowledge-based evolutionary computation to solve nonlinear
  constraint optimization problems: a cultural algorithm approach.
\newblock In: Proceedings of the 1999 Congress on Evolutionary
  Computation-CEC99. vol.~3; 1999. p. 1672--1678.

\bibitem{biyanto2016duelist}
Biyanto TR, Fibrianto HY, Nugroho G, Hatta AM, Listijorini E, Budiati T, et~al.
\newblock Duelist algorithm: an algorithm inspired by how duelist improve their
  capabilities in a duel.
\newblock In: International Conference on Swarm Intelligence; 2016. p. 39--47.

\bibitem{EAemami}
Emami H, Derakhshan F.
\newblock Election algorithm: A new socio-politically inspired strategy.
\newblock AI Communications. 2015;28(3); p. 591--603.

\bibitem{Fadakar20166}
{Fadakar} E, {Ebrahimi} M.
\newblock A new metaheuristic football game inspired algorithm.
\newblock In: 2016 1st Conference on Swarm Intelligence and Evolutionary
  Computation (CSIEC); 2016. p. 6--11.

\bibitem{FIFAAO}
Razmjooy N, Khalilpour M, Ramezan M.
\newblock A New Meta-Heuristic Optimization Algorithm Inspired by FIFA World
  Cup Competitions: Theory and Its Application in PID Designing for AVR System.
\newblock Journal of Control, Automation and Electrical Systems. 2016;27(4); p.
  419--440.

\bibitem{Osaba2014}
Osaba E, Diaz F, Onieva E.
\newblock Golden ball: a novel meta-heuristic to solve combinatorial
  optimization problems based on soccer concepts.
\newblock Applied Intelligence. 2014;41(1); p. 145--166.

\bibitem{GCO}
Eita MA, Fahmy MM.
\newblock Group counseling optimization.
\newblock Applied Soft Computing. 2010;22; p. 585--604.

\bibitem{Daskin2011761}
Daskin A, Kais S.
\newblock Group leaders optimization algorithm.
\newblock Molecular Physics. 2011;109(5); p. 761--772.

\bibitem{GPO}
{Lenord Melvix} JSM.
\newblock Greedy Politics Optimization: Metaheuristic inspired by political
  strategies adopted during state assembly elections.
\newblock In: 2014 IEEE International Advance Computing Conference (IACC);
  2014. p. 1157--1162.

\bibitem{Kapoor2023}
Kapoor M, Pathak BK, Kumar R.
\newblock A nature-inspired meta-heuristic knowledge-based algorithm for
  solving multiobjective optimization problems.
\newblock Journal of Engineering Mathematics. 2023;143; p. 5.

\bibitem{GTOA}
Zhang Y, Jin Z.
\newblock Group Teaching Optimization Algorithm: A Novel Metaheuristic Method
  for Solving Global Optimization Problems.
\newblock Expert Systems with Applications. 2020;148; p. 113246.

\bibitem{HEM}
Montiel O, Castillo O, Melin P, Rodríguez~Díaz A, Sepúlveda R.
\newblock Human evolutionary model: A new approach to optimization.
\newblock Information Sciences. 2007;177(10); p. 2075--2098.

\bibitem{Thammano20101628}
Thammano A, Moolwong J.
\newblock A new computational intelligence technique based on human group
  formation.
\newblock Expert Systems with Applications. 2010;37(2); p. 1628--1634.

\bibitem{HIA}
{Zhang} LM, {Dahlmann} C, {Zhang} Y.
\newblock Human-Inspired Algorithms for continuous function optimization.
\newblock In: 2009 IEEE International Conference on Intelligent Computing and
  Intelligent Systems; 2009. p. 318--321.

\bibitem{HUA}
Ghasemian H, Ghasemian F, Vahdat-Nejad H.
\newblock Human urbanization algorithm: A novel metaheuristic approach.
\newblock Mathematics and Computers in Simulation. 2020;178; p. 1--15.

\bibitem{KKOA}
Srivastava A, Das DK.
\newblock A new Kho-Kho optimization Algorithm: An application to solve
  combined emission economic dispatch and combined heat and power economic
  dispatch problem.
\newblock Engineering Applications of Artificial Intelligence. 2020;94; p.
  103763.

\bibitem{LCA}
Kashan AH.
\newblock League Championship Algorithm (LCA): An algorithm for global
  optimization inspired by sport championships.
\newblock Applied Soft Computing. 2014;16; p. 171--200.

\bibitem{LCBO}
Khatri A, Gaba A, Rana K, Kumar V.
\newblock A novel life choice-based optimizer.
\newblock Soft Computing. 2020;24(12); p. 9121--9141.

\bibitem{Gonzalez-Fernandez2015776}
{Gonzalez-Fernandez} Y, {Chen} S.
\newblock Leaders and followers - A new metaheuristic to avoid the bias of
  accumulated information.
\newblock In: 2015 IEEE Congress on Evolutionary Computation (CEC); 2015. p.
  776--783.

\bibitem{OBA}
Hu TC, Kahng AB, Tsao CWA.
\newblock Old Bachelor Acceptance: A New Class of Non-Monotone Threshold
  Accepting Methods.
\newblock ORSA Journal on Computing. 1995;7(4); p. 417--425.

\bibitem{Zhang20082856}
Zhang X, Chen W, Dai C.
\newblock Application of oriented search algorithm in reactive power
  optimization of power system.
\newblock In: 2008 Third International Conference on Electric Utility
  Deregulation and Restructuring and Power Technologies; 2008. p. 2856--2861.

\bibitem{Askari2020}
Askari Q, Younas I, Saeed M.
\newblock Political {O}ptimizer: A novel socio-inspired meta-heuristic for
  global optimization.
\newblock Knowledge-Based Systems. 2020;195; p. 105709.

\bibitem{POA}
Borji A, Hamide M.
\newblock A new approach to global optimization motivated by parliamentary
  political competitions.
\newblock International Journal of Innovative Computing, Information and
  Control. 2009;5; p. 1643--1653.

\bibitem{PRO}
{Samareh Moosavi} SH, Bardsiri VK.
\newblock Poor and rich optimization algorithm: A new human-based and multi
  populations algorithm.
\newblock Engineering Applications of Artificial Intelligence. 2019;86; p.
  165--181.

\bibitem{ZHANG2018464}
Zhang J, Xiao M, Gao L, Pan Q.
\newblock Queuing search algorithm: A novel metaheuristic algorithm for solving
  engineering optimization problems.
\newblock Applied Mathematical Modelling. 2018;63; p. 464--490.

\bibitem{SAR}
Shabani A, Asgarian B, Gharebaghi SA, Salido MA, Giret A.
\newblock A new optimization algorithm based on search and rescue operations.
\newblock Mathematical Problems in Engineering. 2019;2019; p. 1--24.

\bibitem{SBO1}
Ray T, Liew KM.
\newblock Society and Civilization: An Optimization Algorithm Based on the
  Simulation of Social Behavior.
\newblock IEEE Transactions On Evolutionary Computation. 2003;7(4); p.
  386--396.

\bibitem{SCOXie}
Xie XF, Zhang WJ, Yang ZL.
\newblock Social cognitive optimization for nonlinear programming problems.
\newblock In: Proceedings. International Conference on Machine Learning and
  Cybernetics. vol.~2; 2002. p. 779---783.

\bibitem{Wei201011}
{Wei} Z, {Cui} Z, {Zeng} J.
\newblock Social Cognitive Optimization Algorithm with Reactive Power
  Optimization of Power System.
\newblock In: 2010 International Conference on Computational Aspects of Social
  Networks; 2010. p. 11--14.

\bibitem{SEA}
Xu Y, Cui Z, Zeng J.
\newblock Social Emotional Optimization Algorithm for Nonlinear Constrained
  Optimization Problems.
\newblock In: Swarm, Evolutionary, and Memetic Computing; 2010. p. 583--590.

\bibitem{Emami2022stock}
Emami H.
\newblock Stock exchange trading optimization algorithm: a human-inspired
  method for global optimization.
\newblock The Journal of Supercomputing. 2022;78(2); p. 2125--2174.

\bibitem{Weibo2008583}
Weibo W, Quanyuan F, Yongkang Z.
\newblock A novel particle swarm optimization algorithm with stochastic
  focusing search for real-parameter optimization.
\newblock In: 2008 11th IEEE Singapore International Conference on
  Communication Systems; 2008. p. 583--587.

\bibitem{SGO}
Dwi~Purnomo H.
\newblock Soccer Game Optimization: Fundamental Concept.
\newblock Jurnal Sistem Komputer. 2012;4(1); p. 25--36.

\bibitem{SPBO}
Das B, Mukherjee V, Das D.
\newblock Student psychology based optimization algorithm: A new population
  based optimization algorithm for solving optimization problems.
\newblock Advances in Engineering Software. 2020;146; p. 102804.

\bibitem{Nemati2024}
Nemati M, Zandi Y, Agdas AS.
\newblock Application of a novel metaheuristic algorithm inspired by stadium
  spectators in global optimization problems.
\newblock Scientific Reports. 2024;14; p. 3078.

\bibitem{TTA}
Rashid MFFA.
\newblock Tiki-taka algorithm: a novel metaheuristic inspired by football
  playing style.
\newblock Engineering Computations. 2020;38; p. 313--343.

\bibitem{teamgame}
Mahmoodabadi MJ, Rasekh M, Zohari T.
\newblock TGA: Team game algorithm.
\newblock Future Computing and Informatics Journal. 2018;3(2); p. 191--199.

\bibitem{TLBO}
Rao RV, Savsani VJ, Vakharia DP.
\newblock Teaching–learning-based optimization: A novel method for
  constrained mechanical design optimization problems.
\newblock Computer-Aided Design. 2011;43(3); p. 303--315.

\bibitem{TPOA}
Bagheri H, Ara AL, Hosseini R.
\newblock Thieves and Police, a New Optimization Algorithm: Theory and
  Application in Probabilistic Power Flow.
\newblock IETE Journal of Research. 2021;67(6); p. 951--968.

\bibitem{Li2024}
Li Z, Gao X, Huang X, Gao J, Yang X, Li MJ.
\newblock Tactical unit algorithm: A novel metaheuristic algorithm for optimal
  loading distribution of chillers in energy optimization.
\newblock Applied Thermal Engineering. 2024;238; p. 122037.

\bibitem{TWO}
Kaveh A, Zolghadr A.
\newblock A Novel Meta-Heuristic Algorithm: Tug Of War Optimization.
\newblock International Journal Of Optimization In Civil Engineering.
  2014;6(4); p. 469--492.

\bibitem{Ardjmand2012233}
Ardjmand E, Amin-Naseri MR.
\newblock Unconscious Search - A New Structured Search Algorithm for Solving
  Continuous Engineering Optimization Problems Based on the Theory of
  Psychoanalysis.
\newblock In: Advances in Swarm Intelligence; 2012. p. 233--242.

\bibitem{VPL}
Moghdani R, Salimifard K.
\newblock Volleyball Premier League Algorithm.
\newblock Applied Soft Computing. 2018;64; p. 161--185.

\bibitem{Yampolskiy2011358}
Yampolskiy RV, EL-Barkouky A.
\newblock Wisdom of artificial crowds algorithm for solving NP-hard problems.
\newblock International Journal of Bio-Inspired Computation. 2011;3(6); p.
  358--369.

\bibitem{FOA1}
Ghaemi M, Feizi-Derakhshi MR.
\newblock Forest Optimization Algorithm.
\newblock Expert Systems with Applications. 2014;41(15); p. 6676--6687.

\bibitem{artificialflora}
Cheng L, Wu Xh, Wang Y.
\newblock Artificial Flora (AF) Optimization Algorithm.
\newblock Applied Sciences. 2018;8(3); p. 329.

\bibitem{APO}
{Zhao} Z, {Cui} Z, {Zeng} J, {Yue} X.
\newblock Artificial Plant Optimization Algorithm for Constrained Optimization
  Problems.
\newblock In: 2011 Second International Conference on Innovations in
  Bio-inspired Computing and Applications; 2011. p. 120--123.

\bibitem{BVOA}
Ghaemidizaji M, Dadkhah C, Leung H.
\newblock A new optimization algorithm based on the behavior of BrunsVigia
  flower.
\newblock In: 2018 IEEE International Conference on Systems, Man, and
  Cybernetics (SMC); 2018. p. 263--267.

\bibitem{OngMeng2021}
Ong KM, Ong P, Sia CK.
\newblock A carnivorous plant algorithm for solving global optimization
  problems.
\newblock Applied Soft Computing. 2021;98; p. 106833.

\bibitem{Korani2019}
Korani W, Mouhoub M.
\newblock Discrete {M}other {T}ree {O}ptimization for the {T}raveling
  {S}alesman {P}roblem.
\newblock In: Neural Information Processing: 27th International Conference,
  ICONIP 2020, Bangkok, Thailand, November 23–27, 2020, Proceedings, Part II;
  2020. p. 25–--37.

\bibitem{FPA}
Yang XS.
\newblock Flower Pollination Algorithm for Global Optimization.
\newblock In: Unconventional Computation and Natural Computation, Proceeding;
  2012. p. 240--249.

\bibitem{Dalirinia2023}
Dalirinia E, Jalali M, Yaghoobi M, Tabatabaee H.
\newblock Lotus effect optimization algorithm {{(LEA)}}: a lotus
  nature-inspired algorithm for engineering design optimization.
\newblock The Journal of Supercomputing. 2023;80; p. 761--799.

\bibitem{Moez2016}
Moez H, Kaveh A, Taghizadieh N.
\newblock Natural Forest Regeneration Algorithm: A New Meta-Heuristic.
\newblock Iranian Journal of Science and Technology, Transactions of Civil
  Engineering. 2016;40(4); p. 311--326.

\bibitem{PGOplants}
{Cai} W, {Yang} W, {Chen} X.
\newblock A Global Optimization Algorithm Based on Plant Growth Theory: Plant
  Growth Optimization.
\newblock In: 2008 International Conference on Intelligent Computation
  Technology and Automation (ICICTA). vol.~1; 2008. p. 1194--1199.

\bibitem{PPA1}
Sulaiman M, Salhi A, Selamoglu BI, Kirikchi OB.
\newblock A Plant Propagation Algorithm for Constrained Engineering
  Optimisation Problems.
\newblock Mathematical Problems in Engineering. 2014;2014; p. 1--10.

\bibitem{PFA}
{Premaratne} U, {Samarabandu} J, {Sidhu} T.
\newblock A new biologically inspired optimization algorithm.
\newblock In: 2009 International Conference on Industrial and Information
  Systems (ICIIS); 2009. p. 279--284.

\bibitem{plantroot}
{Xiaoxian He}, {Shigeng Zhang}, {Jie Wang}.
\newblock A novel algorithm inspired by plant root growth with self-similarity
  propagation.
\newblock In: 2015 1st International Conference on Industrial Networks and
  Intelligent Systems (INISCom); 2015. p. 157--162.

\bibitem{yacineplants}
Labbi Y, ben attous D, A~Gabbar H, Mahdad B, Zidan A.
\newblock A new rooted tree optimization algorithm for economic dispatch with
  valve-point effect.
\newblock International Journal of Electrical Power \& Energy Systems. 2016;79;
  p. 298--311.

\bibitem{RRA}
Merrikh-Bayat F.
\newblock The runner-root algorithm: A metaheuristic for solving unimodal and
  multimodal optimization problems inspired by runners and roots of plants in
  nature.
\newblock Applied Soft Computing. 2015;33; p. 292--303.

\bibitem{Karci2007450}
Karci A.
\newblock Theory of Saplings Growing Up Algorithm.
\newblock In: Adaptive and Natural Computing Algorithms; 2007. p. 450--460.

\bibitem{SDMA}
Caraveo C, Valdez F, Castillo O.
\newblock A new optimization meta-heuristic algorithm based on self-defense
  mechanism of the plants with three reproduction operators.
\newblock Soft Computing. 2018;22(15); p. 4907--4920.

\bibitem{Emami2022}
Emami H.
\newblock Seasons optimization algorithm.
\newblock Engineering with Computers. 2022;38(2); p. 1845--1865.

\bibitem{Strawberryplant}
Merrikh-Bayat F. A Numerical Optimization Algorithm Inspired by the Strawberry
  Plant; 2014.

\bibitem{Naseri2020}
Naseri NK, Sundararajan EA, Ayob M, Jula A.
\newblock Smart {R}oot {S}earch {{(SRS)}}: A {N}ovel {N}ature-{I}nspired
  {S}earch {A}lgorithm.
\newblock Symmetry. 2020;12(12); p. 2025.

\bibitem{CHERAGHALIPOUR2018393}
Cheraghalipour A, Hajiaghaei-Keshteli M, Paydar MM.
\newblock Tree Growth Algorithm (TGA): A novel approach for solving
  optimization problems.
\newblock Engineering Applications of Artificial Intelligence. 2018;72; p.
  393--414.

\bibitem{plantphysiology}
Halim H, Ismail I.
\newblock Tree Physiology Optimization in Constrained Optimization Problem.
\newblock Telkomnika (Telecommunication Computing Electronics and Control).
  2018;16; p. 876--882.

\bibitem{KIRAN20156686}
Kiran MS.
\newblock TSA: Tree-seed algorithm for continuous optimization.
\newblock Expert Systems with Applications. 2015;42(19); p. 6686--6698.

\bibitem{YYOP}
Punnathanam V, Kotecha P.
\newblock Yin-Yang-pair Optimization: A novel lightweight optimization
  algorithm.
\newblock Engineering Applications of Artificial Intelligence. 2016;54; p.
  62--79.

\bibitem{SA1}
{Felipe} D, {Goldbarg} EFG, {Goldbarg} MC.
\newblock Scientific algorithms for the Car Renter Salesman Problem.
\newblock In: 2014 IEEE Congress on Evolutionary Computation (CEC); 2014. p.
  873--879.

\bibitem{fardsocial}
Fathollahi-Fard AM, Hajiaghaei-Keshteli M, Tavakkoli-Moghaddam R.
\newblock The Social Engineering Optimizer (SEO).
\newblock Engineering Applications of Artificial Intelligence. 2018;72; p.
  267--293.

\bibitem{SFS}
Salimi H.
\newblock Stochastic Fractal Search: A powerful metaheuristic algorithm.
\newblock Knowledge-Based Systems. 2015;75; p. 1--18.

\bibitem{Toz2023re}
Toz M, Toz G.
\newblock Re-formulated snowflake optimization algorithm {{(SFO-R)}}.
\newblock Evolutionary Intelligence. 2023;p. 1--20.

\bibitem{SGA}
Gonçalves MS, Lopez RH, Fadel~Miguel LF.
\newblock Search group algorithm: A new metaheuristic method for the
  optimization of truss structures.
\newblock Computers and Structures. 2015;153; p. 165--184.

\bibitem{hasanccebi2012efficient}
Hasan{\c{c}}ebi O, Azad SK.
\newblock An efficient metaheuristic algorithm for engineering optimization:
  SOPT.
\newblock International Journal of Optimization in Civil Engineering.
  2012;2(4); p. 479--487.

\bibitem{Chu2024}
Chu SC, Wang TT, Yildiz AR, Pan JS.
\newblock Ship {R}escue {O}ptimization: A {N}ew {M}etaheuristic {A}lgorithm for
  {S}olving {E}ngineering {P}roblems.
\newblock Journal of Internet Technology. 2024;25(1); p. 61--78.

\bibitem{SWO1}
Du H, Wu X, Zhuang J.
\newblock Small-World Optimization Algorithm for Function Optimization.
\newblock In: Advances in Natural Computation; 2006. p. 264--273.

\bibitem{Dueck199386}
Dueck G.
\newblock New optimization heuristics; The great deluge algorithm and the
  record-to-record travel.
\newblock Journal of Computational Physics. 1993;104(1); p. 86--92.

\bibitem{WDO}
{Bayraktar} Z, {Komurcu} M, {Werner} DH.
\newblock Wind Driven Optimization (WDO): A novel nature-inspired optimization
  algorithm and its application to electromagnetics.
\newblock In: 2010 IEEE Antennas and Propagation Society International
  Symposium; 2010. p. 1--4.

\bibitem{ACM}
{Gao-Wei} Y, {Zhanju} H.
\newblock A Novel Atmosphere Clouds Model Optimization Algorithm.
\newblock In: 2012 International Conference on Computing, Measurement, Control
  and Sensor Network; 2012. p. 217--220.

\bibitem{ACS}
Civicioglu P.
\newblock Artificial cooperative search algorithm for numerical optimization
  problems.
\newblock Information Sciences. 2012;229; p. 58--76.

\bibitem{AIG}
Pijarski P, Kacejko P.
\newblock A new metaheuristic optimization method: the algorithm of the
  innovative gunner (AIG).
\newblock Engineering Optimization. 2019;51(12); p. 1--21.

\bibitem{ANS}
Wu G.
\newblock Across neighborhood search for numerical optimization.
\newblock Information Sciences. 2016;329; p. 597--618.

\bibitem{Hubalovska2024}
Hub{\'a}lovsk{\'a} M, Hub{\'a}lovsk{\`y} {\v{S}}, Trojovsk{\`y} P.
\newblock Botox {O}ptimization {A}lgorithm: A {N}ew {H}uman-{B}ased
  {M}etaheuristic {A}lgorithm for {S}olving {O}ptimization {P}roblems.
\newblock Biomimetics. 2024;9(3); p. 137.

\bibitem{BRO}
Rahkar~Farshi T.
\newblock Battle royale optimization algorithm.
\newblock Neural Computing and Applications. 2021;33(4); p. 1139--1157.

\bibitem{DelAcebo200818}
Del~Acebo E, De~La~Rosa JL.
\newblock Introducing bar systems: A class of swarm intelligence optimization
  algorithms.
\newblock In: AISB 2008 Convention: Communication, Interaction and Social
  Intelligence - Proceedings of the AISB 2008 Symposium on Swarm Intelligence
  Algorithms and Applications; 2008. p. 18--23.

\bibitem{BSO1}
Civicioglu P.
\newblock Backtracking Search Optimization Algorithm for numerical optimization
  problems.
\newblock Applied Mathematics and Computation. 2012;219(15); p. 8121--8144.

\bibitem{Zhu201255}
{Zhu} C, {Ni} J.
\newblock Cloud Model-Based Differential Evolution Algorithm for Optimization
  Problems.
\newblock In: 2012 Sixth International Conference on Internet Computing for
  Science and Engineering; 2012. p. 55--59.

\bibitem{Li1998409}
Li B, Jiang W.
\newblock Optimizing complex functions by chaos search.
\newblock Cybernetics and Systems. 1998;29(4); p. 409--419.

\bibitem{CSA}
Nunes~de Castro L, Von~Zuben FJ.
\newblock The Clonal Selection Algorithm with Engineering Applications.
\newblock In: Workshop Proceedings of GECCO. vol.~10; 2000. p. 36--37.

\bibitem{CVA}
Hosseini E, Ghafoor KZ, Sadiq AS, Guizani M, Emrouznejad A.
\newblock COVID-19 Optimizer Algorithm, Modeling and Controlling of Coronavirus
  Distribution Process.
\newblock IEEE Journal of Biomedical and Health Informatics. 2020;24(10); p.
  2765--2775.

\bibitem{DGO}
Dehghani M, Montazeri Z, Malik OP.
\newblock DGO: Dice game optimizer.
\newblock Gazi University Journal of Science. 2019;32(3); p. 871--882.

\bibitem{dialecticsearch}
Kadioglu S, Sellmann M.
\newblock Dialectic Search.
\newblock In: Principles and Practice of Constraint Programming - CP 2009;
  2009. p. 486--500.

\bibitem{DSA}
Civicioglu P.
\newblock transforming geocentric cartesian coordinates to geodetic coordinates
  by using differential search algorithm.
\newblock Computers and Geosciences. 2012;46; p. 229--247.

\bibitem{EMA}
Ghorbani N, Babaei E.
\newblock Exchange market algorithm.
\newblock Applied Soft Computing. 2014;19; p. 177--187.

\bibitem{Boettcher:1999:EOM:2933923.2934033}
Boettcher S, Percus AG.
\newblock Extremal Optimization: Methods Derived from Co-evolution.
\newblock In: Proceedings of the 1st Annual Conference on Genetic and
  Evolutionary Computation. vol.~1; 1999. p. 825--832.

\bibitem{Faramarzi2020}
Faramarzi A, Heidarinejad M, Stephens B, Mirjalili S.
\newblock Equilibrium optimizer: A novel optimization algorithm.
\newblock Knowledge-Based Systems. 2020;191; p. 105190.

\bibitem{FAO}
Tan Y, Zhu Y.
\newblock Fireworks Algorithm for Optimization.
\newblock In: Advances in Swarm Intelligence; 2010. p. 355--364.

\bibitem{FFA}
Shayanfar H, Gharehchopogh FS.
\newblock Farmland fertility: A new metaheuristic algorithm for solving
  continuous optimization problems.
\newblock Applied Soft Computing. 2018;71; p. 728--746.

\bibitem{GEM}
Ahrari A, Atai AA.
\newblock Grenade Explosion Method—A novel tool for optimization of
  multimodal functions.
\newblock Applied Soft Computing. 2010;10; p. 1132--1140.

\bibitem{tanyildizi2017golden}
Tanyildizi E, Demir G.
\newblock Golden sine algorithm: a novel math-inspired algorithm.
\newblock Advances in Electrical and Computer Engineering. 2017;17(2); p.
  71--79.

\bibitem{Husseinzadeh2024}
Husseinzadeh~Kashan A, Karimiyan S, Kulkarni AJ.
\newblock The {G}olf {S}port {I}nspired {S}earch metaheuristic algorithm and
  the game theoretic analysis of its operators’ effectiveness.
\newblock Soft Computing. 2024;28(2); p. 1073--1125.

\bibitem{HO1}
Hatamlou A.
\newblock Heart: a novel optimization algorithm for cluster analysis.
\newblock Progress in Artificial Intelligence. 2014;2(2); p. 167--173.

\bibitem{hyperparamdialectic}
Sellmann M, Tierney K.
\newblock Hyper-parameterized Dialectic Search for Non-linear Box-Constrained
  Optimization with Heterogenous Variable Types.
\newblock In: Learning and Intelligent Optimization; 2020. p. 102--116.

\bibitem{hosseini2017ideological}
Hosseini SH, Ebrahimi A. Ideological Sublations: Resolution of Dialectic in
  Population-based Optimization; 2017.

\bibitem{ISA}
Gandomi AH.
\newblock Interior search algorithm (ISA): A novel approach for global
  optimization.
\newblock ISA Transactions. 2014;53(4); p. 1168--1183.

\bibitem{KA}
Hajiaghaei-Keshteli M, Aminnayeri M.
\newblock Solving the integrated scheduling of production and rail
  transportation problem by Keshtel algorithm.
\newblock Applied Soft Computing. 2014;25; p. 184--203.

\bibitem{KA1}
Jaddi NS, Alvankarian J, Abdullah S.
\newblock Kidney-inspired algorithm for optimization problems.
\newblock Communications in Nonlinear Science and Numerical Simulation.
  2017;42; p. 358--369.

\bibitem{KP}
De~Melo VV.
\newblock Kaizen Programming.
\newblock In: Proceedings of the 2014 Annual Conference on Genetic and
  Evolutionary Computation; 2014. p. 895--902.

\bibitem{Houssein2023}
Houssein EH, Oliva D, Samee NA, Mahmoud NF, Emam MM.
\newblock Liver {C}ancer {A}lgorithm: A novel bio-inspired optimizer.
\newblock Computers in Biology and Medicine. 2023;165; p. 107389.

\bibitem{NiLei2024}
Ni L, Ping Y, Yao N, Jiao J, Wang G.
\newblock Literature {R}esearch {O}ptimizer: A {N}ew {H}uman-{B}ased
  {M}etaheuristic {A}lgorithm for {O}ptimization {P}roblems.
\newblock Arabian Journal for Science and Engineering. 2024;p. 1--49.

\bibitem{Nishida200655}
Nishida TY.
\newblock Membrane Algorithms: Approximate Algorithms for NP-Complete
  Optimization Problems.
\newblock In: Applications of Membrane Computing. Springer Berlin Heidelberg;
  2006. p. 303--314.

\bibitem{MBA}
Sadollah A, Bahreininejad A, Eskandar H, Hamdi M.
\newblock Mine blast algorithm: A new population based algorithm for solving
  constrained engineering optimization problems.
\newblock Applied Soft Computing. 2013;13(5); p. 2592--2612.

\bibitem{asil2017new}
Asil~Gharebaghi S, Ardalan~Asl M.
\newblock New meta-heuristic optimization algorithm using neuronal
  communication.
\newblock Iran University of Science \& Technology. 2017;7(3); p. 413--431.

\bibitem{Khouni2024}
Khouni SE, Menacer T.
\newblock Nizar optimization algorithm: A novel metaheuristic algorithm for
  global optimization and engineering applications.
\newblock The Journal of Supercomputing. 2024;80(3); p. 3229--3281.

\bibitem{Kaveh2020}
Kaveh A, Akbari H, Hosseini SM.
\newblock Plasma generation optimization: A new physically-based metaheuristic
  algorithm for solving constrained optimization problems.
\newblock Engineering Computations. 2020;38(4); p. 1554--1606.

\bibitem{chan2012hyper}
Chan CY, Xue F, Ip W, Cheung C.
\newblock A hyper-heuristic inspired by pearl hunting.
\newblock In: International Conference on Learning and Intelligent
  Optimization; 2012. p. 349--353.

\bibitem{PVS}
Savsani P, Savsani V.
\newblock Passing vehicle search (PVS): A novel metaheuristic algorithm.
\newblock Applied Mathematical Modelling. 2016;40(5--6); p. 3951--3978.

\bibitem{RDA}
Jiang Q, Wang L, Hei X, Fei R, Yang D, Zou F, et~al.
\newblock Optimal approximation of stable linear systems with a novel and
  efficient optimization algorithm.
\newblock In: Proceedings of the {IEEE} Congress on Evolutionary Computation,
  {CEC}; 2014. p. 840--844.

\bibitem{reactivedialectic}
Ans\'{o}tegui C, Pon J, Sellmann M, Tierney K.
\newblock Reactive Dialectic Search Portfolios for MaxSAT.
\newblock In: Proceedings of the Thirty-First AAAI Conference on Artificial
  Intelligence; 2017. p. 765–--772.

\bibitem{ACOBook}
Dorigo M, St{\"u}tzle T.
\newblock {Ant Colony Optimization}.
\newblock MIT Press; 2004.

\bibitem{metaphor}
S\"{o}rensen K.
\newblock Metaheuristics - the metaphor exposed.
\newblock International Transactions in Operational Research. 2015;22; p.
  3--18.

\bibitem{review11}
Garc{\'i}a-Mart{\'i}nez C, Guti{\'e}rrez PD, Molina D, Lozano M, Herrera F.
\newblock Since CEC 2005 competition on real-parameter optimisation: a decade
  of research, progress and comparative analysis's weakness.
\newblock Soft Computing. 2017;21(19); p. 5573--5583.

\bibitem{PerfTuned}
Liao T, Molina D, Stützle T.
\newblock Performance evaluation of automatically tuned continuous optimizers
  on different benchmark sets.
\newblock Applied Soft Computing. 2015;27; p. 490--503.

\bibitem{Bosman2018}
Bosman PAN, Gallagher M.
\newblock The importance of implementation details and parameter settings in
  black-box optimization: a case study on Gaussian estimation-of-distribution
  algorithms and circles-in-a-square packing problems.
\newblock Soft Computing. 2018;22(4); p. 1209--1223.

\bibitem{Biedrzycki2019247}
Biedrzycki R.
\newblock On Equivalence of Algorithm's Implementations: The {{CMA}}-{{ES}}
  Algorithm and Its Five Implementations.
\newblock In: Proceedings of the {{Genetic}} and {{Evolutionary Computation
  Conference Companion}}; 2019. p. 247--248.

\bibitem{ParadisEO}
Liefooghe A, Jourdan L, Talbi EG.
\newblock A software framework based on a conceptual unified model for
  evolutionary multiobjective optimization: ParadisEO-MOEO.
\newblock European Journal of Operational Research. 2011;209(2); p. 104--112.

\bibitem{jMetal}
Durillo JJ, Nebro AJ.
\newblock jMetal: A Java framework for multi-objective optimization.
\newblock Advances in Engineering Software. 2011;42(10); p. 760--771.

\bibitem{NiaPyJOSS2018}
Vrban{\v{c}}i{\v{c}} G, Brezo{\v{c}}nik L, Mlakar U, Fister D, {Fister Jr } I.
\newblock {NiaPy: Python microframework for building nature-inspired
  algorithms}.
\newblock {Journal of Open Source Software}. 2018;3(23); p. 613.

\bibitem{BENITEZHIDALGO2019100598}
Benítez-Hidalgo A, Nebro AJ, García-Nieto J, Oregi I, Ser JD.
\newblock jMetalPy: A Python framework for multi-objective optimization with
  metaheuristics.
\newblock Swarm and Evolutionary Computation. 2019;51; p. 100598.

\bibitem{PlatEMO}
Tian Y, Cheng R, Zhang X, Jin Y.
\newblock {PlatEMO}: A {MATLAB} platform for evolutionary multi-objective
  optimization.
\newblock IEEE Computational Intelligence Magazine. 2017;12(4); p. 73--87.

\bibitem{valadi2014applications}
Valadi J, Siarry P.
\newblock Applications of metaheuristics in process engineering. vol.~31.
\newblock Springer; 2014.

\bibitem{gandomi2013metaheuristic}
Gandomi AH, Yang XS, Talatahari S, Alavi AH.
\newblock Metaheuristic {A}pplications in {S}tructures and {I}nfrastructures.
\newblock New York: Springer; 2013.

\bibitem{griffis2012metaheuristics}
Griffis SE, Bell JE, Closs DJ.
\newblock Metaheuristics in logistics and supply chain management.
\newblock Journal of Business Logistics. 2012;33(2); p. 90--106.

\bibitem{Starodubcev2022}
Starodubcev NO, Nikitin NO, Kalyuzhnaya AV.
\newblock Surrogate-Assisted Evolutionary Generative Design Of Breakwaters
  Using Deep Convolutional Networks.
\newblock In: 2022 IEEE Congress on Evolutionary Computation (CEC); 2022. p.
  1--8.

\bibitem{Lohn2005evolved}
Lohn JD, Hornby GS, Linden DS.
\newblock An evolved antenna for deployment on nasa’s space technology 5
  mission.
\newblock Genetic Programming Theory and Practice II. 2005;p. 301--315.

\bibitem{Schmit2009}
Schmidt M, Lipson H.
\newblock Symbolic regression of implicit equations.
\newblock In: Genetic programming theory and practice VII. Springer; 2009. p.
  73--85.

\bibitem{Linan2023}
Li N, Ma L, Xing T, Yu G, Wang C, Wen Y, et~al.
\newblock Automatic design of machine learning via evolutionary computation: A
  survey.
\newblock Applied Soft Computing. 2023;143; p. 110412.

\bibitem{Song2019}
Song H, Triguero I, {\"O}zcan E.
\newblock A review on the self and dual interactions between machine learning
  and optimisation.
\newblock Progress in Artif Intell. 2019;8(2); p. 143--165.

\bibitem{automlzero}
Real E, Liang C, So D, Le Q.
\newblock {A}uto{ML}-{Z}ero: Evolving {M}achine {L}earning {A}lgorithms {F}rom
  {S}cratch.
\newblock In: Proceedings of the 37th International Conference on Machine
  Learning. vol. 119; 2020. p. 8007--8019.

\bibitem{Yuqiao2023}
Liu Y, Sun Y, Xue B, Zhang M, Yen GG, Tan KC.
\newblock {A Survey on Evolutionary Neural Architecture Search}.
\newblock IEEE Transactions on Neural Networks and Learning Systems.
  2023;34(2); p. 550--570.

\bibitem{Dufourq2017}
Dufourq E, Bassett BA.
\newblock {EDEN}: {E}volutionary deep networks for efficient machine learning.
\newblock In: Pattern Recognition Association of South Africa and Robotics and
  Mechatronics (PRASA-RobMech); 2017. p. 110--115.

\bibitem{Assunccao2019}
Assun{\c{c}}ao F, Louren{\c{c}}o N, Machado P, Ribeiro B.
\newblock {{DENSER}}: deep evolutionary network structured representation.
\newblock Genetic Programming and Evolvable Machines. 2019;20; p. 5--35.

\bibitem{Wang2019}
Wang R, Lehman J, Clune J, Stanley KO. Paired {O}pen-{E}nded {T}railblazer
  {{(POET)}}: {E}ndlessly {G}enerating {I}ncreasingly {C}omplex and {D}iverse
  {L}earning {E}nvironments and {T}heir {S}olutions; 2019.

\bibitem{Akiba2024}
Akiba T, et~al.. Evolutionary {O}ptimization of {M}odel {M}erging {R}ecipes;
  2024.

\bibitem{He2021}
He X, Zhao K, Chu X.
\newblock AutoML: A survey of the state-of-the-art.
\newblock Knowledge-Based Systems. 2021;212; p. 106622.

\bibitem{JasonMa2023}
Ma YJ, Liang W, Wang G, Huang DA, Bastani O, Jayaraman D, et~al.. Eureka:
  {H}uman-{L}evel {R}eward {D}esign via {C}oding {L}arge {L}anguage {M}odels;
  2023.

\bibitem{Zhao2023}
Zhao H, Ning X, Liu X, Wang C, Liu J.
\newblock What makes evolutionary multi-task optimization better: A
  comprehensive survey.
\newblock Appl Soft Comput. 2023 Sep;145; p. 110545.

\bibitem{Lopez2021}
L\'{o}pez-ib\'{a}nez M, Branke J, Paquete L.
\newblock Reproducibility in Evolutionary Computation.
\newblock ACM Transactions on Evolutionary Learning and Optimization.
  2021;1(4).

\bibitem{Wu2024evolutionary}
Wu X, hao Wu S, Wu J, Feng L, Tan KC. Evolutionary {C}omputation in the {E}ra
  of {L}arge {L}anguage {M}odel: Survey and {R}oadmap; 2024.

\bibitem{Triguero2024}
Triguero I, Molina D, Poyatos J, {Del Ser} J, Herrera F.
\newblock General {P}urpose {A}rtificial {I}ntelligence {S}ystems {{(GPAIS)}}:
  Properties, definition, taxonomy, societal implications and responsible
  governance.
\newblock Information Fusion. 2024;103; p. 102135.

\bibitem{Guo2024}
Guo Q, et~al.
\newblock Connecting {L}arge {L}anguage {M}odels with {E}volutionary
  {A}lgorithms {Y}ields {P}owerful {P}rompt {O}ptimizers.
\newblock In: The Twelfth International Conference on Learning Representations;
  2024. .

\bibitem{Chen2023}
Chen S, Chen S, Hou W, Ding W, You X.
\newblock {EGANS}: {E}volutionary {G}enerative {A}dversarial {N}etwork {S}earch
  for {Z}ero-{S}hot {L}earning.
\newblock IEEE Transactions on Evolutionary Computation. 2023;p. In Press.

\end{thebibliography}

\end{document}